\documentclass[twoside,11pt]{article}
\usepackage{jair, theapa, rawfonts}

\jairheading{1}{1993}{1-15}{6/91}{9/91}
\ShortHeadings{Inductive Logic Programming At 30: A New Introduction}{Cropper and Dumančić}
\firstpageno{1}

\usepackage[utf8]{inputenc}

\usepackage[sfdefault,scaled=.85]{FiraSans}
\usepackage[T1]{fontenc}
\usepackage{textcomp}
\usepackage{amsmath,amsthm}
\usepackage[cmintegrals]{newtxsf}

\usepackage{mathtools}
\usepackage{inconsolata}
\usepackage{booktabs}
\usepackage{listings}
\usepackage{xcolor}
\usepackage{url}

\usepackage{scalerel}
\newcommand\fork{\mathbin{\ThisStyle{{\supset}\kern-\dimexpr.5\LMex+3pt\relax{-}}}}

\usepackage{amsthm}
\usepackage{amsmath}
\usepackage{tablefootnote}
\usepackage{enumitem}
\usepackage[normalem]{ulem}

\theoremstyle{definition}
\newtheorem{definition}{Definition}
\newtheorem{example}{Example}

\usepackage{tikz}
\usetikzlibrary{arrows}
\usetikzlibrary{arrows.meta}
\usetikzlibrary{decorations.markings}
\usetikzlibrary{positioning}

\newcommand{\mc}[1]{$\mathcal{#1}$}
\newcommand{\dilp}{$\partial$ILP}
\newcommand{\tw}[1]{\texttt{#1}}

\newenvironment{bk}
{
\ttfamily
\[\tw{B} = \left\{\begin{array}{l}
}
{
\end{array}\right\}\]
\par
}

\newenvironment{modes}
{
\ttfamily
\[\tw{M} = \left\{\begin{array}{l}
}
{
\end{array}\right\}\]
\par
}

\newenvironment{hyp}
{\[\tw{H} = \left\{\begin{array}{l}}
{\end{array}\right\}\]}

\newenvironment{epos}
{\[\tw{E$^+$} = \left\{\begin{array}{l}}
{\end{array}\right\}\]}

\newenvironment{exs}
{\[\tw{E} = \left\{\begin{array}{l}}
{\end{array}\right\}\]}

\newenvironment{code}
{
\ttfamily
\begin{center}
\begin{tabular}{l}
}
{
\end{tabular}
\end{center}
\par
}

\title{Inductive Logic Programming At 30: A New Introduction}

\author{
       \name Andrew Cropper \email andrew.cropper@cs.ox.ac.uk\\
       \addr University of Oxford
       \AND
       \name Sebastijan Dumančić \email s.dumancic@tudelft.nl\\
       \addr TU Delft
}

\begin{document}

\maketitle

\begin{abstract}
Inductive logic programming (ILP) is a form of machine learning.
The goal of ILP is to induce a hypothesis (a set of logical rules) that generalises training examples.
As ILP turns 30, we provide a new introduction to the field.
We introduce the necessary logical notation and the main learning settings; describe the building blocks of an ILP system; compare several systems on several dimensions; describe four systems (Aleph, TILDE, ASPAL, and Metagol); highlight key application areas; and, finally, summarise current limitations and directions for future research.
\end{abstract}

\section{Introduction}
\label{sec:intro}

A remarkable feat of human intelligence is the ability to learn knowledge.
A key form of learning is \emph{induction}: the process of forming general rules (hypotheses) from specific observations (examples).
For instance, suppose you draw 10 red balls out of a bag, then you might induce a hypothesis (a rule) that all the balls in the bag are red.
Having induced this hypothesis, you can predict the colour of the next ball out of the bag.

Machine learning (ML) automates induction.
ML induces a hypothesis (also called a \emph{model}) that generalises training examples (observations).
For instance, given labelled images of cats and dogs, the goal of ML is to induce a hypothesis that predicts whether an unlabelled image is a cat or a dog.
Inductive logic programming (ILP) \shortcite{mugg:ilp} is a form of ML.
As with other forms of ML, the goal of ILP is to induce a hypothesis that generalises training examples.
However, whereas most forms of ML use tables\footnote{
    Table-based learning is \emph{attribute-value} learning.
    See \shortciteA{luc:book} for an overview of the hierarchy of representations. Note that not \emph{all} other forms of ML use tables. For instance, \shortciteA{DBLP:conf/nips/Rocktaschel017} use embeddings.
} to represent data (examples and hypotheses), ILP uses \emph{logic programs} (sets of logical rules).
Moreover, whereas most forms of ML learn functions, ILP learns relations.
We illustrate ILP using four scenarios.

\subsection{Scenario 1: Concept Learning}
\label{ex:happy}
Suppose we want to predict whether someone is happy.
To do so, we ask four people (\emph{alice}, \emph{bob}, \emph{claire}, and \emph{dave}) whether they are happy.
We also ask for additional information, specifically their job, their company, and whether they like lego.
Many ML approaches, such as a decision tree or neural network learner, would represent this data as a table, such as Table \ref{tab:intro-data-table}.
Using standard ML terminology, each row represents a training example, the first three columns (\emph{name}, \emph{job}, and \emph{enjoys lego}) represent \emph{features}, and the final column (\emph{happy}) represents the \emph{label} or \emph{classification}.
Given this table as input, the goal is to induce a hypothesis that generalises the training examples.
For instance, a neural network learner would learn a table of numbers that weight the importance of the features (or hidden features in a multi-layer network).
We can then use the hypothesis to predict labels for unseen examples.

\begin{table}[ht]
\centering
\begin{tabular}{@{}c|ccc@{}}
\textbf{Name}   & \textbf{Job}  &\textbf{Enjoys lego} & \textbf{Happy} \\
\midrule
\emph{alice}  & lego builder & yes & yes   \\
\emph{bob}    & lego builder & no & no    \\
\emph{claire} & estate agent & yes & no     \\
\emph{dave}   & estate agent & no & no    \\
\end{tabular}
\caption{A table representation of a ML task.}
\label{tab:intro-data-table}
\end{table}

\noindent
Rather than represent data as tables, ILP represents data as logic programs, sets of logical rules.
The main building block of a logic program is an \emph{atom}.
An atom is of the form $p(x_1,\dots,x_n)$, where $p$ is a \emph{predicate} symbol of arity $n$ (takes $n$ arguments) and each $x_i$ is a \emph{term}.
A logic program uses atoms to represent data.
For instance, we can represent that \emph{alice} enjoys lego as the atom \tw{enjoys\_lego(alice)} and that \emph{bob} is a lego builder as \tw{lego\_builder(bob)}.

An ILP task is formed of three sets ($B$, $E^+$, $E^-$).
The set $B$ is \emph{background knowledge} (BK).
BK is similar to features but can contain relations and information indirectly associated with the examples.
We can represent the data in Table \ref{tab:intro-data-table} as the set $B$:
\begin{bk}
    \tw{lego\_builder(alice).}\\
    \tw{lego\_builder(bob).}\\
    \tw{estate\_agent(claire).}\\
    \tw{estate\_agent(dave).}\\
    \tw{enjoys\_lego(alice).}\\
    \tw{enjoys\_lego(claire).}
\end{bk}
\noindent
ILP usually follows the \emph{closed world assumption} \shortcite{cwa}, so if anything is not explicitly true we assume it is false.
With this assumption, we do not need to explicitly state that  \tw{enjoys\_lego(bob)} and \tw{enjoys\_lego(dave)} are false.

The sets $E^+$ and $E^-$ represent positive and negative examples respectively.
We can represent the examples in Table \ref{tab:intro-data-table} as:
\[
\tw{E}^+ = \left\{
\begin{array}{l}
\tw{happy(alice).}
\end{array}
\right\}
\tw{E}^- = \left\{
\begin{array}{l}
\tw{happy(bob).}\\
    \tw{happy(claire).}\\
    \tw{happy(dave).}
\end{array}
\right\}
\]

\noindent
Given these sets, the goal of ILP is to induce a hypothesis that with the BK logically entails as many positive and as few negative examples as possible.
A hypothesis ($H$) in ILP is a set of logical rules, such as:
\[
\tw{H} = \left\{\begin{array}{l}
\forall A. \;\; \tw{lego\_builder(A)} \land \tw{enjoys\_lego(A)} \rightarrow \tw{happy(A)}\\
\end{array}\right\}
\]

\noindent
This hypothesis contains one rule that says that for all \tw{A}, if \tw{A} is a lego builder (\tw{lego\_builder(A)}) and enjoys lego (\tw{enjoys\_lego(A)}), then \tw{A} is happy (\tw{happy(A)}).
Having induced a rule, we can deduce knowledge from it.
For instance, this rule says if \tw{lego\_builder(alice)} and \tw{enjoys\_lego(alice)} are true then \tw{happy(alice)} must also be true.

The above rule is written in a standard first-order logic notation.
We usually write logic programs in reverse implication form:
\[\tw{head} \tw{:- } \tw{body}_1, \tw{body}_2, \dots, \tw{body}_n \]
A rule in this form states that the \emph{head} atom is true when every \emph{body}$_i$ atom is true.
A comma denotes conjunction.
In logic programming, every variable is assumed to be universally quantified, so we drop quantifiers.
We also flip the direction of the implication symbol $\rightarrow$ to $\leftarrow$ and often replace it with \tw{:-} because it is easier to use when writing computer programs.
Therefore, in logic programming notation, the above hypothesis is:
\begin{hyp}
    \tw{happy(A):- lego\_builder(A),enjoys\_lego(A).}
\end{hyp}
Logic programs are \emph{declarative} which means that the order of atoms in a rule does not change its semantics\footnote{
    This statement about the declarative nature of logic programs is imprecise.
    The order of rules often matters in practice, such as in Prolog.
    We discuss this issue in detail in Section \ref{sec:logprogs}.
}.
For instance, the above hypothesis is semantically identical to this one:
\begin{hyp}
    \tw{happy(A):- enjoys\_lego(A),lego\_builder(A).}
\end{hyp}
\noindent




\subsection{Scenario 2: Data Curation}
\label{sec:intro:string}

Suppose we want to learn a string transformation programs from \emph{input} $\mapsto$ \emph{output} examples, such as program that returns the last character of a string:
\begin{center}
\begin{tabular}{@{}c|c@{}}
\textbf{Input} & \textbf{Output} \\
\midrule
machine & e \\
learning & g \\
algorithm & m \\
\end{tabular}
\end{center}
\noindent
Many forms of ML would represent these examples as a table, such as using a \emph{one-hot-encoding} technique\footnote{
  Using the simplest binary one-hot-encoding approach, we would have a feature for every letter and an example would have the value 1 for that feature if the letter appears in the example; otherwise, the value will be 0.
}.
By contrast, ILP represents these examples as atoms, such as:
\begin{epos}
    \tw{last([m,a,c,h,i,n,e], e).}\\
    \tw{last([l,e,a,r,n,i,n,g], g).}\\
    \tw{last([a,l,g,o,r,i,t,m], m).}\\
\end{epos}
\noindent
The symbol \tw{last} is the \emph{target predicate} that we want to learn (the relation to generalise).
The first argument of each atom represents an input list and the second argument represents an output value.
To induce a hypothesis for these examples, we need to provide suitable BK, such as common list operations:

\begin{center}
\small
\begin{tabular}{@{}l|l|l@{}}
\textbf{Name} & \textbf{Description} & \textbf{Example}\\
\midrule
empty(A) & \tw{A} is an empty list & \tw{empty([]).}\\
\midrule
head(A,B) & \tw{B} is the head of the list \tw{A} & \tw{head([c,a,t],c).}\\
\midrule
tail(A,B) & \tw{B} is the tail of the list \tw{A} & \tw{tail([c,a,t],[a,t]).}\\
\end{tabular}
\end{center}

\noindent
Given the aforementioned examples and BK with the above list operations,  the goal is to search for a hypothesis that generalises the examples.
At a high level, an ILP system builds a hypothesis by combining information from the BK and examples.
The set of all possible hypotheses is called the \emph{hypothesis space}.
In other words, the goal of an ILP system is to search the hypothesis space for a hypothesis that generalises the examples.

In this data curation scenario, an ILP system could induce the hypothesis:
\begin{hyp}
    \tw{last(A,B):- head(A,B),tail(A,C),empty(C).}\\
    \tw{last(A,B):- tail(A,C),last(C,B).}
\end{hyp}
\noindent
This hypothesis contains two rules.
The first rule says that \tw{B} is the last element of \tw{A} when \tw{B} is the head of \tw{A} and the tail of \tw{A} is empty.
The second rule says that \tw{B} is the last element of \tw{A} when
\tw{B} is the last element of the tail of \tw{A}.

\subsection{Scenario 3: Program Synthesis}
Suppose we have the following positive and negative examples, again represented as atoms, where the first argument is an unsorted list and the second argument is a sorted list:
\[
\tw{E}^+ = \left\{
\begin{array}{l}
\tw{sort([2,1],[1,2]).}\\
\tw{sort([5,3,1],[1,3,5]).}
\end{array}
\right\}
\tw{E}^- = \left\{
\begin{array}{l}
\tw{sort([2,1],[2,1]).}\\
\tw{sort([1,3,1],[1,1,1]).}
\end{array}
\right\}
\]
\noindent
Also suppose that as BK we have the same \tw{empty}, \tw{head}, and \tw{tail} relations from the string transformation scenario and two additional relations:
\begin{center}
\small
\begin{tabular}{@{}l|p{5cm}|p{6cm}@{}}
\textbf{Name} & \textbf{Description} & \textbf{Example}\\
\midrule
partition(Pivot,A,L,R) & \tw{L} is a sublist of \tw{A} containing elements less than or equal to \tw{Pivot} and \tw{R} is a sublist of \tw{A} containing elements greater than the pivot & \tw{pivot(3,[4,1,5,2],[1,2],[4,5]).}\\
\midrule
append(A,B,C) & true when \tw{C} is the concatenation of \tw{A} and \tw{B} &     \tw{append([a,b,c],[d,e],[a,b,c,d,e]).} \\
\end{tabular}
\end{center}

\noindent
Given these sets, an ILP system could induce the hypothesis:

\begin{hyp}
    \tw{sort(A,B):- empty(A),empty(B).}\\
    \tw{sort(A,B):- head(A,Pivot),partition(Pivot,A,L1,R1),}\\
    \hspace{23mm} \tw{sort(L1,L2),sort(R1,R2),append(L2,R2,B).}
\end{hyp}

\noindent
This hypothesis corresponds to the \emph{quicksort} algorithm \shortcite{qsort} and generalises to lists of arbitrary length and elements not seen in the training examples.
This scenario shows that ILP is a form of \emph{program synthesis} \shortcite{mis}, where the goal is to automatically build executable programs.

\subsection{Scenario 4: Scientific Discovery}

As \shortciteA{srinivasan1994mutagenesis} state, \emph{``There is more to scientific theory formulation than data fitting. To be acceptable, a theory must be understandable and open to critical analysis''}.
For this reason, ILP has been widely used for scientific discovery.
For instance, \shortciteA{King92} use ILP to model structure-activity relationships for drug design.
In this work, an ILP system takes as input positive and negative examples and BK.
The positive examples are paired examples of greater activity.
For instance, the positive example \tw{great(d20,di5)} states that \emph{drug 20} has higher activity than \emph{drug 15}.
Negative examples are examples of drug pairings with lower activity.
The BK contains information about the chemical structures of drugs and the properties of \emph{substituents}\footnote{An atom or group other than hydrogen on a molecule.}.
For instance, the atom \tw{struc(d35,no2,nhcoch3,h)} states that
\emph{drug 35} has \emph{no2} substituted at position \emph{3}, \emph{nhcoch3} substituted at position 4, and no substitution at position 5, and the atom \tw{flex(no2,3)} states that \emph{no2} has flexibility 3.
Given such examples and BK, the ILP system Golem \shortcite{golem} induces multiple rules to explain the examples, including this one:

\begin{code}
\tw{great(A,B):- struc(A,C,D,E),struc(B,F,h,h),h\_donor(C,hdonO),}\\
\hspace{25mm} \tw{polarisable(C,polaril),flex(F,G),flex(C,H),}\\
\hspace{25mm} \tw{great\_flex(G,H),great6\_flex(G).}\\
\end{code}

\noindent
This rule says that Drug A is better than drug B if
drug B has no substitutions at positions 4 and 5,
and drug B at position 3 has flexibility >6,
and drug A at position 3 has polarisability = 1,
and drug A at position 3 has hydrogen donor = 0,
and drug A at position 3 is less flexible than drug B at position 3.

As this scenario illustrates, ILP can learn human-readable hypotheses.
The interpretability of such rules is crucial to allow domain experts to gain insight.

\subsection{Why ILP?}
Most ML approaches rely on statistical inference.
By contrast, ILP relies on logical inference and uses techniques from \emph{automated reasoning} and \emph{knowledge representation}.
Table \ref{tab:ml-diffs} shows a simplified comparison between ILP and statistical ML approaches.
We briefly discuss these differences.

\begin{table}[ht]
\centering
\begin{tabular}{@{}l|ll@{}}
& \textbf{Statistical ML} & \textbf{ILP} \\
\midrule
\textbf{Examples} & Many & Few \\
\textbf{Data} & Tables & Logic programs \\
\textbf{Hypotheses} & Propositional/functions & First-/higher-order relations\\
\textbf{Explainability} & Difficult & Possible\\
\textbf{Knowledge transfer} & Difficult & Easy\\
\end{tabular}
\caption{A simplified comparison between ILP and statistical ML approaches based on the table by \shortciteA{cacm:ip}.}
\label{tab:ml-diffs}
\end{table}

\paragraph{Examples.}
Many forms of ML are notorious for their inability to generalise from small numbers of training examples, notably deep learning \shortcite{marcus:2018,chollet:2019,bengio:2019}.
As \shortciteA{dilp} point out, if we train a neural system to add numbers with 10 digits, it can generalise to numbers with 20 digits, but when tested on numbers with 100 digits, the predictive accuracy drastically decreases  \shortcite{nandopoo,DBLP:journals/corr/KaiserS15}.
By contrast, ILP can induce hypotheses from small numbers of examples, often from a single example \shortcite{metabias,mugg:vision}.
This data efficiency is important when we only have small amounts of training data.
For instance, \shortciteA{flashfill} applies techniques similar to ILP to induce programs from user-provided examples in Microsoft Excel to solve string transformation problems, where it is infeasible to ask a user for thousands of examples.
This data efficiency has made ILP attractive in many real-world applications, especially in drug design, where large numbers of examples are not always easy to obtain.

\paragraph{Data.}
Using logic programs to represent data allows ILP to learn with complex relational information and for easy integration of expert knowledge.
For instance, if learning causal relations in causal networks, a user can encode constraints about the network \shortcite{inoue:mla}.
If learning to recognise events, a user could provide the axioms of the event calculus \shortcite{iled}.
Relational BK allows us to succinctly represent infinite relations.
For instance, it is trivial to define a summation relation over the infinite set of natural numbers (\tw{add(A,B,C):- C = A+B}).
By contrast, tabled-based ML approaches are mostly restricted to finite data and cannot represent this information.
For instance, it is impossible to provide a decision tree learner \shortcite{id3,c45} this infinite relation because it would require an infinite feature table.
Even if we restricted ourselves to a finite set of $n$ natural numbers, a table-based approach would still need $n^3$ features to represent the complete summation relation.

\paragraph{Hypotheses.}
Because they are closely related to relational databases, logic programs naturally support relational data such as graphs.
Because of the expressivity of logic programs, ILP can learn complex relational theories, such as cellular automata \shortcite{inoue:lfit,apperception}, event calculus theories \shortcite{iled}, and Petri nets \shortcite{DBLP:journals/ml/BainS18}, and various forms of non-monotonic programs \shortcite{muggbain:nonmon,DBLP:conf/ijcai/InoueK97,sakama2001nonmonotomic}.
Because of the symbolic nature of logic programs, ILP can reason about hypotheses, which allows it to learn \emph{optimal} programs, such as minimal time-complexity programs \shortcite{metaopt} and secure access control policies \shortcite{law:fastlas}.
Moreover, because induced hypotheses have the same language as the BK, they can be stored in the BK, making transfer learning trivial \shortcite{metabias}.

\paragraph{Explainability.}
Because of logic's similarity to natural language, logic programs can be easily read by humans, which is crucial for explainable AI\footnote{\shortciteA{mugg:compmlj} (also explored by \shortciteA{mugg:beneficial}) evaluate the comprehensibility of ILP hypotheses using \shortciteS{usml} notion of \emph{ultra-strong ML}, where a learned hypothesis is expected to not only be accurate but to also demonstrably improve the performance of a human when provided with the learned hypothesis.}.
Because of this interpretability, ILP has long been used for \emph{scientific discovery}\footnote{
    \shortciteA{DBLP:journals/cacm/Muggleton99} provides a (slightly outdated) summary of scientific discovery using ILP.} \shortcite{King92,DBLP:journals/ai/SrinivasanMSK96,DBLP:conf/ilp/SrinivasanKMS97,DBLP:journals/ml/SrinivasanPCK06,Kaalia16}.
For instance, the \emph{Robot Scientist} \shortcite{king2004functional} is a system that uses ILP to generate hypotheses to explain data and can also automatically devise experiments to test the hypotheses, physically run the experiments, interpret the results, and then repeat the cycle.
Whilst researching yeast-based functional genomics, the Robot Scientist became the first machine to independently discover new scientific knowledge \shortcite{king2009automation}.

\paragraph{Knowledge transfer.}
Most ML algorithms are single-task learners and cannot reuse learned knowledge.
For instance, although AlphaGo \shortcite{alphago} has super-human Go ability, it cannot reuse this knowledge to play other games, nor the same game with a slightly different board.
By contrast, because of its symbolic representation, ILP naturally supports lifelong and transfer learning \shortcite{DBLP:conf/ilp/TorreySWM07,playgol}, which is considered essential for human-like AI \shortcite{lake:ai}.
For instance, when inducing solutions to a set of string transformation tasks, such as those in Scenario 2, \shortciteA{metabias} show that an ILP system can automatically identify easier problems to solve, learn programs for them, and then \emph{reuse} the learned programs to help learn programs for more difficult problems.
Moreover, they show that this knowledge transfer approach leads to a hierarchy of reusable programs, where each program builds on simpler programs.

\subsection{How Does ILP Work?}
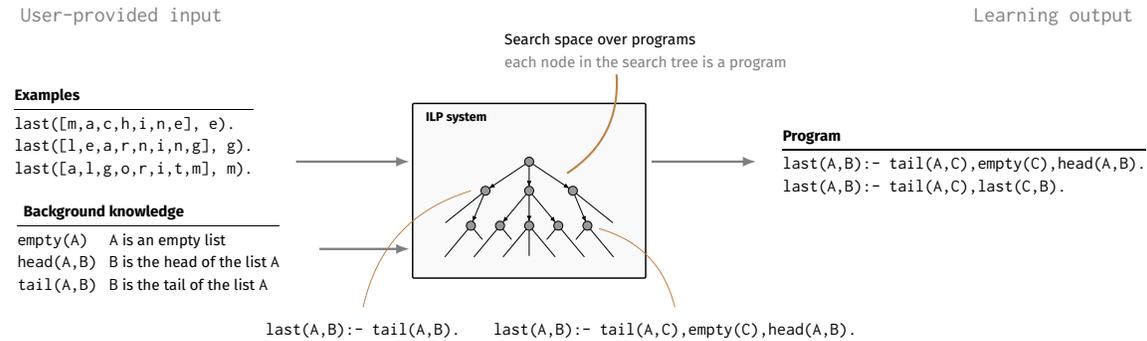
\begin{figure}[ht]
	\centering
	\resizebox{\textwidth}{!}{
	\begin{tikzpicture}[scale=1]

		\node[gray] at (-3, 4) [align=left] {\huge \tw{User-provided input}};
		\node[gray] at (29, 4) [align=left] {\huge \tw{Learning output}};

		\node at (-2.5,0) [align=left] {
			\LARGE
			\begin{tabular}{@{}l@{}}
				\textbf{Examples} \\
				\midrule
				\tw{last([m,a,c,h,i,n,e], e).} \\
				\tw{last([l,e,a,r,n,i,n,g], g).}\\
    			\tw{last([a,l,g,o,r,i,t,m], m).}\\
			\end{tabular}
		};

		\node at (-2,-4) [align=left] {
			\LARGE 
			\begin{tabular}{@{}ll@{}}
				\multicolumn{2}{l}{\textbf{Background knowledge}} \\
				\midrule
				\tw{empty(A)} & \tw{A} is an empty list \\
				\tw{head(A,B)} & \tw{B} is the head of the list \tw{A} \\
				\tw{tail(A,B)} & \tw{B} is the tail of the list \tw{A} \\
			\end{tabular}

		};

		\draw[-{Latex},gray,line width=0.8mm] (3,-1)--(6.9,-1);
		\draw[-{Latex},gray,line width=0.8mm] (3.8,-4)--(6.9,-4);

		\draw[rounded corners=0.5mm,line width=0.5mm, fill=gray, fill opacity=0.05] (7,1)--(15,1)--(15,-5)--(7,-5)--cycle;

		\node at (8.5,0.5) {\Large \textbf{ILP system}};

		\node[draw,circle,minimum size=2mm,fill=gray,fill opacity=0.8] (l1) at (11,-1) {};
		\node[draw,circle,minimum size=2mm,fill=gray,fill opacity=0.8] (l21) at (9.5,-2) {};
		\node[draw,circle,minimum size=2mm,fill=gray,fill opacity=0.8] (l22) at (11,-2) {};
		\node[draw,circle,minimum size=2mm,fill=gray,fill opacity=0.8] (l23) at (12.5,-2) {};

		\node[draw,circle,minimum size=2mm,fill=gray,fill opacity=0.8] (l31) at (11,-3.2) {};
		\node[draw,circle,minimum size=2mm,fill=gray,fill opacity=0.8] (l32) at (10,-3.2) {};
		\node[draw,circle,minimum size=2mm,fill=gray,fill opacity=0.8] (l33) at (12,-3.2) {};
		\node[draw,circle,minimum size=2mm,fill=gray,fill opacity=0.8] (l34) at (9,-3.2) {};
		\node[draw,circle,minimum size=2mm,fill=gray,fill opacity=0.8] (l35) at (13,-3.2) {};
		\node[circle,minimum size=2mm] (l36) at (8,-3.2) {};
		\node[circle,minimum size=2mm] (l37) at (14,-3.2) {};

		\node[circle,minimum size=2mm] (l41) at (14,-4.4) {};
		\node[circle,minimum size=2mm] (l42) at (13,-4.4) {};
		\node[circle,minimum size=2mm] (l43) at (12,-4.4) {};
		\node[circle,minimum size=2mm] (l44) at (11,-4.4) {};
		\node[circle,minimum size=2mm] (l45) at (10,-4.4) {};
		\node[circle,minimum size=2mm] (l46) at (9,-4.4) {};
		\node[circle,minimum size=2mm] (l47) at (8,-4.4) {};

		\draw[-{Latex}] (l1)--(l21);
		\draw[-{Latex}] (l1)--(l22);
		\draw[-{Latex}] (l1)--(l23);
		\draw[-{Latex}] (l22)--(l31);
		\draw[-{Latex}] (l22)--(l32);
		\draw[-{Latex}] (l22)--(l33);
		\draw[-{Latex}] (l21)--(l34);
		\draw[-{Latex}] (l23)--(l35);

		\draw[-] (l21)--(l36);
		\draw[-] (l23)--(l37);
		\draw[-] (l31)--(l44);
		\draw[-] (l31)--(l45);
		\draw[-] (l31)--(l43);
		\draw[-] (l32)--(l46);
		\node[circle] (t1) at (10.5,-3.8) {};
		\draw[-] (l32)--(t1);
		\draw[-] (l34)--(l47);
		\node[circle] (t2) at (9.5,-3.8) {};
		\draw[-] (l34)--(t2);
		\draw[-] (l33)--(l42);
		\node[circle] (t3) at (11.5,-3.8) {};
		\draw[-] (l33)--(t3);
		\draw[-] (l35)--(l41);
		\node[circle] (t4) at (12.5,-3.8) {};
		\draw[-] (l35)--(t4);

		\draw[-{Latex},gray,line width=0.8mm] (15.2,-1)--(18.8,-1);

		\node at (26,-1) [align=left] {
			\LARGE
			\begin{tabular}{@{}l@{}}
			\textbf{Program} \\
			\midrule
			\tw{last(A,B):- tail(A,C),empty(C),head(A,B).}\\
    		\tw{last(A,B):- tail(A,C),last(C,B).}
			\end{tabular}
		};

		\path[-,brown,line width=0.7mm,bend left] (14,2) edge (12.3,-1.4);
		\node at (15,2.8) {\LARGE
		\begin{tabular}{@{}l@{}}
			Search space over programs \\
			\textcolor{gray}{each node in the search tree is a program}
		\end{tabular}
		};

		\path[-,brown,line width=0.3mm,bend right] (9,-2) edge (5.3,-6);
		\node at (5.3, -6.8) {
			\LARGE
			\tw{last(A,B):- tail(A,B).}
		};

		\path[-,brown,line width=0.3mm,bend left] (13.4,-3.3) edge (16,-6);
		\node at (16,-6.8) {
			\LARGE
			\tw{last(A,B):- tail(A,C),empty(C),head(A,B).}
		};

	\end{tikzpicture}
	}
	\caption{An ILP system learns programs from examples and BK.
	The system learns by searching a space of possible programs which are constructed from BK.}
	\label{fig:ilp_pipe}
\end{figure}

Building an ILP system (Figure \ref{fig:ilp_pipe}) requires making several choices or assumptions.
Understanding these assumptions is key to understanding ILP.
We discuss these assumptions in Section \ref{sec:building} but briefly summarise them now.

\paragraph{Learning setting.}
The central choice is how to represent examples.
The examples in the scenarios in this section include boolean concepts (\emph{lego\_builder}) and input-output examples (string transformation and sorting).
Although boolean concepts and input-output examples are common representations, there are other representations, such as interpretations \shortcite{tilde} and transitions \shortcite{inoue:lfit}.
The representation determines the learning setting which in turn defines what it means for a program to solve the ILP problem.

\paragraph{Representation language.}
ILP represents data as logic programs.
There are, however, many logic programming languages, each with strengths and weaknesses.
For instance, Prolog is a Turing-complete logic programming language.
Datalog is a syntactical subset of Prolog that sacrifices features (such as data structures) and expressivity (it is not Turing-complete) to gain efficiency and decidability.
Some languages support non-monotonic reasoning, such as answer set programming \shortcite{asp}.
Choosing a suitable representation language is crucial in determining which problems a system can solve.

\paragraph{Defining the hypothesis space.}
The fundamental ILP problem is to search the hypothesis space for a suitable hypothesis.
The hypothesis space contains all possible programs that can be built in the chosen representation language.
Unrestricted, the hypothesis space is infinite, so it is vital to restrict it to make the search feasible.
As with all ML techniques, ILP restricts the hypothesis space by enforcing an \emph{inductive bias} \shortcite{tm:book}.
A \emph{language bias} enforces restrictions on hypotheses, such as how many variables or relations can be in a hypothesis.
Choosing an appropriate language bias is necessary for efficient learning and is a major challenge.

\paragraph{Search method.}
Having defined the hypothesis space, the problem is to efficiently search it.
The traditional way to categorise approaches is whether they use a \emph{top-down} or \emph{bottom-up} search, where \emph{generality} orders the search space\footnote{
A hypothesis $h_1$ is more general than $h_2$ if $h_1$ entails at least as many examples as $h_2$.
A hypothesis $h_1$ is more specific than $h_2$ if $h_1$ entails fewer examples then $h_2$.
}.
Top-down approaches \shortcite{foil,tilde,hyper,toplog,DBLP:conf/ilp/RibeiroI14} start with an overly general hypothesis and try to specialise it.
Bottom-up approaches \shortcite{duce,cigol,golem,inoue:lfit} start with an overly specific hypothesis and try to generalise it.
Some approaches combine the two \shortcite{progol,aleph,dcc}.
A third approach has recently emerged called \emph{meta-level} ILP \shortcite{inoue:mla,mugg:metagold,inoue:flap,law:alp,popper}.
This approach represents an ILP problem as a \emph{meta-level} logic program, i.e.~ a program that reasons about programs.
Meta-level approaches often delegate the search for a hypothesis to an off-the-shelf solver \shortcite{aspal,mugg:metalearn,ilasp,hexmil,apperception,popper} after which the meta-level solution is translated back to a standard solution for the ILP problem.

\subsection{A Brief History}

That ILP is a form of ML surprises many researchers who only associate ML with statistical techniques.
However, if we follow \shortciteS{tm:book} definition of ML\footnote{
According to \shortciteS{tm:book}, an algorithm is said to learn from experience $E$ with respect to some class of tasks $T$ and performance measure $P$, if its performance at tasks in $T$, as measured by $P$, improves with experience $E$.} then ILP is no different from other ML approaches: it improves given more examples.
The confusion seems to come from ILP's use of logic as a representation for learning.
However, as \shortciteA{pedro:master} points out, there are generally five areas of ML: symbolists, connectionists, Bayesian, analogisers, and evolutionists.
ILP is in the symbolic learning category.

Turing can be seen as one of the first symbolist, as he proposed using a logical representation to build thinking machines \shortcite{turing:mind,mugg:logic-and-learning}.
\shortciteA{jm:commonsense} made the first comprehensive proposal for the use of logic in AI with his \emph{advice seeker}.
Much work on using logic for ML soon followed.
Recognising the limitations of table-based representations, \shortciteA{banerji1964language} proposed using predicate logic as a representation language for learning.
\citeS{aq} work on the AQ algorithm, which induces rules using a set covering algorithm, has greatly influenced many ILP systems.
\shortciteS{plotkin:thesis} work on subsumption and least general generalisation has influenced nearly all of ILP, especially theory.
Other notable work includes \shortciteA{vere1975} on induction algorithms for predicate calculus and \shortciteS{marvin} MARVIN system, one of the first to learn executable programs.
\shortciteS{mis} work on inducing Prolog programs made major contributions to ILP, including the concepts of backtracking and refinement operators.
\shortciteS{foil} FOIL system is one of the most well-known ILP systems and is a natural extension of ID3 \shortcite{id3} from the propositional setting to the first-order setting.
Other notable contributions include \emph{inverse resolution} \shortcite{cigol}, which was also one of the earliest approaches at predicate invention.
ILP as a field was founded by \shortciteA{mugg:ilp}, who stated that it lies at the intersection of ML and knowledge representation.

\subsection{Contributions}

There are several excellent ILP survey papers \shortcite{sammut:history,mugg:ilp94,DBLP:journals/ai/Muggleton99,DBLP:journals/jmlr/PageS03,ilp20} and books \shortcite{ilp:book,luc:book}.
In this paper, we want to provide a new introduction to the field aimed at a general AI reader interested in symbolic learning.
We differ from existing surveys by including, and mostly focusing on, recent developments \shortcite{ilp30}, such as new methods for learning recursive programs, predicate invention, and meta-level search.
Although we cover work on inducing Datalog and answer set programs, we mostly focus on approaches that induce definite programs, and in particular Prolog programs.
We do not detail work that combines neural networks with ILP, to which there are already suitable survey papers \shortcite{DBLP:journals/flap/GarcezGLSST19,DBLP:conf/ijcai/RaedtDMM20}.

The rest of the paper is organised as follows:
\begin{itemize}
\setlength\itemsep{1pt}
\setlength\parskip{1pt}
    \item We describe necessary logic programming notation (Section \ref{sec:logic}).
    \item We define the standard ILP learning settings (Section \ref{sec:ilp}).
    \item We describe the basic assumptions required to build an ILP system (Section \ref{sec:building}).
    \item We compare many ILP systems and describe the features they support (Section \ref{sec:compare}).
    \item We describe four ILP systems in detail (Aleph, TILDE, ASPAL, and Metagol) (Section \ref{sec:cases}).
    \item We summarise some of the key application areas of ILP (Section \ref{sec:apps}).
    \item We briefly survey related work (Section \ref{sec:related}).
    \item We conclude by outlining the main current limitations of ILP and suggesting directions for future research (Section \ref{sec:cons})
\end{itemize}
\section{Logic Programming}
\label{sec:logic}

ILP uses logic programs \shortcite{kowalski:lp} to represent BK, examples, and hypotheses.
A logic program is fundamentally different from an imperative program (e.g. C, Java, Python) and very different from a functional program (e.g. Haskell, OCaml).
Imperative programming views a program as a sequence of step-by-step instructions where computation is the process of executing the instructions.
By contrast, logic programming views a program as a logical theory (a set of logical rules) where computation is various forms of deduction over the theory, such as searching for a proof, refutation, or a model of it.
Another major difference is that a logic program is \emph{declarative} \shortcite{decprog} because it allows a user to state \emph{what} a program should do, rather than \emph{how} it should work.
This declarative nature means that the order of rules in a logic program does not (usually) matter.

In the rest of this section, we introduce the basics of logic programming necessary to understand the rest of this paper.
We cover the syntax and semantics and briefly introduce different logic programming languages.
We focus on concepts necessary for understanding ILP and refer the reader to more detailed expositions of logic programming \shortcite{ilp:book,luc:book,lloyd:book}, Prolog \shortcite{artofprolog,bratko:prolog}, and ASP \cite{asp} for more information.
We, therefore, omit descriptions of many important concepts in logic programming, such as stratified negation.
Readers comfortable with logic can skip this section.

\subsection{Syntax}
\label{sec:syntax}

We first define the syntax of a logic program:

\begin{itemize}
    \setlength\itemsep{1pt}
    \setlength\parskip{0pt}
\item A \emph{variable} is a string of characters starting with an uppercase letter, e.g. $A$, $B$, and $C$.
\item A \emph{function} symbol is a string of characters starting with a lowercase letter.
\item A \emph{predicate} symbol is a string of characters starting with a lowercase letter, e.g. \tw{job} or \tw{happy}. The \emph{arity} $n$ of a function or predicate symbol $p$ is the number of arguments it takes and is denoted as $p/n$, e.g. \tw{happy/1}, \tw{head/2}, and \tw{append/3}.
\item A \emph{constant} symbol is a function symbol with zero arity, e.g. \tw{alice}  or \tw{bob}.
\item A \emph{term} is a variable, or a constant/function symbol of arity $n$ immediately followed by a tuple of $n$ terms.
\item A term is \emph{ground} if it contains no variables.
\item An \emph{atom} is a formula $p(t_1,\dots , t_n)$, where $p$ is a predicate symbol of arity $n$ and each $t_i$ is a term, e.g. \tw{lego\_builder(alice)}, where \tw{lego\_builder} is a predicate symbol of arity 1 and \tw{alice} is a constant symbol.
\item An atom is \emph{ground} if all of its terms are ground, e.g. \tw{lego\_builder(alice)} is ground but \tw{lego\_builder(A)}, where \tw{A} is a variable, is not ground.
\item The symbol \tw{not} denotes \emph{negation as failure}, where an atom is false if it cannot be proven true.
\item A \emph{literal} is an atom \tw{A} (a \emph{positive literal}) or its negation \tw{not A} (a \emph{negative literal}). For instance, \tw{lego\_builder(alice)} is both an atom and a literal but \tw{not lego\_builder(alice)} is only a literal because it includes the negation symbol \tw{not}.
\item A \emph{clause} is of the form \tw{h$_1$, $\dots$, h$_n$} :- \tw{b$_1$, $\dots$,b$_m$} where each \tw{h$_i$} and \tw{b$_j$} is a literal and the symbol \tw{,} denotes conjunction.
The symbols \tw{h$_i$} are called the \emph{head} of the clause.
The symbols \tw{b$_i$} are called the \emph{body} of the clause.
We sometimes use the name \emph{rule} instead of \emph{clause}.
\item A \emph{Horn} clause is a clause with at most one positive literal.
\item A \emph{definite} clause is clause of the form \tw{h :- b$_1$, b$_2$, $\dots$, b$_n$}, i.e. a clause with only one head literal, e.g. \tw{qsort(A,B):- empty(A),empty(B).}
Informally, a definite clause states that the head is true if the body is true, i.e.~all of the body literals are proven true.
For instance, the rule \tw{happy(A):- lego\_builder(A),enjoys\_lego(A)} says that \tw{happy(A)} is true when both \tw{lego\_builder(A)} and \tw{enjoys\_lego(A)} are true.
\item A clause is \emph{ground} if it contains no variables.
\item A clausal \emph{theory} is a set of clauses.
\item A \emph{goal} (also called a \emph{constraint}) is a clause of the form \tw{:- b$_1$, b$_2$, $\dots$, b$_n$}, i.e. a clause without a head, e.g. \tw{:- head(A,B),head(B,A)}.
\item A \emph{unit} clause is a clause with no body. For unit clauses, we usually omit the \tw{:-} symbol, e.g. \tw{loves(alice,X).}
\item A \emph{fact} is a ground unit clause \tw{loves(andrew,laura).}
\item Simultaneously replacing variables $v_1,\dots,v_n$ in a clause with terms $t_1,\dots,t_n$ is called a \emph{substitution} and is denoted as $\theta = \{v_1/t_1,\dots,v_n/t_n\}$.
For instance, applying the substitution $\theta=\{A/bob\}$ to \tw{loves(alice,A)} results in \tw{loves(alice,bob)}.
\item A substitution $\theta$ \emph{unifies} atoms $A$ and $B$ in the case $A\theta = B\theta$.
Note that atoms $A$ and $B$ need to have a distinct set of variables, i.e., they should not have a variable with the same name, for unification to work properly.
\end{itemize}


\subsection{Semantics}
\label{sec:semantics}

The semantics of logic programs is based on the concepts of a Herbrand \emph{universe}, \emph{base}, and \emph{interpretation}.
All three concepts build upon a given \emph{vocabulary} $\mathcal{V}$ containing all constants, functions, and predicate symbols of a program.
The Herbrand universe is the set of all ground terms that can be formed from the constants and functions symbols in $\mathcal{V}$.
For instance, the Herbrand universe of the lego builder example (Section \ref{ex:happy}) is
\begin{code}
	\{alice, bob, claire, dave\}
\end{code}
\noindent
If the example also contained the function symbol \texttt{age/1}, then the Herbrand universe would be the infinite set:
\begin{code}
	\{alice, bob, claire, dave, age(alice), age(bob), age(age(alice)), \dots\}
\end{code}
\noindent
The Herbrand base is the set of all ground atoms that can be formed from the predicate symbols in $\mathcal{V}$ and the terms in the corresponding Herbrand universe.
For instance, the Herbrand base of the lego builder example is:
\[
\small
\left\{
\begin{array}{l}
    \tw{happy(alice), happy(bob), happy(claire), happy(dave),}\\
    \tw{lego\_builder(alice), lego\_builder(bob), lego\_builder(claire), lego\_builder(dave),}\\
    \tw{estate\_agent(alice), estate\_agent(bob), estate\_agent(claire), estate\_agent(dave),}\\
    \tw{enjoys\_lego(alice), enjoys\_lego(bob), enjoys\_lego(claire), enjoys\_lego(dave)}
\end{array}
\right\}
\]
\noindent
A Herbrand interpretation assigns truth values to the elements of a Herbrand base.
By convention, a Herbrand interpretation includes true ground atoms, assuming that every atom not included is false.
For instance, the Herbrand interpretation corresponding to the example in Section \ref{ex:happy} is:
\[
\left\{
\begin{array}{l}
    \tw{happy(alice), lego\_builder(alice), lego\_builder(bob),  estate\_agent(claire),  }\\
    \tw{estate\_agent(dave), enjoys\_lego(alice), enjoys\_lego(claire)}
\end{array}
\right\}
\]
\noindent
A Herbrand interpretation $I$ is a Herbrand model for a set of clauses $C$ if for all clauses \tw{h$_1$;...;h$_n$:- b$_1$,..,b$_m$} $\in C$ and for all ground substitutions $\theta$: \tw{\{b$_1\theta$,...,b$_m\theta$\}} $\subset I \rightarrow$ \tw{\{h$_1\theta$,...,h$_n\theta$\}} $\cap I \neq \emptyset$.
That is, a Herbrand interpretation $I$ is a Herbrand model of a clause if for all substitutions $\theta$ for which the body literals, after applying the substitution $\theta$, are true in $I$, at least one of the head literals is also true in $I$.
For instance, the Herbrand interpretation from the previous paragraph is a model for the clause:
\begin{code}
	happy(A):- lego\_builder(A), enjoys\_lego(A).
\end{code}
\noindent
because every substitution that makes the body ($\theta$\tw{=\{A/alice\}}) true also makes the head true.
By contrast, the following interpretation is not a model of the clause because the substitution $\theta$\tw{=\{A/dave\}} makes the body true but not the head:
\begin{code}
	enjoys\_lego(A):- estate\_agent(A).
\end{code}
\noindent
\noindent
A definite clause \tw{c} is a logical consequence of a theory \tw{T} if every Herbrand model of \tw{T} is also a model of \tw{c}.

This brings us to one of the core concepts in ILP \emph{entailment}.
When a clause \tw{c} is a logical consequence of a theory \tw{T}, we say that \tw{c} is \emph{entailed} by \tw{T}, written \tw{T} $\models$ \tw{c}.
The concept of entailment will reappear throughout this text, most prominently when we discuss various learning settings in ILP (Section \ref{sec:ilp}).
It is therefore important to have a firm grasp of its meaning.


\subsection{Logic Programming Languages}
\label{sec:logprogs}

There are various logic programming languages.
We now cover some of the most important ones for ILP.
Logic programming is based on clausal logic.
Clausal programs are sets of clauses.
\shortciteA{resolution} showed that a single rule of inference (the resolution principle) is both sound and refutation complete for clausal logic.
However, reasoning about full clausal logic is computationally expensive \shortcite{ilp:book}.
Therefore, most work in ILP focuses on fragments of clausal logic, such as Horn programs: clauses with at most one positive literal.
All programs mentioned in the introduction are Horn programs, such as the program for extracting the last element of the list:
\begin{code}
	last(A,B):- tail(A,C),empty(C),head(A,B). \\
    last(A,B):- tail(A,C),last(C,B).
\end{code}

\noindent
One reason for focusing on Horn theories, rather than full clausal theories, is SLD-resolution \shortcite{kowalski:sld}, an inference rule that sacrifices expressibility for efficiency.
For instance, the clause $p(a) \lor p(b)$ cannot be expressed in Horn logic because it has two positive literals.
Horn logic is, however, still Turing complete \shortcite{tarnlund:hornclause}.
%
%

Prolog \shortcite{DBLP:journals/cacm/Kowalski88,DBLP:conf/hopl/ColmerauerR93} is a logic programming language based on SLD-resolution and is therefore restricted to Horn clauses.
Most Prolog implementations \shortcite{swipl,yap} allow extra-logical features, such as cuts.
Prolog is not purely declarative because of constructs like cut, which means that a procedural reading of a Prolog program is needed to understand it.
In other words, the order of clauses in a Prolog program has a major influence on its execution and results.

Datalog is a fragment of definite clausal theories (clausal theories that contain only definite clauses).
The main two restrictions are (i) every variable in the head literal must also appear in a body literal, and (ii) complex terms as arguments of predicates are disallowed, e.g. \tw{p(f(1),2)} or lists.
Therefore, the list manipulation programs from previous sections cannot (easily) be expressed in Datalog\footnote{It is possible to represent lists as a set of facts.}.
Datalog is, however, sufficient for the \emph{happy} Scenario because structured terms are unnecessary.
Compared to definite programs, the main advantage of Datalog is decidability \shortcite{datsin:datalog}.
However, this decidability comes at the cost of expressivity as Datalog is not Turing complete.
By contrast, definite programs with function symbols have the expressive power of Turing machines and are consequently undecidable \shortcite{tarnlund:hornclause}.
Unlike Prolog, Datalog is purely declarative.


\subsubsection{Non-monotonic Logic}

\label{sec:nlp}
A logic is monotonic when adding knowledge to it does not reduce the logical consequences of that theory.
A logic is non-monotonic if some conclusions can be removed/invalidated by adding more knowledge.
Definite programs are monotonic because anything that could be deduced before a (definite) clause is added to it can still be deduced after it is added.
In other words, adding a (definite) clause to a definite program cannot remove the logical consequences of the program.
For instance, consider the following propositional program:
\begin{code}
sunny.\\
happy:- sunny.
\end{code}
\noindent
This program states it is sunny and that I am happy if it is sunny.
We can therefore deduce that I am happy because it is sunny.
Now suppose that we added another rule:
\begin{code}
sunny.\\
happy:- sunny.\\
happy:- rich.
\end{code}
\noindent
This new rule states that I am also happy if I am rich.
Note that by the closed world assumption, we know I am not rich.
After adding this rule, we can still deduce that I am happy from the first rule.

The logic of definite clauses with negation as failure (NAF) \shortcite{naf} is non-monotonic, which brings us to the special class of \emph{normal logic programs}, which take the following form:
\begin{code}
	h :- b$_1$,...,b$_n$, not b$_{n+1}$,..., not b$_m$.
\end{code}
\noindent
Informally, a normal rule states that the head is true if all \tw{b$_1$,...,b$_n$} are true and all \tw{b$_{n+1}$,...,b$_m$} are false.
That is, an atom is false if it cannot be proven true.
That an atom cannot be proven true does not mean that it is missing from a knowledge base.
Instead, it additionally means that there no rule can prove it true.
Assuming that a logical statement is false if it cannot be proven true is known as following the \emph{closed world assumption}.

Now consider the following non-monotonic program:
\begin{code}
sunny.\\
happy:- sunny, not weekday.
\end{code}
\noindent
This program states it is sunny and I am happy if it is sunny and it is not a weekday.
By the closed world assumption, we can deduce that it is not a weekday, so we can deduce that I am happy because it is sunny and it is not a weekday.
Now suppose we added knowledge that it is a weekday.
\begin{code}
sunny.\\
happy:- sunny, not weekday.\\
weekday.
\end{code}
\noindent
Then we can no longer deduce that I am happy.
In other words, by adding knowledge that it is a weekday, the conclusion that I am happy no longer holds.

There are many different semantics ascribed to non-monotonic programs, including completion \shortcite{naf}, well-founded \shortcite{wellfounded}, and stable model (answer set) \shortcite{stablemodel} semantics.
Discussing the differences of these semantics is beyond the scope of this paper. 



\paragraph{Answer set programming.}
Answer set programming is a form of logic programming based on stable model (answer set) semantics \shortcite{stablemodel}.
Whereas a definite logic program has only one model (the least Herbrand model), an ASP program can have one, many, or even no models (answer sets).
This makes ASP particularly attractive for expressing common-sense reasoning \shortcite{law:aij}.
Similar to Datalog, an answer set program is purely declarative.
ASP also supports additional language features, such as aggregates and weak and hard constraints.
Computation in ASP is the process of finding models.
Answer set solvers perform the search and thus generate models.
Most ASP solvers \shortcite{clasp}, in principle, always terminate (unlike Prolog query evaluation, which may lead to an infinite loop).
We refer the reader to the excellent book by \shortciteA{asp} for more information.


\subsection{Generality}
\label{sec:generality}

A key concept in ILP is the generality order over hypotheses (logic programs).
A generality order helps structure the search over the hypothesis space by reasoning about the relative properties of two programs.
A generality ordering tells us whether a program \tw{p$_1$} is more general than a program \tw{p$_2$}, i.e., whether all logical consequence of \tw{p$_2$} are also logical consequences of \tw{p$_1$}.
Equivalently, if \tw{p$_1$} is \emph{more general} than \tw{p$_2$}, then \tw{p$_2$} is \emph{more specific} than \tw{p$_1$}.

Most approaches reason about the generality of programs syntactically through $\theta$-subsumption (or \emph{subsumption} for short) \shortcite{plotkin:thesis}.
To understand subsumption, we need to understand that a clause can be seen as a finite (possibly empty) set of literals, implicitly representing their disjunction.
For instance, the clause:
\begin{code}
happy(A) :- lego\_builder(A), enjoys\_lego(A).
\end{code}
\noindent
is equivalent to (where $\neg$ denotes classical negation):
\[\{\tw{happy(A)}, \neg \tw{lego\_builder(A)}, \neg \tw{enjoys\_lego(A)}\}.\]

%
%
%

\noindent
We representing clauses as sets, we can define subsumption:
\begin{definition}[Clausal subsumption]
\label{def:clausesub}
A clause $C_1$ \emph{subsumes} a clause $C_2$ if and only if there exists a substitution $\theta$ such that $C_1\theta \subseteq C_2$.
\end{definition}

\begin{example}[Clausal subsumption]
Let \tw{C$_1$} and \tw{C$_2$} be the clauses:
\[
    \begin{array}{l}
    \tw{C$_1$ = f(A,B):- head(A,B)}\\
    \tw{C$_2$ = f(X,Y):- head(X,Y),odd(Y)}.
    \end{array}
\]
Then $\tw{C}_1$ subsumes $\tw{C}_2$ because
\begin{code}
	\big\{f(A,B), $\neg$head(A,B)\big\}$\theta \subseteq$ \big\{f(X,Y), $\neg$head(X,Y),$\neg$odd(Y)\big\}
\end{code}
\noindent
 with $\theta = \{A/X,Y/B\}$.
\end{example}
\noindent
Conversely, a clause $C_2$ is \emph{more specific} than a clause $C_1$ if $C_1$ subsumes $C_2$.\footnote{This notion of subsumption is known as {\emph{weak subsumption}}. An alternative notion is \emph{strong subsumptions} which additionally performs factoring, i.e., it remove redundant literals. As an example, \tw{p(X,Y) :- q(X,Y), q(Y,X).} strongly subsumes \tw{p(Z,Z) :- q(Z,Z).} with $\theta = \{X/Z, Y/Z\}$ because \tw{p(Z,Z) :- q(Z,Z), q(Z,Z).} is equivalent to \tw{p(Z,Z) :- q(Z,Z).}. It does not, however, weakly subsume it as the number of literals is different.}

In principle, we could check the generality of two programs by comparing the consequences they entail.
However, a program might entail an infinite set of consequences (e.g. when structured terms, such as lists, are involved) which would prevent us from establishing the generality relation between two programs.
In other words, checking entailment between clauses is undecidable \shortcite{church:problem}.
By contrast, checking subsumption between clauses is decidable \shortcite{plotkin:thesis}, although, in general, deciding subsumption is a NP-complete problem \shortcite{ilp:book}.

\section{Inductive Logic Programming}
\label{sec:ilp}


In the introduction, we described four ILP scenarios.
In each case, the problem was formed of three sets $B$ (background knowledge), $E^+$ (positive examples), and $E^-$ (negative examples).
We informally stated the ILP problem is to induce a hypothesis $H$ that with $B$ \emph{generalises} $E^+$ and $E^-$.
We now formalise this problem.

According to \shortciteA{luc:settings}, there are three main ILP learning settings: learning from \emph{entailment} (LFE), \emph{interpretations} (LFI), and \emph{satisfiability} (LFS).
LFE and LFI are by far the most popular learning settings, so we only cover these two.
Other recent work focuses on \emph{learning from transitions} \shortcite{inoue:lfit,apperception,TRMLJ2020} and \emph{learning from answer sets} \shortcite{ilasp}.
We refer the reader to these other works for an overview of those learning settings.

In each setting, the symbol \mc{X} denotes the \emph{example/instance} space, the set of examples for which a concept is defined; \mc{B} denotes the language of \emph{background knowledge}, the set of all clauses that could be provided as background knowledge; and \mc{H} denotes the \emph{hypothesis space}, the set of all possible hypotheses.

\subsection{Learning From Entailment}

LFE is by far the most popular ILP setting \shortcite{mis,duce,cigol,golem,foil,progol,hyper,aleph,xhail,atom,mugg:metagold,metagol,hexmil,popper}.
The LFE problem is based on the notion of \textit{entailment}, which we discussed in Section \ref{sec:semantics}, and two properties of the hypotheses: completeness and consistency.
A hypothesis is \textit{complete} if it entails all positive examples.
A hypothesis is \textit{consistent} if it does not entail any negative example.

The LFE problem is:

\begin{definition}[\textbf{Learning from entailment}]
\label{def:lfe}
Given a tuple $(B,E^+,E^-)$ where:
\begin{itemize}
    \setlength\itemsep{1pt}
    \setlength\parskip{1pt}
\item $B \subseteq $\mc{B} denotes background knowledge
\item $E^{+} \subseteq \mathcal{X}$ denotes positive examples of the concept
\item $E^{-} \subseteq \mathcal{X}$ denotes negative examples of the concept
\end{itemize}
The goal LFE is to return a hypothesis $H \in \mathcal{H}$ such that:
\begin{itemize}
    \setlength\itemsep{1pt}
    \setlength\parskip{1pt}
    \item $\forall e \in E^+, \; H \cup B \models e$ (i.e.~$H$ is \emph{complete})
    \item $\forall e \in E^-, \; H \cup B \not\models e$ (i.e.~$H$ is \emph{consistent})
\end{itemize}
\noindent
\end{definition}


\noindent
A hypothesis can be a single clause or multiple clauses.
Often, a single clause is insufficient to describe a target concept.
For instance, to learn a definition of a recursive concept, the hypothesis needs to capture at least the base and recursive case.
The setup in which the hypothesis needs to capture at least two dependent clauses is known as multi-clause learning \shortcite{mctoplog}.


\begin{example}
Consider the LFE tuple:

\[
\tw{B} = \left\{
\begin{array}{l}
    \tw{lego\_builder(alice).}\\
    \tw{lego\_builder(bob).}\\
    \tw{estate\_agent(claire).}\\
    \tw{estate\_agent(dave).}\\
    \tw{enjoys\_lego(alice).}\\
    \tw{enjoys\_lego(claire).}
\end{array}
\right\}
\tw{E}^+ = \left\{
\begin{array}{l}
\tw{happy(alice).}
\end{array}
\right\}
\tw{E}^- = \left\{
\begin{array}{l}
    \tw{happy(bob).}\\
    \tw{happy(claire).}\\
    \tw{happy(dave).}
\end{array}
\right\}
\]

\noindent
Also assume we have the hypothesis space:

\[
\mathcal{H} = \left\{
\begin{array}{l}
    \tw{h$_1$: happy(A):- lego\_builder(A).}\\
    \tw{h$_2$: happy(A):- estate\_agent(A).}\\
    \tw{h$_3$: happy(A):- likes\_lego(A).}\\
    \tw{h$_4$: happy(A):- lego\_builder(A),estate\_agent(A).}\\
    \tw{h$_5$: happy(A):- lego\_builder(A),enjoys\_lego(A).}\\
    \tw{h$_6$: happy(A):- estate\_agent(A),enjoys\_lego(A).}
\end{array}\right\}
\]

\noindent
Then we can consider which hypotheses an ILP system should return:
\begin{itemize}
    \setlength\itemsep{1pt}
    \setlength\parskip{1pt}
\item $B \; \cup $ \tw{h}$_1 \models happy(bob)$ so is inconsistent
\item $B \; \cup $ \tw{h}$_2 \not\models happy(alice)$ so is incomplete
\item $B \; \cup $ \tw{h}$_3 \models happy(claire)$ so is inconsistent
\item $B \; \cup $ \tw{h}$_4 \not\models happy(alice)$ so is incomplete
\item $B \; \cup $ \tw{h}$_5$ is both complete and consistent
\item $B \; \cup $ \tw{h}$_6 \not\models happy(alice)$ so is incomplete
\end{itemize}
\end{example}

\noindent
The LFE problem in Definition \ref{def:lfe} is general.
ILP systems impose strong restrictions on \mc{X}, \mc{B}, and \mc{H}.
For instance, some restrict \mc{X} to only contain atoms whereas others allow clauses.
Some restrict \mc{H} to contain only Datalog clauses.
We discuss these biases in Section \ref{sec:building}.

According to Definition \ref{def:lfe}, a hypothesis must entail every positive example (be \emph{complete}) and no negative examples (be \emph{consistent}).
However, training examples are often noisy, so it is difficult to find a hypothesis that is both complete and consistent.
Therefore, most approaches relax this definition and try to find a hypothesis that entails as many positive and as few negative examples as possible.
Precisely what this means depends on the system.
For instance, the default cost function in Aleph \shortcite{aleph} is \emph{coverage}, defined as the number of positive examples entailed subtracted by the number of negative examples entailed by the hypothesis.
Other systems also consider the size of a hypothesis, typically the number of clauses or literals in it.
We discuss noise handling in Section \ref{sec:noise}.

\subsection{Learning From Interpretations}



The second most popular \shortcite{claudien,tilde,ilasp} learning setting is LFI where an example is an interpretation, i.e.~a set of facts.
The LFI problem is:

\begin{definition}[\textbf{Learning from interpretations}]
\label{def:lfi}
Given a tuple $(B,E^+,E^-)$ where:
\begin{itemize}
\setlength\itemsep{1pt}
\setlength\parskip{1pt}
\item $B \subseteq $\mc{B} denotes background knowledge
\item $E^{+} \subseteq \mathcal{X}$ denotes positive examples of the concept, each example being a set of facts
\item $E^{-} \subseteq \mathcal{X}$ denotes negative examples of the concept, each example being a set of facts
\end{itemize}
The goal of LFI is to return a hypothesis $H \in \mathcal{H}$ such that:
\begin{itemize}
    \setlength\itemsep{1pt}
    \setlength\parskip{1pt}
    \item $\forall e \in E^+, \; e \textit{ is a model of } H \cup B$
    \item $\forall e \in E^-, \; e \textit{ is not a model of } H \cup B$
\end{itemize}
\noindent
\end{definition}

\noindent
When learning from interpretations, it is implicitly assumed that every example is completely specified. 
That is, every atom in the interpretation has to be true or false, and there is no room for missing values.
As providing a complete interpretation might be unfeasible in many cases, many ILP systems focus on partial interpretations~\cite{luc:settings}.

\begin{example}
	To illustrate LFI, we use the example from \shortcite{DBLP:conf/ilp/RaedtK08}.
	Consider the BK:
	\[
	\tw{B} = \left\{
	\begin{array}{lll}
		\tw{father(henry,bill).} & \tw{father(alan,betsy).} & \tw{father(alan,benny).} \\
		\tw{mother(beth,bill).} & \tw{mother(ann,betsy).} & \tw{mother(alice,benny).}  \\
	\end{array}
	\right\}
	\]
	and the examples:
	\[
	\tw{E}^+ = \left\{
		\begin{array}{l}
			\tw{e$_1$} = \left\{
				\begin{array}{l}
					\tw{carrier(alan).} \\
					\tw{carrier(ann).} \\
					\tw{carrier(betsy).} \\
				\end{array}
				\right\} \\
			\tw{e$_2$} = \left\{
				\begin{array}{l}
					\tw{carrier(benny).} \\
					\tw{carrier(alan).} \\
					\tw{carrier(alice).}
				\end{array}
			\right\}
		\end{array}
	\right\}
	\]
	\[
	\tw{E}^- = \left\{
		\begin{array}{l}
			\tw{e$_3$} = \left\{
				\begin{array}{l}
					\tw{carrier(henry).} \\
					\tw{carrier(beth).} \\
				\end{array}
				\right\}
		\end{array}
		\right\}
	\]

	\noindent
	Also assume the hypothesis space:

	\[
		\mathcal{H} = \left\{
		\begin{array}{l}
 		   \tw{h$_1$ = carrier(X):- mother(Y,X),carrier(Y),father(Z,X),carrier(Z).}\\
 		   \tw{h$_2$ = carrier(X):- mother(Y,X),father(Z,X).}\\
		\end{array}\right\}
	\]

	\noindent
	To solve the LFI problem (Definition \ref{def:lfi}), we need to find a hypothesis $H$ such that \tw{e$_1$} and \tw{e$_2$} are models of $H \cup B$ and \tw{e$_3$} is not.
	That is, we need to find a hypothesis $H$ that satisfies the following property for every example $\tw{e}_i 
	\in \tw{E}^+$: for every substitution $\theta$ such that $body(\tw{h}_1)\theta \subseteq B \cup \tw{e}_i$ holds, it also holds that $head(\tw{h}_1)\theta \subseteq B \cup \tw{e}_i$.
	\tw{e$_3$} is not the model of \tw{h$_1$} as there exists a substitution $\theta = \{\tw{X}/\tw{bill}, \tw{Y}/\tw{beth}, \tw{Z}/\tw{henry}\}$ such that body holds but the head does not.
	For the same reason, none of the examples is a model of \tw{h$_2$}.
\end{example}

\noindent
We often say that a hypothesis \emph{covers} an example.
The meaning of a hypothesis covering an example changes depending on the learning setup.
In LFE, the hypothesis $H$ covers an example if the example is entailed by $H \cup B$. 
In LFI, the hypothesis $H$ covers an example if the example is a model of $H \cup B$.






\section{Building An ILP System}
\label{sec:building}

Building an ILP system requires making several choices or assumptions, which are part of the \emph{inductive bias} of a learner.
An inductive bias is essential and all ML approaches impose an inductive bias \shortcite{tm:book}.
Understanding these assumptions is key to understanding ILP.
The choices can be categorised as:

\begin{itemize}
    \setlength\itemsep{1pt}
    \setlength\parskip{1pt}
    \item \textbf{Learning setting}: how to represent examples
    \item \textbf{Representation language}: how to represent BK and hypotheses
    \item \textbf{Language bias}: how to define the hypothesis space
    \item \textbf{Search method}: how to search the hypothesis space
\end{itemize}

\noindent
Table \ref{tab:build} shows the assumptions of some systems.
This table excludes many important systems, including interactive systems, such as Marvin \shortcite{marvin}, MIS \shortcite{mis}, DUCE \shortcite{duce}, Cigol \shortcite{cigol}, and Clint \shortcite{clint}, theory revision systems, such as FORTE \shortcite{forte}, and probabilistic systems, such as  SLIPCOVER \shortcite{DBLP:journals/tplp/BellodiR15} and ProbFOIL \shortcite{probfoil}.
Covering all systems is beyond the scope of this paper.
We discuss these differences/assumptions.

\begin{table}[ht]
\centering
\footnotesize
\begin{minipage}{\textwidth}
\centering

\begin{tabular}{l|ccccc}

\textbf{System} & \textbf{Setting} & \textbf{Hypotheses} & \textbf{BK} & \textbf{Language Bias} & \textbf{Search method} \\

\midrule
\textbf{FOIL} \shortcite{foil}                & LFE     & Definite      & Definite\tablefootnote{
    The original FOIL setting is more restricted than the table shows and can only have BK in the form of facts and does not allow functions \shortcite{luc:book}.
} & n/a\tablefootnote{The FOIL paper does not discuss its language bias.}           & TD            \\
\midrule
\textbf{Progol} \shortcite{progol}         & LFE     & Normal        & Normal   & Modes         & BU+TD         \\
\midrule
\textbf{Claudien} \shortcite{claudien}         & LFI     & Clausal        & Definite   & DLab         & TD         \\
\midrule
\textbf{TILDE} \shortcite{tilde}               & LFI     & Logical trees & Normal   & Modes         & TD            \\
\midrule
\textbf{Aleph} \shortcite{aleph}         & LFE     & Normal        & Normal   & Modes         & BU+TD         \\
\midrule
\textbf{XHAIL} \shortcite{xhail}                & LFE     & Normal        & Normal   & Modes         & BU            \\
\midrule
\textbf{ASPAL} \shortcite{aspal}                & LFE     & Normal        & Normal   & Modes         & ML            \\
\midrule
\textbf{Atom} \shortcite{atom}                & LFE     & Normal        & Normal   & Modes         & BU+TD            \\
\midrule
\textbf{QuickFOIL} \shortcite{quickfoil}                & LFE     & Definite        & Facts   & Schema         & TD            \\
\midrule
\textbf{LFIT} \shortcite{inoue:lfit}                & LFT     & Normal        & None   & n/a\tablefootnote{
  LFIT employs many implicit language biases.
}        & BU+TD\tablefootnote{
    The LFIT approach of \shortciteA{inoue:lfit} is bottom-up and the approach of \shortciteA{DBLP:conf/ilp/RibeiroI14} it top-down.}            \\
\midrule
\textbf{ILASP\footnote{
    ILASP is a suite of ILP systems.
    For simplicity, we refer to all the systems as ILASP.
    However, some features, such as noise handling, are not in the original ILASP paper.
}} \shortcite{ilasp}                 & LFI\footnote{
    ILASP learns from answer sets.
    However, to avoid having to introduce a problem setting for a single system, we have classified it as learning from interpretations, which is the most similar setting.
    See the original ILASP paper for more information \shortcite{ilasp}.
}
     & ASP        & ASP   & Modes         & ML            \\
\midrule
\textbf{Metagol} \shortcite{mugg:metagold}              & LFE     & Definite      & Normal   & Metarules     & ML            \\
\midrule
\textbf{\dilp{}} \shortcite{dilp} & LFE     & Datalog       & Facts    & Metarules\tablefootnote{
    \dilp{} uses rule templates which can be seen as a generalisation of metarules.
}     & ML            \\
\midrule
\textbf{HEXMIL} \shortcite{hexmil}               & LFE     & Datalog      & Datalog   & Metarules  & ML\\
\midrule
\textbf{FastLAS} \shortcite{law:fastlas}                & LFI     & ASP        & ASP   & Modes         & ML            \\
\midrule
\textbf{Apperception} \shortcite{apperception}               & LFT     & Datalog\textsuperscript{$\fork$}      & None   & Types  & ML\\
\midrule
\textbf{Popper} \shortcite{popper}               & LFE     & Definite      & Normal   & Declarations  & ML\\
\end{tabular}
\caption{
    Assumptions of popular ILP systems.
    LFE stands for \emph{learn from entailment}, LFI stands for \emph{learning from interpretations}, LFT stands for \emph{learning from transitions}.
    TD stands for \emph{top-down}, BU stands for \emph{bottom-up}, and ML stands for \emph{meta-level}.
    This table is meant to provide a very high-level overview of the systems.
    Therefore, the table entries are coarse and should not be taken absolutely literally.
    For instance, Progol, Aleph, and ILASP support other types of language biases, such as constraints on clauses.
    Popper also, for instance, supports ASP programs as BK, but \emph{usually} takes normal programs.
}
\label{tab:build}
\end{minipage}
\end{table}

\subsection{Learning Setting}

The two main learning settings are LFE and LFI (Section \ref{sec:ilp}).
Within the LFE setting, there are further distinctions.
Some systems, such as Progol \shortcite{progol}, allow for clauses as examples.
Most systems, however, learn from sets of facts, so this dimension of comparison is not useful.

\subsection{Hypotheses}

Although some systems learn propositional programs, such as Duce \shortcite{duce}, most learn first-order (or higher-order) programs.
For systems that learn first-order programs, there are classes of programs that they learn.
Some systems induce full (unrestricted) clausal theories, such as Claudien \shortcite{claudien} and CF-induction \shortcite{cfinduction}.
However, reasoning about full clausal theories is computationally expensive, so most systems learn fragments of clausal logic, usually definite programs.
Systems that focus on program synthesis \shortcite{mis,hyper,atom,metagol,metaopt,popper} tend to induce definite programs, typically as Prolog programs.

\subsubsection{Normal Programs}
One motivation for learning normal programs (Section \ref{sec:nlp}) is that many practical applications require non-monotonic reasoning.
Moreover, it is often simpler to express a concept with negation as failure (NAF).
For instance, consider the following problem by \shortciteA{xhail}:

\[
\tw{B} = \left\{
\begin{array}{l}
\tw{bird(A):- penguin(A)}\\
\tw{bird(alvin)}\\
\tw{bird(betty)}\\
\tw{bird(charlie)}\\
\tw{penguin(doris)}
\end{array}
\right\}
\tw{E}^+ = \left\{
\begin{array}{l}
\tw{flies(alvin)}\\
\tw{flies(betty)}\\
\tw{flies(charlie)}
\end{array}
\right\}
\tw{E}^- = \left\{
\begin{array}{l}
    \tw{flies(doris)}
\end{array}
\right\}
\]

\noindent
Without NAF it is difficult to induce a general hypothesis for this problem.
By contrast, with NAF a system could learn the hypothesis:

\begin{hyp}
\tw{flies(A):- bird(A), not penguin(A)}
\end{hyp}

\noindent
ILP approaches that learn normal logic programs can further be characterised by their semantics, such as whether they are based on completion \shortcite{naf}, well-founded \shortcite{wellfounded}, or stable model (answer set) \shortcite{stablemodel} semantics.
Discussing the differences between these semantics is beyond the scope of this paper.

\subsubsection{Answer Set Programs}
\label{sec:asp}

There are many benefits to learning ASP programs \shortcite{otero:asp,ilasp}.
When learning Prolog programs with NAF, the programs must be stratified; otherwise, the learned program may loop under certain queries \shortcite{law:aij}.
By contrast, some systems can learn unstratified ASP programs \cite{ilasp}.
In addition, ASP programs support rules that are not available in Prolog, such as choice rules and weak and hard constraints.
For instance, ILASP \shortcite{ilasp}, can learn the following definition of a Hamiltonian graph (taken from \shortciteA{law:alp}) as an ASP program:
\begin{code}
0\{in(V0, V1)\}1 :- edge(V0, V1).\\
reach(V0):- in(1, V0).\\
reach(V1):- reach(V0), in(V0, V1).\\
:- not reach(V0), node(V0).\\
:- V1 != V2, in(V0, V2), in(V0, V1).
\end{code}
\noindent
This program illustrates useful language features of ASP.
The first rule is a \emph{choice} rule, which means that an atom \emph{can} be true.
In this example, the rule indicates that there can be an in edge from the vertex \emph{V1} to \emph{V0}.
The last two rules are \emph{hard constraints}, which essentially enforce integrity constraints.
The first hard constraint states that it is impossible to have a node that is not reachable.
The second hard constraint states that it is impossible to have a vertex with two in edges from distinct nodes.
For more information about ASP we recommend the book by \shortciteA{asp}.

Approaches to learning ASP programs can be divided into two categories: \emph{brave learners}, which aim to learn a program such that at least one answer set covers the examples, and \emph{cautious learners}, which aim to find a program which covers the examples in all answer sets.
We refer to existing work of \shortciteA{otero:asp,DBLP:journals/ml/SakamaI09,DBLP:journals/ml/SakamaI09,law:aij} for more information about these different approaches.

\subsubsection{Higher-order Programs}
\label{sec:ho}
As many programmers know, there are benefits to using higher-order representations.
For instance, suppose you have some encrypted/decrypted strings represented as Prolog facts:

\begin{epos}
\tw{decrypt([d,b,u],[c,a,t])}\\
\tw{decrypt([e,p,h],[d,o,g])}\\
\tw{decrypt([h,p,p,t,f],[g,o,o,s,e])}
\end{epos}

\noindent
Given these examples and suitable BK, a system could learn the first-order program:

\begin{hyp}
\tw{decrypt(A,B):- empty(A),empty(B).}\\
\tw{decrypt(A,B):- head(A,C),chartoint(C,D),prec(D,E),inttochar(E,F),}\\
    \hspace{29mm} \tw{head(B,F),tail(A,G),tail(B,H),decrypt(G,H).}
\end{hyp}

\noindent
This program defines a Caesar cypher which shifts each character back once (e.g. \emph{z} $\mapsto$ \emph{y}, \emph{y} $\mapsto$ \emph{x}, etc).
Although correct (ignoring the modulo operation for simplicity), this program is long and difficult to read.
To overcome this limitation, some systems \shortcite{metaho} learn higher-order programs, such as:

\begin{hyp}
\tw{decrypt(A,B):- map(A,B,inv)}\\
\tw{inv(A,B):- char\_to\_int(A,C),prec(C,D),int\_to\_char(D,B)}
\end{hyp}

\noindent
This program is higher-order because it allows literals to take predicate symbols as arguments.
The symbol \tw{inv} is \emph{invented} (we discuss \emph{predicate invention} in Section \ref{sec:pi}) and is used as an argument for \tw{map} in the first rule and as a predicate symbol in the second rule.
The higher-order program is smaller than the first-order program because the higher-order background relation \tw{map} abstracts away the need to learn a recursive program.
\shortciteA{metaho} show that inducing higher-order programs can drastically improve learning performance in terms of predictive accuracy, sample complexity, and learning times.

\subsection{Background Knowledge}
\label{sec:bk}

BK is similar to features used in other forms of ML.
However, whereas features are finite tables, BK is a logic program.
Using logic programs to represent data allows ILP to learn with complex relational information.
For instance, suppose we want to learn list or string transformation programs, we might want to supply helper relations, such as \tw{head}, \tw{tail}, and \tw{last} as BK:

\begin{bk}
\tw{head([H|\_],H).}\\
\tw{tail([\_|T],T).}\\
\tw{last([H],H).}\\
\tw{last([\_|T1],A):- tail(T1,T2),last(T2,A).}\\
\end{bk}
\noindent
These relations hold for lists of any length and any type.

As a second example, suppose you want to learn the definition of a prime number.
Then you might want to give a system the ability to perform arithmetic reasoning, such as using the Prolog relations:

\begin{bk}
\tw{even(A):- 0 is mod(A,2).}\\
\tw{odd(A):- 1 is mod(A,2).}\\
\tw{sum(A,B,C):- C is A+B.}\\
\tw{gt(A,B):- A>B.}\\
\tw{lt(A,B):- A<B.}
\end{bk}

\noindent
These relations are general and hold for arbitrary numbers and we do not need to pre-compute all the logical consequences of the definitions, which is impossible because there are infinitely many.
By contrast, table-based ML approaches are restricted to finite propositional data.
For instance, it is impossible to use the greater than relation over the set of natural numbers in a decision tree learner because it would require an infinite feature table.

\subsubsection{Constraints}
BK allows a human to encode prior knowledge of a problem.
As a trivial example, if learning banking rules to determine whether two companies can lend to each other, you may encode a prior constraint to prevent two companies from lending to each other if they are owned by the same parent company:

\begin{code}

:- lend(A,B), parent\_company(A,C), parent\_company(B,C).
\end{code}

\noindent
Constraints are widely used in ILP \shortcite{quickfoil,evans:thesis,popper}.
For instance, \shortciteA{inoue:mla} represent knowledge as a causal graph and use constraints to denote impossible connections between nodes.
\shortciteA{apperception} use constraints to induce theories to explain sensory sequences.
For instance, one requirement of their \emph{unity condition} is that objects (constants) are connected via chains of binary relations.
The authors argue that such constraints are necessary for the induced solutions to achieve good predictive accuracy.



\subsubsection{Discussion}
\label{sec:bk:discuss}

As with choosing appropriate features, choosing appropriate BK in ILP is crucial for good learning performance.
ILP has traditionally relied on predefined and hand-crafted BK, often designed by domain experts.
However, it is often difficult and expensive to obtain such BK.
Indeed, the over-reliance on hand-crafted BK is a common criticism of ILP \shortcite{dilp}.
The difficulty is finding the balance of having enough BK to solve a problem, but not too much that a system becomes overwhelmed.
We discuss these two issues.

\paragraph{Too little BK.}
If we use too little or insufficient BK then we may exclude a good hypothesis from the hypothesis space.
For instance, reconsider the string transformation problem from the introduction, where we want to learn a program that returns the last character of a string from examples.

\begin{epos}
\tw{last([m,a,c,h,i,n,e], e)}\\
\tw{last([l,e,a,r,n,i,n,g],g)}\\
\tw{last([a,l,g,o,r,i,t,m], m)}
\end{epos}

\noindent
To induce a hypothesis from these examples, we need to provide an ILP system with suitable BK.
For instance, we might provide BK that contains relations for common list/string operations, such as \tw{empty}, \tw{head}, and \tw{tail}.
Given these three relations, an ILP system could learn the program:

\begin{hyp}
\tw{last(A,B):- tail(A,C),empty(C),head(A,B).}\\
\tw{last(A,B):- tail(A,C),last(C,B).}
\end{hyp}

\noindent
However, suppose that the user had not provided \tw{tail} as BK.
Then how could a system learn the above hypothesis?
This situation is a major problem, as most systems can only use BK provided by a user.
To mitigate this issue, there is research on enabling a system to automatically \emph{invent} new predicate symbols, known as \emph{predicate invention}, which we discuss in Section \ref{sec:pi}, which has been shown to mitigate missing BK \shortcite{crop:incomplete}.
However, ILP still heavily relies on much human input to solve a problem.
Addressing this limitation is a major challenge.

\paragraph{Too much BK.}
As with too little BK, a major challenge is too much irrelevant BK.
Too many relations (assuming that they can appear in a hypothesis) is often a problem because the size of the hypothesis space is a function of the size of the BK.
Empirically, too much irrelevant BK is detrimental to learning performance \shortcite{ashwin:1995,ashwin:badbk,forgetgol}, this also includes irrelevant language biases \shortcite{reduce}.
Addressing the problem of too much BK has been under-researched.
In Section \ref{sec:cons}, we suggest that this topic is a promising direction for future work, especially when considering the potential for ILP to be used for lifelong learning (Section \ref{sec:lifelong}).

\subsection{Language Bias}
\label{sec:bias}

The fundamental ILP problem is to search the hypothesis space for a suitable hypothesis.
The hypothesis space contains all possible programs that can be built in the chosen representation language.
Unrestricted, the hypothesis space is infinite, so it is important to restrict it to make the search feasible.
To restrict the hypothesis space, systems enforce an \emph{inductive bias} \shortcite{tm:book}.
A \emph{language bias} enforces restrictions on hypotheses, such as restricting the number of variables, literals, and rules in a hypothesis.
These restrictions can be categorised as either \emph{syntactic} bias, restrictions on the form of a rule in a hypothesis, and \emph{semantic} bias, restrictions on the behaviour of induced hypotheses \shortcite{decbias}.
For instance, in the happy example (Example \ref{ex:happy}), we assumed that a hypothesis only contains predicate symbols that appear in the BK or examples.
However, we need to encode this bias to give an ILP system.
There are several ways of encoding a language bias, such as \emph{grammars} \shortcite{cohen:grammarbias}, Dlabs \shortcite{claudien}, production fields \shortcite{cfinduction}, and predicate declarations \shortcite{popper}.
We focus on \emph{mode declarations} \shortcite{progol} and \emph{metarules} \shortcite{reduce}, two popular language biases.

\subsubsection{Mode Declarations}
\label{sec:modes}
Mode declarations are the most popular form of language bias \shortcite{progol,tilde,aleph,xhail,tal,aspal,raspal,atom,ilasp,iled}.
Mode declarations state which predicate symbols may appear in a rule, how often, and also their argument types.
In the mode language, \emph{modeh} declarations denote which literals may appear in the head of a rule and \emph{modeb} declarations denote which literals may appear in the body of a rule.
A mode declaration is of the form:
\[mode(recall,pred(m_1,m_2,\dots,m_a))\]
The following are all valid mode declarations:

\begin{code}
modeh(1,happy(+person)).\\
modeb(*,member(+list,-element)).\\
modeb(1,head(+list,-element)).\\
modeb(2,parent(+person,-person)).\\
\end{code}

\noindent
The first argument of a mode declaration is an integer denoting the \emph{recall}.
Recall is the maximum number of times that a mode declaration can be used in a rule\footnote{
    The recall of a \emph{modeh} declaration is almost always useless and is often set to 1.}.
Another way of understanding recall is that it bounds the number of alternative solutions for a literal.
Providing a recall is a hint to a system to ignore certain hypotheses.
For instance, if using the \emph{parent} kinship relation, then we can set the recall to two, as a person has at most two parents. If using the \emph{grandparent} relation, then we can set the recall to four, as a person has at most four grandparents.
If we know that a relation is functional, such as \tw{head}, then we can bound the recall to one.
The symbol \emph{*} denotes no bound.

The second argument denotes that the predicate symbol that may appear in the head (\emph{modeh}) or body (\emph{modeb}) of a rule and the type of arguments it takes.
The symbols $+$, $-$, and $\#$ denote whether the arguments are \emph{input}, \emph{output}, or \emph{ground} arguments respectively.
An \emph{input} argument specifies that, at the time of calling the literal, the corresponding argument must be instantiated.
In other words, the argument needs to be bound to a variable that already appears in the rule.
An \emph{output} argument specifies that the argument should be bound after calling the corresponding literal.
A \emph{ground} argument specifies that the argument should be ground and is often used to learn rules with constant symbols in them.

To illustrate mode declarations, consider the modes:

\begin{code}
modeh(1,target(+list,-char)).\\
modeb(*,head(+list,-char)).\\
modeb(*,tail(+list,-list)).\\
modeb(1,member(+list,-list)).\\
modeb(1,equal(+char,-char)).\\
modeb(*,empty(+list)).
\end{code}

\noindent
Given these modes, the rule \tw{target(A,B):- head(A,C),tail(C,B)} is mode inconsistent because \tw{modeh(1,target(+list,-char))} requires that the second argument of target (\tw{B}) is char and the mode \tw{modeb(*,tail(+list,-list))} requires that the second argument of tail (\tw{B}) is a list, so this rule is mode inconsistent.
The rule \tw{target(A,B):- empty(A),head(C,B)} is also mode inconsistent because \tw{modeb(*,head(+list,-char))} requires that the first argument of head (\tw{C}) is instantiated but the variable \tw{C} is never instantiated in the rule.

By contrast, the following rules are all mode consistent:

\begin{code}
target(A,B):- tail(A,C),head(C,B).\\
target(A,B):- tail(A,C),tail(C,D),equal(C,D),head(A,B).\\
target(A,B):- tail(A,C),member(C,B).
\end{code}

\noindent
Depending on the specific system, modes can also support the introduction of constant symbols.
In Aleph, an example of such a declaration is \tw{modeb(*,length(+list,\#int))}, which would allow integer values to be included in rules.

Different systems use mode declarations in slightly different ways.
Progol and Aleph use mode declarations with input/output argument types because they induce Prolog programs, where the order of literals in a rule matters.
By contrast, ILASP induces ASP programs, where the order of literals in a rule does not matter, so ILASP does not use input/output arguments.

\subsubsection{Metarules}
\label{sec:metarules}
\emph{Metarules}\footnote{Metarules were introduced as \emph{clause schemata} by \shortciteA{emde:metarules} and were notably used in Mobal system \shortcite{mobal}. Metarules are also called \emph{second-order schemata} \shortcite{clint} and \emph{program schemata} \shortcite{dialogs}, amongst many other names.} are a popular form of syntactic bias and are used by many systems \shortcite{clint,dialogs,mobal,wang2014structure,mugg:metagold,metagol,hexmil,dilp,DBLP:journals/ml/BainS18}.
Metarules are second-order rules which define the structure of learnable programs which in turn defines the hypothesis space.
For instance, to learn the \tw{grandparent} relation given the \tw{parent} relation, the \emph{chain} metarule would be suitable:

\begin{code}
P(A,B):- Q(A,C), R(C,B).
\end{code}

\noindent
The letters $P$, $Q$, and $R$ denote second-order variables (variables that can be bound to predicate symbols) and the letters $A$, $B$ and $C$ denote first-order variables (variables that can be bound to constant symbols). Given the \emph{chain} metarule, the background \tw{parent} relation, and examples of the \tw{grandparent} relation, ILP approaches will try to find suitable substitutions for the second-order variables, such as the substitutions \{P/grandparent, Q/parent, R/parent\} to induce the theory:

\begin{code}
grandparent(A,B):- parent(A,C),parent(C,B).
\end{code}

\noindent
Despite their widespread use, there is little work determining which metarules to use for a given learning task.
Instead, these approaches assume suitable metarules as input or use metarules without any theoretical guarantees.
In contrast to other forms of bias in ILP, such as modes or grammars, metarules are themselves logical statements, which allows us to reason about them.
For this reason, there is preliminary work in reasoning about metarules to identify universal sets suitable to learn certain fragments of logic programs \shortcite{minmeta,reduce-jelia,reduce}.
Despite this preliminary work, deciding which metarules to use for a given problem is still a major challenge, which future work must address.

\subsubsection{Discussion}
Choosing an appropriate language bias is essential to make an ILP problem tractable because it defines the hypothesis space.
If the bias is too \emph{weak}, then the search can become intractable.
If the bias is too  \emph{strong} then we risk excluding a good solution from the hypothesis space.
This trade-off is one of the major problems holding ILP back from being widely used\footnote{
    The Blumer bound \shortcite{blumer:bound} (the bound is a reformulation of Lemma 2.1) helps explain this trade-off.
This bound states that given two hypothesis spaces, searching the smaller space will result in fewer errors compared to the larger space, assuming that the target hypothesis is in both spaces.
Here lies the problem: how to choose a learner's hypothesis space so that it is large enough to contain the target hypothesis yet small enough to be efficiently searched.
To know more about this aspect of ILP, we first recommend Chapter 7 of \shortciteS{tm:book} Machine Learning book, which is, in our view, still the best introductory exposition of computational learning theory, and then work specific to ILP \shortcite{cohen:pac1,cohen:pac2,cohen:pac3,DBLP:conf/ilp/GottlobLS97}.
}.
To understand the impact of an inappropriate language bias, consider the string transformation example in Section \ref{sec:intro:string}.
Even if all necessary background relations are provided, not providing a recursive metarule (e.g. \tw{R(A,B):- P(A,C), R(C,B)}) would prevent a metarule-based system from inducing a program that generalises to lists of any length.
Similarly, not providing a recursive mode declaration for the target relation would prevent a mode-based system from finding a good hypothesis.

Different language biases offer different benefits.
Mode declarations are expressive enough to enforce a strong bias to significantly prune the hypothesis space.
They are especially appropriate when a user has much knowledge about their data and can, for instance, determine suitable recall values.
If a user does not have such knowledge, then it can be very difficult to determine suitable mode declarations.
Moreover, if a user provides weak mode declarations (for instance with infinite recall, a single type, and no input/output arguments), then the search quickly becomes intractable.
Although there is some work on learning mode declarations \shortcite{modelearning,DBLP:conf/ilp/FerilliEBM04,DBLP:conf/sigmod/PicadoTFP17}, it is still a major challenge to choose appropriate ones.

A benefit of metarules is that they require little knowledge of the BK and a user does not need to provide recall values, types, or specify input/output arguments.
Because they precisely define the form of hypotheses, they can greatly reduce the hypothesis space, especially if the user knows about the class of programs to be learned.
However, as previously mentioned, the major downside with metarules is determining which metarules to use for an arbitrary learning task.
Although there is some preliminary work in identifying universal sets of metarules \shortcite{minmeta,reduce-jelia,reduce}, deciding which metarules to use for a given problem is a major challenge, which future work must address.
\subsection{Search Method}
\label{sec:method}

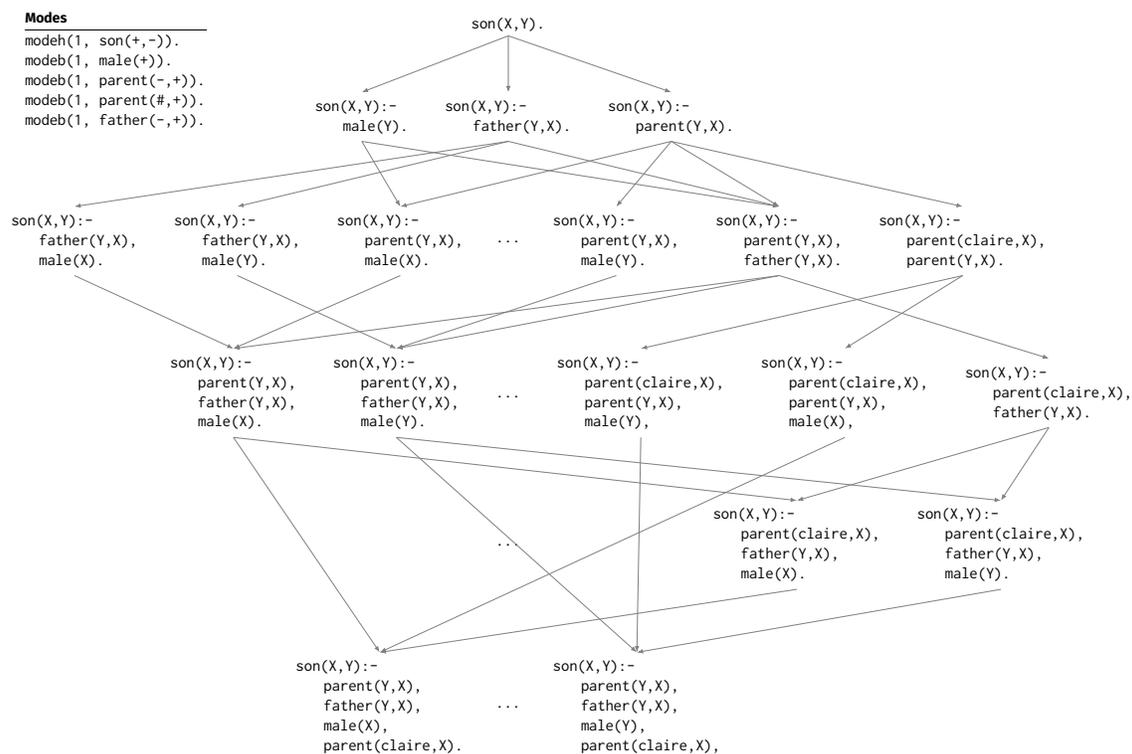
\begin{figure}[t]
	\centering
	\Huge
	\resizebox{\textwidth}{!}{
	\begin{tikzpicture}[scale=1]

		{
		\Huge
		\node (examples) at (0,-0.5) [align=left] {
			\begin{tabular}{@{}l@{}}
				\textbf{Modes} \\
				\midrule
				\tw{modeh(1, son(+,-)).} \\
				\tw{modeb(1, male(+)).} \\
				\tw{modeb(1, parent(-,+)).} \\
				\tw{modeb(1, parent(\#,+)).} \\
				\tw{modeb(1, father(-,+)).} \\
			\end{tabular}
		};
		}
		
		\node (l1_1) at (13,-3) {
			\Huge
			\begin{tabular}{l}
				\tw{son(X,Y):-} \\
				\quad\quad \tw{male(Y).}
			\end{tabular}
		};

		\node[right=of l1_1] (l1_2) {
			\Huge
			\begin{tabular}{l}
				\tw{son(X,Y):-} \\
				\quad\quad \tw{father(Y,X).}
			\end{tabular}
		};

		\node[right=of l1_2] (l1_3) {
			\Huge
			\begin{tabular}{l}
				\tw{son(X,Y):-} \\
				\quad\quad \tw{parent(Y,X).}
			\end{tabular}
		};

%
		
		\node[above=3cm of l1_2] (top) {\tw{son(X,Y).}};
		
		\draw[-{Latex},gray, line width=0.05cm] (top.south)--(l1_1.north);
		\draw[-{Latex},gray, line width=0.05cm] (top.south)--(l1_2.north);
		\draw[-{Latex},gray, line width=0.05cm] (top.south)--(l1_3.north);
		

		\node[below=5cm of l1_2] (l2_center) {
			\tw{...}
		};

		\node[left=of l2_center] (l2_1) {
			\Huge
			\begin{tabular}{l}
				\tw{son(X,Y):-} \\
				\quad\quad \tw{parent(Y,X),} \\
				\quad\quad \tw{male(X).}
			\end{tabular}
		};

		\node[left=of l2_1] (l2_3) {
			\Huge
			\begin{tabular}{l}
				\tw{son(X,Y):-} \\
				\quad\quad \tw{father(Y,X),} \\
				\quad\quad \tw{male(Y).}
			\end{tabular}
		};

		\node[left=of l2_3] (l2_5) {
			\Huge
			\begin{tabular}{l}
				\tw{son(X,Y):-} \\
				\quad\quad \tw{father(Y,X),} \\
				\quad\quad \tw{male(X).}
			\end{tabular}
		};

		\node[right=of l2_center] (l2_2) {
			\Huge
			\begin{tabular}{l}
				\tw{son(X,Y):-} \\
				\quad\quad \tw{parent(Y,X),} \\
				\quad\quad \tw{male(Y).}
			\end{tabular}
		};

		\node[right=of l2_2] (l2_4) {
			\Huge
			\begin{tabular}{l}
				\tw{son(X,Y):-} \\
				\quad\quad \tw{parent(Y,X),} \\
				\quad\quad \tw{father(Y,X).}
			\end{tabular}
		};
		
		\node[right=of l2_4] (l2_6) {
			\Huge
			\begin{tabular}{l}
				\tw{son(X,Y):-} \\
				\quad\quad \tw{parent(claire,X),} \\
				\quad\quad \tw{parent(Y,X).} \\
			\end{tabular}
		};
		
		\draw[-{Latex},gray, line width=0.05cm] (l1_1.south)--(l2_1.north);
		\draw[-{Latex},gray, line width=0.05cm] (l1_1.south)--(l2_4.north);
		\draw[-{Latex},gray, line width=0.05cm] (l1_2.south)--(l2_3.north);
		\draw[-{Latex},gray, line width=0.05cm] (l1_2.south)--(l2_5.north);
		\draw[-{Latex},gray, line width=0.05cm] (l1_2.south)--(l2_4.north);
		\draw[-{Latex},gray, line width=0.05cm] (l1_3.south)--(l2_2.north);
		\draw[-{Latex},gray, line width=0.05cm] (l1_3.south)--(l2_4.north);
		\draw[-{Latex},gray, line width=0.05cm] (l1_3.south)--(l2_6.north);
		\draw[-{Latex},gray, line width=0.05cm] (l1_3.south)--(l2_1.north);

		\node[below=7cm of l2_center] (l3_center) {
			\Huge
			\begin{tabular}{l}
				\tw{...}
			\end{tabular}
		};

		\node[left=of l3_center] (l3_2) {
			\Huge
			\begin{tabular}{l}
				\tw{son(X,Y):-} \\
				\quad\quad \tw{parent(Y,X),} \\
				\quad\quad \tw{father(Y,X),} \\
				\quad\quad \tw{male(Y).}
			\end{tabular}
		};

		\node[right=of l3_center] (l3_6) {
			\Huge
			\begin{tabular}{l}
				\tw{son(X,Y):-} \\
				\quad\quad \tw{parent(claire,X),} \\
				\quad\quad \tw{parent(Y,X),} \\
				\quad\quad \tw{male(Y),} \\
			\end{tabular}
		};
		
		\node[right=of l3_6] (l3_8) {
			\Huge
			\begin{tabular}{l}
				\tw{son(X,Y):-} \\
				\quad\quad \tw{parent(claire,X),} \\
				\quad\quad \tw{parent(Y,X),} \\
				\quad\quad \tw{male(X),} \\
			\end{tabular}
		};

		\node[right=of l3_8] (l3_4) {
			\Huge
			\begin{tabular}{l}
				\tw{son(X,Y):-} \\
				\quad\quad \tw{parent(claire,X),} \\
				\quad\quad \tw{father(Y,X).} \\
			\end{tabular}
		};

		\node[left=of l3_2] (l3_1) {
			\Huge
			\begin{tabular}{l}
				\tw{son(X,Y):-} \\
				\quad\quad \tw{parent(Y,X),} \\
				\quad\quad \tw{father(Y,X),} \\
				\quad\quad \tw{male(X).}
			\end{tabular}
		};

		\draw[-{Latex},gray,  line width=0.05cm] (l2_5.south)--(l3_1.north);
		\draw[-{Latex},gray,  line width=0.05cm] (l2_3.south)--(l3_2.north);
		\draw[-{Latex},gray,  line width=0.05cm] (l2_1.south)--(l3_1.north);
		\draw[-{Latex},gray,  line width=0.05cm] (l2_2.south)--(l3_2.north);
		\draw[-{Latex},gray,  line width=0.05cm] (l2_4.south)--(l3_2.north);
		\draw[-{Latex},gray,  line width=0.05cm] (l2_4.south)--(l3_1.north);
		\draw[-{Latex},gray,  line width=0.05cm] (l2_4.south)--(l3_4.north);
		\draw[-{Latex},gray,  line width=0.05cm] (l2_6.south)--(l3_6.north);
		\draw[-{Latex},gray,  line width=0.05cm] (l2_6.south)--(l3_8.north);
		
		\node[below=7cm of l3_center] (lm_center) {\Huge \tw{...}};
		
		\node[right=8cm of lm_center] (lm1_center) {};

		\node[right=of lm1_center] (lm_1) {
			\Huge
			\begin{tabular}{l}
				\tw{son(X,Y):-} \\
				\quad\quad \tw{parent(claire,X),} \\
				\quad\quad \tw{father(Y,X),} \\
				\quad\quad \tw{male(X).}
			\end{tabular}	
		};

		\node[right=of lm_1] (lm_2) {
			\Huge
			\begin{tabular}{l}
				\tw{son(X,Y):-} \\
				\quad\quad \tw{parent(claire,X),} \\
				\quad\quad \tw{father(Y,X),} \\
				\quad\quad \tw{male(Y).}
			\end{tabular}	
		};

		\node[below=8cm of lm_center] (l4_center) {
			\Huge
			\tw{...}
		};
		
		\draw[-{Latex},gray,  line width=0.05cm] (l3_4.south)--(lm_1.north);
		\draw[-{Latex},gray,  line width=0.05cm] (l3_4.south)--(lm_2.north);
		\draw[-{Latex},gray,  line width=0.05cm] (l3_2.south)--(lm_2.north);
		\draw[-{Latex},gray,  line width=0.05cm] (l3_1.south)--(lm_1.north);

		\node[left=of l4_center] (l4_1) {
			\Huge
			\begin{tabular}{l}
				\tw{son(X,Y):-} \\
				\quad\quad \tw{parent(Y,X),} \\
				\quad\quad \tw{father(Y,X),} \\
				\quad\quad \tw{male(X),} \\
				\quad\quad \tw{parent(claire,X).}\\
			\end{tabular}	
		};

		\node[right=of l4_center] (l4_2) {
			\Huge
			\begin{tabular}{l}
				\tw{son(X,Y):-} \\
				\quad\quad \tw{parent(Y,X),} \\
				\quad\quad \tw{father(Y,X),} \\
				\quad\quad \tw{male(Y),} \\
				\quad\quad \tw{parent(claire,X),} 
			\end{tabular}	
		};

		\draw[-{Latex},gray,  line width=0.05cm] (l3_2.south)--(l4_2.north);
		\draw[-{Latex},gray,  line width=0.05cm] (l3_1.south)--(l4_1.north);
		\draw[-{Latex},gray,  line width=0.05cm] (l3_6.south)--(l4_2.north);
		\draw[-{Latex},gray,  line width=0.05cm] (l3_8.south)--(l4_1.north);
		\draw[-{Latex},gray,  line width=0.05cm] (lm_1.south)--(l4_1.north);
		\draw[-{Latex},gray,  line width=0.05cm] (lm_2.south)--(l4_2.north);

	\end{tikzpicture}
	}
	\caption{The generality relation orders the hypothesis space into a lattice (an arrow connects a hypothesis with its specialisation). The hypothesis space is built from the modes and only shown partially (\tw{\#} indicates that a constant needs to be used as an argument; only \tw{claire} is used as a constant here). The most general hypothesis sits on the top of the lattice, while the most specific hypotheses are at the bottom. The \textit{top-down} lattice traversal starts at the top, with the most general hypothesis, and specialises it moving downwards through the lattice. The \textit{bottom-up} traversal starts at the bottom, with the most specific  hypothesis, and generalises it moving upwards through the lattice.}
	\label{fig:lattice}
\end{figure}

Having defined the hypothesis space, the next problem is to efficiently search it.
There are two traditional search methods: \emph{bottom-up} and \emph{top-down}.
These methods rely on notions of generality, where one program is more \emph{general} or more \emph{specific} than another (Section \ref{sec:generality}).
A generality relation imposes an order over the hypothesis space.
Figure \ref{fig:lattice} shows this order using theta-subsumption, the most popular ordering relation.
A system can exploit this ordering during the search for a hypothesis.
For instance, if a clause does not entail a positive example, then there is no need to explore any of its specialisations because it is logically impossible for them to entail the example.
Likewise, if a clause entails a negative example, then there is no need to explore any of its generalisations because they will also entail the example.

The above paragraph refers only to generality orders on single clauses, as many systems employ the covering algorithm whereby hypotheses are constructed iteratively clause by clause \cite{foil,claudien,progol,tilde,aleph}.
However, some systems \cite{mis,hyper,popper} induce theories formed of multiple clauses and thus require generality orders over clausal \emph{theories}.
We refer an interested reader to Chapter 7 in the work of \shortciteA{luc:book} for more information about inducing theories.

\subsubsection{Top-down}
Top-down algorithms, \shortcite{foil,tilde,hyper,toplog} start with a general hypothesis and then specialise it.
For instance, HYPER \shortcite{hyper} searches a tree in which the nodes correspond to hypotheses.
Each child of a hypothesis in the tree is more specific than or equal to its predecessor in terms of theta-subsumption, i.e.~a hypothesis can only entail a subset of the examples entailed by its parent.
The construction of hypotheses is based on \emph{hypothesis refinement} \shortcite{mis,ilp:book}.
If a hypothesis is considered that does not entail all the positive examples, it is immediately discarded because it can never be refined into a complete hypothesis.

\subsubsection{Bottom-up}
\label{search:bottom-up}
Bottom-up algorithms start with the examples and generalise them \shortcite{duce,cigol,golem,progolem,inoue:lfit}.
For instance, Golem \shortcite{golem} generalises pairs of examples based on relative least-general generalisation (RLGG) \shortcite{rlgg}.
To introduce RLGG, we start by introducing \shortciteS{plotkin:thesis} notion of least-general generalisation (LGG), which tells us how to generalise two clauses.
Given two clauses, the LGG operator returns the most specific single clause that is more general than both of them.

To define the LGG of two clauses, we start with the LGG of terms:
\begin{itemize}
    \setlength\itemsep{1pt}
    \setlength\parskip{1pt}
    \item \tw{lgg(f(s$_1$,$\ldots$,s$_n$),  f(t$_1$,$\ldots$,t$_m$))} = \tw{f(lgg(s$_1$,t$_1$),$\dots$,lgg(s$_n$,t$_n$))}.
    \item \tw{lgg(f(s$_1$,$\ldots$,s$_n$),  g(t$_1$,$\ldots$,t$_m$))} = \tw{V} (a variable),
    \item \tw{lgg(f(s$_1$,$\ldots$,s$_n$),  V)} = \tw{V$^\prime$} (a new variable).
\end{itemize}
\noindent Note that a constant is a functor with zero arguments, and thus the above rules apply.
We define the LGG of literals:
\begin{itemize}
    \setlength\itemsep{1pt}
    \setlength\parskip{1pt}
    \item \tw{lgg(p(s$_1$,$\ldots$,s$_n$),  p(t$_1$,$\ldots$,t$_n$))} = \tw{p(lgg(s$_1$,t$_1$),$\dots$,lgg(s$_n$,t$_n$))}.
    \item \tw{lgg($\neg$p(s$_1$,$\ldots$,s$_n$),  $\neg$p(t$_1$,$\ldots$,t$_n$))} = \tw{$\neg$p(lgg(s$_1$,t$_1$),$\dots$,lgg(s$_n$,t$_n$))}
    \item \tw{lgg(p(s$_1$,$\ldots$,s$_n$),  q(t$_1$,$\ldots$,t$_n$))} is undefined
    \item \tw{lgg(p(s$_1$,$\ldots$,s$_n$),  $\neg$p(t$_1$,$\ldots$,t$_n$))} is undefined
    \item \tw{lgg($\neg$p(s$_1$,$\ldots$,s$_n$),  p(t$_1$,$\ldots$,t$_n$))} is undefined.
\end{itemize}

Using the set representation of a clause, we define the LGG of two clauses:
\begin{center}
    \tw{lgg(cl$_1$,cl$_2$)} = \{ \tw{lgg(l$_1$,l$_2$)} for \tw{l$_1 \in $ cl$_1$} and \tw{l$_2 \in $ cl$_2$}, such that \tw{lgg(l$_1$,l$_2$) is defined}  \}
\end{center}
In other words, the LGG of two clauses is a LGG of all \emph{pairs} of literals of the two clauses.

Having defined the notion of LGG, we move to defining RLGG.
\shortciteS{rlgg} notion of \emph{relative least-general generalisation} computes a LGG of two examples \emph{relative} to the BK (assumed to be a set of facts):
\begin{center}
	\tw{rlgg(e$_1$,e$_2$)} = \tw{lgg(e$_1$ :- BK, e$_2$ :- BK)}.
\end{center}

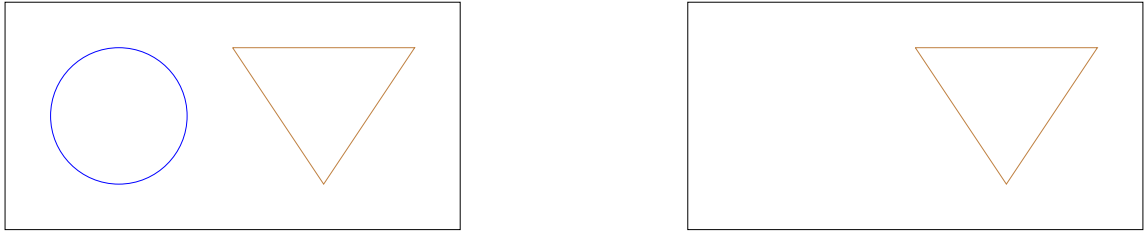
\begin{figure}[t]
	\centering
	\resizebox{\textwidth}{!}{
	\begin{tikzpicture}[scale=1]
		
		\draw (0,0)--(10,0)--(10,5)--(0,5)--cycle;
		
		\draw (15,0)--(25,0)--(25,5)--(15,5)--cycle;
		
		\draw[blue] (2.5,2.5) circle (1.5);
		\draw[brown] (5,4)--(9,4)--(7,1)--(5,4);

		\draw[brown] (20,4)--(24,4)--(22,1)--(20,4);

	\end{tikzpicture}
	}
	\caption{Bongard problems}
	\label{fig:bongard}
\end{figure}
\begin{example}
	To illustrate RLGG, consider the Bongard problems in Figure \ref{fig:bongard} where the goal is to spot the common factor in both images.
	Assume that the images are described with the BK:
	\begin{bk}
		\tw{triangle(o1).}\\
		\tw{triangle(o3).}\\
		\tw{circle(o2).} \\
		\tw{points(o1,down).}\\
		\tw{points(o3,down).}\\
		\tw{contains(1,o1).}\\
		\tw{contains(1,o2).} \\
		\tw{contains(2,o3).}\\
	\end{bk}
\noindent That is, the BK states that objects \tw{o1} and \tw{o3} are triangles, object \tw{o2} is a circle, objects \tw{o1} and \tw{o2} point down, image \tw{1} contains objects \tw{o1} and \tw{o2}, while image \tw{2} contains object 3.
We can use RLGG to identify the common factor, i.e., to find a program representing the common factor.
We will denote the example images as \tw{bon(1)} and \tw{bon(2)}.
We start by formulating the clauses describing examples relative to BK and removing irrelevant parts of BK:

	{\small
	\begin{tabular}{lll}
		\tw{lgg(} & & \\
			      & \big(\tw{bon(1) :-} & \tw{contains(1,o1), contains(1,o2), triangle(o1), points(o1,down),}\\
			      &						& \tw{circle(o2)	, \sout{contains(2,o3), triangle(o3), points(o3,down)}.}\big), \\
			      & \big(\tw{bon(2) :-} & \tw{\sout{contains(1,o1), contains(1,o2), triangle(o1), points(o1,down),}}\\
			      &						& \tw{\sout{circle(o2)}, contains(2,o3), triangle(o3), points(o3,down).}\big) \\
		\tw{)} &
	\end{tabular}
	}

\noindent We proceed by computing LGG for the heads and the bodies of the two clauses separately\footnote{We can do this because when a clause is converted to the set representation, the literals in the body and head have different signs (body literals are negative, while the head literals are positive) which results in an undefined LGG.}.
The LGG of the head literals is \tw{lgg(bon(1),bon(2))} = \tw{bon(lgg(1,2))} = \tw{bon(X)}.
An important thing to note here is that we have to use the \emph{same variable} for the \emph{same ordered pair of terms} everywhere.
For instance, we have used the variable \tw{X} for \tw{lgg(1,2)} and we have to use the same variable every time we encounter the same pair of terms.
To compute the LGG of the body literals, we compute the LGG for all pairs of body literals:
	{\scriptsize
	\begin{code}
		\big\{ \tw{lgg(contains(1,o1),contains(2,o3)), \sout{lgg(contains(1,o1),triangle(o3))}, \sout{lgg(contains(1,o1),points(o3,down))}}\\
		 \tw{lgg(contains(1,o2),contains(2,o3)), \sout{lgg(contains(1,o2),triangle(o3))}, \sout{lgg(contains(1,o1),points(o3,down))}, } \\
		 \tw{\sout{lgg(triangle(o1),contains(2,o3))}, lgg(triangle(o1),triangle(o3)), \sout{lgg(triangle(o1),points(o3,down))},} \\
		 \tw{\sout{lgg(points(o1,down),contains(2,03))}, \sout{lgg(points(o1,down),triangle(o3))}, lgg(points(o1,down),points(o3,down)), } \\
		 \tw{\sout{lgg(circle(o2),contains(2,o3))}, \sout{lgg(circle(o2),triangle(o3))}, \sout{lgg(circle(o2),points(o3,down))}}\big\}.
	\end{code}
	}
	
\noindent Eliminating undefined LGGs leaves us with:
	\begin{code}
		\big\{ \tw{lgg(contains(1,o1),contains(2,o3)), }
		 \tw{lgg(contains(1,o2),contains(2,o3)), } \\
		 \tw{ lgg(triangle(o1),triangle(o3)),} 
		 \tw{lgg(points(o1,down),points(o3,down)) }
		 \big\}.
	\end{code}
\noindent Finally, computing the individual LGGs\footnote{with the following LGGs on terms: \tw{lgg(1,2) = X, lgg(o1,o3) = Y, lgg(o2,o3) = Z}}	
	\begin{code}
		\big\{\tw{contains(X,Y), }
		 \tw{contains(X,Z), } 
		 \tw{triangle(Y),} 
		 \tw{points(Y,down) }
		 \big\}.
	\end{code}
\noindent and eliminating the redundant literal \tw{contains(X,Z)} (as it is subsumed by \tw{contains(X,Y)}) gives us the clause:
	\begin{center}
		\tw{bon(X):- contains(X,Y),triangle(Y),points(Y,down).}
	\end{center}

\noindent
We suggest the book by \shortciteA{luc:book} for more information about generality orders.

\end{example}

\subsubsection{Top-down And Bottom-up}
Progol is one of the most important systems and has inspired many other approaches \shortcite{aleph,xhail,atom}, including Aleph, which we cover in detail in Section \ref{sec:aleph}.
Progol is, however, slightly confusing because it is a top-down system but it first uses a bottom-up approach to bound the search space.
Indeed, many authors only consider it a top-down approach.
Progol uses a set covering algorithm.
Starting with an empty program, Progol picks an uncovered positive example to generalise.
To generalise an example, Progol uses mode declarations (Section \ref{sec:modes}) to build the \emph{bottom clause} \shortcite{progol}, the logically most-specific clause that explains the example.
The use of a bottom clause bounds the search from above (the empty set) and below (the bottom clause).
In this way, Progol is a bottom-up approach because it starts with a bottom clause and tries to generalise it.
However, to find a generalisation of the bottom clause, Progol uses an A* algorithm to search for a generalisation in a top-down (general-to-specific) manner and uses the other examples to guide the search\footnote{The A* search strategy employed by Progol can easily be replaced by alternative search algorithms, such as stochastic search \shortcite{DBLP:journals/ml/MuggletonT08}.}.
In this way, Progol is a top-down approach.
When the search for a generalisation of the bottom clause has finished, Progol adds the clause to its hypothesis (and thus makes it more general) and removes any positive examples entailed by the new hypothesis.
It repeats this process until there are no more positive examples uncovered.
In Section \ref{sec:aleph}, we discuss this approach in more detail when we describe Aleph \shortcite{aleph}, a system similar to Progol.

\subsubsection{Meta-level}
A third new approach has recently emerged called \emph{meta-level} ILP \shortcite{inoue:mla,mugg:metagold,inoue:flap,law:alp,popper}.
There is no agreed-upon definition for what meta-level ILP means, but most approaches encode the ILP problem as a meta-level logic program, i.e.~a program that reasons about programs.
Such meta-level approaches often delegate the search for a hypothesis to an off-the-shelf solver \shortcite{aspal,raspal,mugg:metalearn,ilasp,hexmil,apperception,brute,popper} after which the meta-level solution is translated back to a standard solution for the ILP problem.
In other words, instead of writing a procedure to search in a top-down or bottom-up manner, meta-level approaches formulate the learning problem as a declarative problem, often as an ASP problem \shortcite{aspal,raspal,mugg:metalearn,ilasp,hexmil,apperception,brute,popper}.
For instance, ASPAL \shortcite{aspal} translates an ILP problem into a meta-level ASP program which describes every example and every possible rule in the hypothesis space (defined by mode declarations).
ASPAL then uses an ASP system to find a subset of the rules that cover all the positive but none of the negative examples.
In other words, ASPAL delegates the search to an ASP solver.
ASPAL uses an ASP optimisation statement to find the hypothesis with the fewest literals.

Meta-level approaches can often learn optimal and recursive programs.
Moreover, meta-level approaches use diverse techniques and technologies.
For instance, Metagol \shortcite{mugg:metagold,metagol} uses a Prolog meta-interpreter to search for a proof of a meta-level Prolog program.
ASPAL \shortcite{aspal}, ILASP \shortcite{ilasp}, HEXMIL \shortcite{hexmil}, and the Apperception Engine \shortcite{apperception}  translate an ILP problem into an ASP problem and use powerful ASP solvers to find a model of the problem -- note that these systems all employ very different algorithms.
\dilp{} \shortcite{dilp} uses neural networks to solve the problem.
Overall, the development of meta-level ILP approaches is exciting because it has diversified ILP from the standard clause refinement approach of earlier systems.

For more information about meta-level reasoning, we suggest the work of \shortciteA{inoue:flap}, who provides an introduction to meta-level reasoning and learning.
\shortciteA{law:alp} also provide an overview of \emph{conflict-driven} ILP, which the systems ILASP3 \shortcite{law:thesis} and Popper \shortcite{popper} adopt.

\subsubsection{Discussion}
The different search methods discussed above have different advantages and disadvantages, and there is no `best' approach.
Moreover, as Progol illustrates, there is not necessarily clear distinctions between \emph{top-down}, \emph{bottom-up}, and \emph{meta-level} approaches.
We can, however, make some general observations about the different approaches.

Bottom-up approaches can be seen as being \emph{data-} or \emph{example-driven}.
The major advantage of these approaches is that they are typically very fast.
However, as \shortciteA{hyper} points out, there are several disadvantages of bottom-up approaches, such as (i) they typically use unnecessarily long hypotheses with many clauses, (ii) it is difficult for them to learn recursive hypotheses and multiple predicates simultaneously, and (iii) they do not easily support predicate invention.

The main advantages of top-down approaches are that they can more easily learn recursive programs and textually minimal programs.
The major disadvantage is that they can be prohibitively inefficient because they can generate many hypotheses that do not cover even a single positive example.
Another disadvantage of top-down approaches is their reliance on iterative improvements.
For instance, TILDE keeps specialising every clause which leads to improvement (i.e., a clause covers fewer negative examples).
As such, TILDE can get stuck with suboptimal solutions if the necessary clauses are very long and intermediate specialisations do not improve the score (coverage) of the clause.
To avoid this issue, these systems rely on lookahead~\shortcite{lookahead} which increases the complexity of learning.

The main advantage of meta-level approaches is that they can learn recursive programs and optimal programs \shortcite{aspal,ilasp,hexmil,dilp,apperception,popper}.
They can also harness the state-of-the-art techniques in constraint solving, notably in ASP.
However, some unresolved issues remain.
A key issue is that many approaches encode an ILP problem as a single (often very large) ASP problem \shortcite{aspal,ilasp,hexmil,apperception}, so struggle to scale to problems with very large domains.
Moreover, since most ASP solvers only work on ground programs \shortcite{clingo}, pure ASP-based approaches are inherently restricted to tasks that have a small and finite grounding.
Although preliminary work attempts to tackle this issue \shortcite{popper,dcc}, work is still needed for these approaches to scale to very large problems.
Many approaches also precompute every possible rule in a hypothesis \shortcite{aspal,ilasp}, so struggle to learn programs with large rules, although preliminary work tries to address this issue \shortcite{brute}.

\section{ILP Features}
\label{sec:compare}

Table \ref{tab:comp} compares the same systems from Table \ref{tab:build} on a small number of dimensions.
This table excludes many other important dimensions of comparison, such as whether a system supports non-observational predicate learning, where examples of the target relations are not directly given \shortcite{progol}.
We discuss these features in turn.

    \begin{table}[ht]
\centering
\footnotesize
\begin{tabular}{l|ccccc}
\textbf{System}                        & \textbf{Noise} & \textbf{Optimality} & \textbf{Infinite domains} & \textbf{Recursion} & \textbf{Predicate invention} \\ \midrule
\textbf{FOIL} \shortcite{foil}                & Yes            & No                  & Yes                       & Partly             & No                           \\ \midrule
\textbf{Progol} \shortcite{progol}         & Yes             & No                 & Yes                    & Partly             & No                           \\ \midrule
\textbf{Claudien} \shortcite{claudien}         & Yes     & No        & Yes   & Party         & No         \\
\midrule
\textbf{TILDE} \shortcite{tilde}                & Yes            & No                  & Yes                       & No                 & No\tablefootnote{
    A logical decision tree learned by TILDE can be translated into a logic program that contains invented predicate symbols.
    However, TILDE is unable to reuse any invented symbols whilst learning.
}                           \\ \midrule
\textbf{Aleph} \shortcite{aleph}         & Yes             & No                 & Yes                    & Partly             & No                           \\ \midrule
\textbf{XHAIL} \shortcite{xhail}                & Yes            & No                  & Yes                       & Partly             & No                           \\ \midrule
\textbf{ASPAL} \shortcite{aspal}                & No             & Yes                 & No                        & Yes                & No                           \\ \midrule
\textbf{Atom} \shortcite{atom}                & Yes             & No                 & Yes                        & Partly                & No                           \\ \midrule
\textbf{QuickFOIL} \shortcite{quickfoil}                & Yes            & No                  & No                       & Partly             & No                           \\ \midrule

\textbf{LFIT} \shortcite{inoue:lfit}                & No            & Yes                 & No                        & No\tablefootnote{
    LFIT does not support recursion in the rules but allows recursion in their usage.
    The input is a set of pairs of interpretations and the output is a logic program that can be recursively applied on its output to produce sequences of interpretations.
}                & No                       \\ \midrule
\textbf{ILASP} \shortcite{ilasp}                & Yes\footnote{ILASP3\cite{law:thesis} supports noisy examples.}            & Yes                 & Partly\tablefootnote{
    ILASP precomputes every rule defined by a given mode declaration $M$ to form a rule space $S_M$.
    Given background knowledge $B$ and an example $E$, ILASP requires that the grounding of $B \cup S_M \cup E$ must be finite.
}                        & Yes                & Partly                       \\ \midrule
\textbf{Metagol} \shortcite{mugg:metagold}              & No             & Yes                 & Yes                       & Yes                & Yes                          \\ \midrule
\textbf{\dilp{}} \shortcite{dilp} & Yes            & Yes                 & No                        & Yes                & Partly                       \\ \midrule
\textbf{HEXMIL} \shortcite{hexmil}               & No             & Yes                 & No                       & Yes                & Yes                           \\
\midrule

\textbf{FastLAS} \shortcite{law:fastlas}                & Yes            & Yes                 & Partly\tablefootnote{
    Given background knowledge $B$ and an example $E$, FastLAS requires that the grounding of $B \cup S_M \cup E$ must be finite.
} & No & No\\
\midrule

\textbf{Apperception} \shortcite{apperception}               & Yes             & Yes                 & No                       & Yes                & Partly                           \\ \midrule
\textbf{Popper} \shortcite{popper}               & Yes             & Yes                 & Yes                       & Yes                & Yes                           \\
\end{tabular}

\caption{
    A vastly simplified comparison of ILP systems.
    As with Table \ref{tab:build}, this table is meant to provide a very high-level overview of some systems.
    Therefore, the table entries are coarse and should not be taken absolutely literally.
    For instance, Metagol does not support noise and thus has the value \emph{no} in the noise column, but there is an extension \shortcite{mugg:vision} that samples examples to mitigate the issue of misclassified examples.
    ILASP and \dilp{} support predicate invention, but only a restricted form.
    See Section \ref{sec:pi} for an explanation.
    FOIL, Progol, and XHAIL can learn recursive programs when given sufficient examples.
    See Section \ref{sec:recursion} for an explanation.
}
\label{tab:comp}
\end{table}

\subsection{Noise}
\label{sec:noise}

Noise handling is important in ML.
In ILP, we can distinguish between three types of noise:
\begin{itemize}
    \setlength\itemsep{1pt}
    \setlength\parskip{1pt}
    \item \textbf{Noisy examples}: where an example is misclassified
    \item \textbf{Incorrect BK}: where a relation holds when it should not (or does not hold when it should)
    \item \textbf{Imperfect BK}: where relations are missing or there are too many irrelevant relations
\end{itemize}

\noindent
We discuss these three types of noise.

\paragraph{Noisy examples.}

The problem definitions from Section \ref{sec:ilp} are too strong to account for noisy (incorrectly labelled) examples because they expect a hypothesis that entails all of the positive and none of the negative examples.
Therefore, most systems relax this constraint and accept a hypothesis that does not necessarily cover all positive examples or that covers some negative examples\footnote{It is, unfortunately, a common misconception that ILP cannot handle mislabelled examples \shortcite{dilp}.}.
Most systems that use a set covering loop naturally support noise handling.
For instance, TILDE extends a decision tree learner \shortcite{id3,c45} to the first-order setting and uses the same information gain methods to induce hypotheses.
The noise-tolerant version of ILASP \shortcite{law:thesis} uses ASP's optimisation abilities to provably learn the program with the best coverage.
In general, handling noisy examples is a well-studied topic in ILP.

\paragraph{Noisy BK.}
Just as training examples can potentially be noisy/misclassified, the facts/rules in the BK can be noisy/misclassified.
For instance, if learning rules to forecast the weather, the BK might include facts about historical weather, which might not be 100\% correct.
However, most systems assume that the BK is perfect, i.e. that atoms are true or false, and there is no room for uncertainty.
This assumption is a major limitation because real-world data, such as images or speech, cannot always be easily translated into a purely noise-free symbolic representation.
We discuss this limitation in more detail in Section \ref{sec:noisyinput}.
One of the key appealing features of \dilp{} is that it takes a differentiable approach to ILP and can be given fuzzy or ambiguous data.
Rather than an atom being true or false, \dilp{} gives atoms continuous semantics, which maps atoms to the real unit interval [0, 1].
The authors successfully demonstrate the approach on the MNIST dataset.

\paragraph{Imperfect BK.}
Handling imperfect BK is an under-explored topic in ILP.
We can distinguish between two types of imperfect BK: missing BK and too much BK, which we discussed in Section \ref{sec:bk:discuss}.
\subsection{Optimality}

There are often multiple (sometimes infinite) hypotheses that solve the ILP problem (or have the same training error).
In such cases, which hypothesis should we choose?

\subsubsection{Occamist Bias}

Many systems try to learn a textually minimal hypothesis.
This approach is justified as following an \emph{Occamist bias} \shortcite{DBLP:journals/ml/Schaffer93}.
The most common interpretation of an Occamist bias is that amongst all hypotheses consistent with the data, the simplest is the most likely\footnote{
\shortciteA{pedro:occam} points out that this interpretation is controversial, partly because Occam’s razor is interpreted in two different ways. Following \shortciteA{pedro:occam}, let the \emph{generalisation error} of a hypothesis be its error on unseen examples and the \emph{training error} be its error on the examples it was learned from.
The formulation of the razor that is perhaps closest to Occam’s original intent is \emph{given two hypotheses with the same generalisation error, the simpler one should be preferred because simplicity is desirable in itself}. The second formulation, which most ILP systems follow, is different and can be stated as \emph{given two hypotheses with the same training error, the simpler one should be preferred because it is likely to have lower generalisation error}. \shortciteA{pedro:occam} points out that the first razor is largely uncontroversial, but the second one, taken literally, is provably and empirically false \shortcite{occams_test}. Many systems do not distinguish between the two cases. We therefore also do not make any distinction.
}.
Most approaches use an Occamist bias to find the smallest hypothesis, measured in terms of the number of clauses \shortcite{mugg:metagold}, literals \shortcite{ilasp}, or description length \shortcite{progol}.
Most systems are not, however, guaranteed to induce the smallest programs.
A key reason for this limitation is that many approaches learn a single clause at a time leading to the construction of sub-programs that are sub-optimal in terms of program size and coverage.
For instance, Aleph, described in detail in the next section, offers no guarantees about the program size and coverage.
Newer systems address this limitation \shortcite{aspal,ilasp,metagol,hexmil,popper} through meta-level reasoning (Section \ref{sec:method}).
For instance, ASPAL \shortcite{aspal} is given as input a hypothesis space with a set of candidate clauses.
The ASPAL task is to find a minimal subset of clauses that entails all the positive and none of the negative examples.
ASPAL uses ASP's optimisation abilities to provably learn the program with the fewest literals.

\subsubsection{Cost-minimal Programs}
Learning efficient logic programs has long been considered a difficult problem \shortcite{mugg:ilp94,ilp20}, mainly because there is no declarative difference between an efficient program, such as mergesort, and an inefficient program, such as bubble sort.
To address this issue, Metaopt \shortcite{metaopt} learns efficient programs.
Metaopt maintains a cost during the hypothesis search and uses this cost to prune the hypothesis space.
To learn minimal time complexity logic programs, Metaopt minimises the number of resolution steps.
For instance, imagine learning a \emph{find duplicate} program, which finds a duplicate element in a list e.g. \tw{[p,r,o,g,r,a,m]} $\mapsto$ \tw{r}, and \tw{[i,n,d,u,c,t,i,o,n]
} $\mapsto$ \tw{i}.
Given suitable input data, Metagol induces the program:

\begin{code}
f(A,B):- head(A,B),tail(A,C),element(C,B).\\
f(A,B):- tail(A,C),f(C,B).
\end{code}

\noindent
This program goes through the elements of the list checking whether the same element exists in the rest of the list.
Given the same input, Metaopt induces the program:

\begin{code}
f(A,B):- mergesort(A,C),f1(C,B).\\
f1(A,B):- head(A,B),tail(A,C),head(C,B).\\
f1(A,B):- tail(A,C),f1(C,B).
\end{code}

\noindent
This program first sorts the input list and then goes through the list to check for duplicate adjacent elements.
Although larger, both in terms of clauses and literals, the program learned by Metaopt is more efficient ($O(n \log n)$) than the program learned by Metagol ($O(n^2)$).

Other systems can also learn optimal programs \shortcite{inspire}.
For instance, FastLAS \shortcite{law:fastlas} follows this idea and takes as input a custom scoring function and computes an optimal solution for the given scoring function.
The authors show that this approach allows a user to optimise domain-specific performance metrics on real-world datasets, such as access control policies.
\subsection{Infinite Domains}

Some systems, mostly meta-level approaches, cannot handle infinite domains \shortcite{aspal,raspal,dilp,hexmil,apperception}.
Pure ASP-based systems \shortcite{aspal,hexmil,apperception} struggle to handle infinite domains because (most) current ASP solvers only work on ground programs, i.e. they need a finite grounding.
ASP can work with infinite domains as long as the grounding is finite.
This finite grounding restriction is often achieved by enforcing syntactic restrictions on the programs, such as finitely-ground programs \shortciteA{DBLP:conf/iclp/CalimeriCIL08}.
ASP systems (which combine a grounder and a solver), such as Clingo \shortcite{clingo}, first take a first-order program as input, ground it using an ASP grounder, and then use an ASP solver to determine whether the ground problem is satisfiable.
This approach leads to the \emph{grounding bottleneck} problem \shortcite{grounding}, where grounding can be so large that it is simply intractable.
This grounding issue is especially problematic when reasoning about complex data structures, such as lists.
For instance, grounding the permutation relation over the ASCII characters would require 128! facts.
The grounding bottleneck is especially a problem when reasoning about real numbers\footnote{Most ASP implementations do not natively support lists nor real numbers, although both can be represented using other means.}.
For instance, ILASP \shortcite{ilasp} can represent real numbers as strings and can delegate reasoning to Python (via Clingo's scripting feature).
However, in this approach, the numeric computation is performed when grounding the inputs, so the grounding must be finite, which makes it impractical.
This grounding problem is not specific to ASP-based systems.
For instance, \dilp{} is an ILP system based on a neural network, but it only works on BK in the form of a finite set of ground atoms.
This grounding problem is essentially the fundamental problem faced by table-based ML approaches that we discussed in Section \ref{sec:bk}.

One approach to mitigate this problem is to use \emph{context-dependent examples} \shortcite{law:context}, where BK can be associated with specific examples so that an ILP systems need only ground part of the BK.
Although this approach is shown to improve the grounding problem compared to not using context-dependent examples, the approach still needs a finite grounding for each example and still struggles as the domain size increases \shortcite{popper}.

\subsection{Recursion}
\label{sec:recursion}

The power of recursion is that an infinite number of computations can be described by a finite recursive program \shortcite{DBLP:books/daglib/0067086}.
In ILP, recursion is often crucial for generalisation.
We illustrate this importance with two examples.

\begin{example}[Reachability]
Consider learning the concept of \emph{reachability} in a graph.
Without recursion, an ILP system would need to learn a separate clause to define reachability of different lengths. For instance, to define reachability depths for 1-4 would require the program:

\begin{code}
reachable(A,B):- edge(A,B).\\
reachable(A,B):- edge(A,C),edge(C,B).\\
reachable(A,B):- edge(A,C),edge(C,D),edge(D,B).\\
reachable(A,B):- edge(A,C),edge(C,D),edge(D,E),edge(E,B).
\end{code}

\noindent
This program does not generalise because it does not define reachability for arbitrary depths.
Moreover, most systems would need examples of each depth to learn such a program.
By contrast, a system that supports recursion can learn the program:

\begin{code}
reachable(A,B):- edge(A,B).\\
reachable(A,B):- edge(A,C),reachable(C,B).
\end{code}

\noindent
Although smaller, this program generalises reachability to any depth.
Moreover, systems can learn this definition from a small number of examples of arbitrary reachability depth.

\end{example}

\begin{example}[String transformations]
Reconsider the string transformation problem from the introduction (Section \ref{sec:intro:string}).
As with the reachability example, without recursion, a system would need to learn a separate clause to find the last element for each list of length $n$, such as:

\begin{code}
last(A,B):- tail(A,C),empty(C),head(A,B).\\
last(A,B):- tail(A,C),tail(C,D),empty(D),head(C,B).\\
last(A,B):- tail(A,C),tail(C,D),tail(D,E),empty(E),head(E,B).
\end{code}

\noindent
By contrast, a system that supports recursion can learn the compact program:

\begin{code}
last(A,B):- tail(A,C),empty(C),head(A,B).\\
last(A,B):- tail(A,C),last(C,B).
\end{code}

\noindent
Because of the symbolic representation and the recursive nature, this program generalises to lists of arbitrary length and which contain arbitrary elements (e.g. integers and characters).
\end{example}

\noindent
Without recursion it is often difficult for a system to generalise from small numbers of examples \shortcite{datacurate}.
Moreover, recursion is vital for many program synthesis tasks, such as the quicksort scenario from the introduction.
Despite its importance, learning recursive programs has long been a difficult problem \shortcite{ilp20}.
Moreover, there are many negative theoretical results on the learnability of recursive programs \shortcite{cohen:pac2}.
As Table \ref{tab:comp} shows, many systems cannot learn recursive programs, or can only learn it in a limited form.

A common limitation is that many systems rely on \emph{bottom clause construction} \shortcite{progol}, which we discuss in detail in Section \ref{sec:aleph}.
In this approach, for each positive example, a system creates the most specific clause that entails the example and then tries to generalise the clause to entail other examples.
However, because a system learns only a single clause per example\footnote{This statement is not true for all systems that employ bottom clause construction. XHAIL \shortcite{xhail}, for instance, can induce multiple clauses per example.}, this covering approach requires examples of both the base and inductive cases, which means that such systems struggle to learn recursive programs, especially from small numbers of examples.

Interest in recursion has resurged recently with the introduction of meta-interpretive learning (MIL) \shortcite{mugg:metalearn,mugg:metagold,metaho} and the MIL system Metagol \shortcite{metagol}.
The key idea of MIL is to use metarules (Section \ref{sec:metarules}) to restrict the form of inducible programs and thus the hypothesis space.
For instance, the \emph{chain} metarule ($P(A,B) \leftarrow Q(A,C), R(C,B)$) allows Metagol to induce programs\footnote{Metagol can induce longer clauses though predicate invention, which we discuss in Section \ref{sec:pi}.} such as:

\begin{code}
f(A,B):- tail(A,C),head(C,B).
\end{code}

\noindent
Metagol induces recursive programs using recursive metarules, such as the \emph{tail recursive} metarule \emph{P(A,B) $\leftarrow$ Q(A,C), P(C,B)}.
Metagol can also learn mutually recursive programs, such as learning the definition of an even number by also inventing and learning the definition of an odd number (\tw{even\_1}):

\begin{code}
even(0).\\
even(A):- successor(A,B),even\_1(B).\\
even\_1(A):- successor(A,B),even(B).
\end{code}

\noindent
Many systems can now learn recursive programs \shortcite{ilasp,dilp,hexmil,apperception,popper}.
With recursion, systems can generalise from small numbers of examples, often a single example \shortcite{metabias,playgol}.
For instance, Popper \shortcite{popper} can learn list transformation programs from only a handful of examples, such as a program to drop the last element of a list:

\begin{code}
droplast(A,B):- tail(A,B),empty(B).\\
droplast(A,B):- tail(A,C),droplast(C,D),head(A,E),cons(E,D,B).
\end{code}

\noindent
The ability to learn recursive programs has opened ILP to new application areas, including learning string transformations programs \shortcite{metabias}, robot strategies \shortcite{metagolo}, context-free grammars \shortcite{mugg:metalearn}, and answer set grammars \shortcite{law:asg}.
\subsection{Predicate Invention}
\label{sec:pi}

Most systems assume that the given BK is suitable to induce a solution.
This assumption may not always hold.
Rather than expecting a user to provide all the necessary BK, the goal of \emph{predicate invention} (PI) is for a system to automatically invent new auxiliary predicate symbols, i.e.~ to introduce new predicate symbols in a hypothesis that are not given the examples nor the BK\footnote{PI is a form of \emph{non-observational predicate learning}, where examples of the target relations are not directly given \shortcite{progol}.}
This idea is similar to when humans create new functions when manually writing programs, such as to reduce code duplication or to improve readability.
For instance, to learn the quicksort algorithm, a learner needs to be able to partition the list given a pivot element and append two lists.
If \tw{partition} and \tw{append} are not provided in BK, the learner would need to invent them.

PI has repeatedly been stated as an important challenge \shortcite{cigol,stahl:pi,DBLP:journals/jetai/Muggleton94,ilp20}.
\shortciteA{russell:humancomp} even argues that the automatic invention of new high-level concepts is the most important step needed to reach human-level AI.
A classical example of PI is learning the definition of \tw{grandparent} from only the background relations \tw{mother} and \tw{father}.
Given suitable examples and no other background relations, a system can learn the program:

\begin{code}
grandparent(A,B):- mother(A,C),mother(C,B).\\
grandparent(A,B):- mother(A,C),father(C,B).\\
grandparent(A,B):- father(A,C),mother(C,B).\\
grandparent(A,B):- father(A,C),father(C,B).
\end{code}

\noindent
Although correct, this program is large and has 4 clauses and 12 literals.
By contrast, consider the program learned by a system which supports PI:

\begin{code}
grandparent(A,B):- inv(A,C),inv(C,B).\\
inv(A,B):- mother(A,B).\\
inv(A,B):- father(A,B).
\end{code}

\noindent
To learn this program, a system has \emph{invented} a new predicate symbol \tw{inv}.
This program is semantically equivalent\footnote{
This use of the term \emph{semantically equivalent} is imprecise.
Whether these two programs are strictly equivalent depends on the definition of logical equivalence, for which there are many \shortcite{DBLP:books/mk/minker88/Maher88}.
Moreover, equivalence between the two programs is further complicated because they have different vocabularies (because of the invented predicate symbol).
Our use of equivalence is based on the two programs having the same logical consequences for the target predicate symbol \tw{grandparent}.
} to the previous one, but is shorter both in terms of the number of literals and clauses.
The invented symbol \tw{inv} can be interpreted as \emph{parent}.
In other words, if we rename \tw{inv} to \tw{parent} we have the program:

\begin{code}
grandparent(A,B):- parent(A,C),parent(C,B).\\
parent(A,B):- mother(A,B).\\
parent(A,B):- father(A,B).
\end{code}

\noindent
As this example shows, PI can help learn smaller programs, which, in general, is preferable because most systems struggle to learn large programs \shortcite{iggp,brute}.





\noindent
PI has been shown to help reduce the size of programs, which in turn reduces sample complexity and improves predictive accuracy \shortcite{curled,playgol,metaho,alps,knorf}.


To further illustrate the power of PI, imagine learning a \tw{droplasts} program, which removes the last element of each sublist in a list, e.g. \tw{[alice,bob,carol]} $\mapsto$ \tw{[alic,bo,caro]}.
Given suitable examples and BK, Metagol$_{ho}$ \shortcite{metaho} learns the higher-order program:

\begin{code}
droplasts(A,B):- map(A,B,droplasts1).\\
droplasts1(A,B):- reverse(A,C),tail(C,D),reverse(D,B).
\end{code}

\noindent
To learn this program, Metagol$_{ho}$ invents the predicate symbol \tw{droplasts1}, which is used twice in the program: once as term in the literal \tw{map(A,B,droplasts1)} and once as a predicate symbol in the literal \tw{droplasts1(A,B)}.
This higher-order program uses \tw{map} to abstract away the manipulation of the list to avoid the need to learn an explicitly recursive program (recursion is implicit in \tw{map}).

Now consider learning a \emph{double droplasts} program (\tw{ddroplasts}), which extends the droplast problem so that, in addition to dropping the last element from each sublist, it also drops the last sublist, e.g. \tw{[alice,bob,carol]} $\mapsto$ \tw{[alic,bo]}.
Given suitable examples, metarules, and BK, Metagol$_{ho}$ learns the program:

\begin{code}
ddroplasts(A,B):- map(A,C,ddroplasts1),ddroplasts1(C,B).\\
ddroplasts1(A,B):- reverse(A,C),tail(C,D),reverse(D,B).
\end{code}

\noindent
This program is similar to the aforementioned \tw{droplasts} program, but additionally reuses the invented predicate symbol \tw{ddroplasts1} in the literal \tw{ddroplasts1(C,B)}.
This program illustrates the power of PI to help learn substantially more complex programs.

Most early attempts at PI were unsuccessful, and, as Table \ref{tab:comp} shows, most systems do not support it.
As \shortciteA{kramer1995predicate} points out, PI is difficult for at least three reasons:
\begin{itemize}
    \setlength\itemsep{1pt}
    \setlength\parskip{1pt}
    \item When should we invent a new symbol? There must be a reason to invent a new symbol; otherwise, we would never invent one.
    \item How should you invent a new symbol? How many arguments should it have?
    \item How do we judge the quality of a new symbol? When should we keep an invented symbol?
\end{itemize}

\noindent
There are many PI techniques.
We briefly discuss some approaches now.

\subsubsection{Inverse Resolution}
Early work on PI was based on the idea of inverse resolution \shortcite{cigol} and specifically \emph{W operators}.
Discussing inverse resolution in depth is beyond the scope of this paper.
We refer the reader to the original work of \shortciteA{cigol} or the overview books by \shortciteA{ilp:book} and \shortciteA{luc:book} for more information.
Although inverse resolution approaches could support PI, they never demonstrated completeness, partly because of the lack of a declarative bias to delimit the hypothesis space \shortcite{mugg:metagold}.

\subsubsection{Placeholders}
One approach to PI is to predefine invented symbols through mode declarations, which \shortciteA{DBLP:conf/ilp/LebanZB08} call \emph{placeholders} and which \shortciteA{law:thesis} calls \emph{prescriptive PI}.
For instance, to invent the parent relation, a suitable modeh declaration would be required, such as:
\begin{code}
modeh(1,inv(person,person)).
\end{code}
\noindent
However, this placeholder approach is limited because it requires that a user manually specify the arity and argument types of a symbol \shortcite{ilasp}, which rather defeats the point, or requires generating all possible invented predicates \shortcite{dilp,apperception}, which is computationally expensive.

\subsubsection{Metarules}
To reduce the complexity of PI, Metagol uses metarules (Section \ref{sec:metarules}) to define the hypothesis space.
For instance, the \emph{chain} metarule ($P(A,B) \leftarrow Q(A,C), R(C,B)$) allows Metagol to induce programs such as:

\begin{code}
f(A,B):- tail(A,C),tail(C,B).
\end{code}

\noindent
This program drops the first two elements from a list.
To induce longer clauses, such as to drop first three elements from a list, Metagol can use the same metarule but can invent a new predicate symbol and then chain their application, such as to induce the program\footnote{
We could unfold \shortcite{unfolding} this program to remove the invented symbol to derive the program \tw{f(A,B):- tail(A,C),tail(C,D),tail(D,B).}}:
\noindent

\begin{code}
f(A,B):- tail(A,C),inv(C,B).\\
inv(A,B):- tail(A,C),tail(C,B).
\end{code}

\noindent
A side-effect of this metarule-driven approach to PI is that problems are forced to be decomposed into smaller problems.
For instance, suppose you want to learn a program that drops the first four elements of a list, then Metagol could learn the following program, where the invented predicate symbol \tw{inv} is used twice:

\begin{code}
f(A,B):- inv(A,C),inv(C,B).\\
inv(A,B):- tail(A,C),tail(C,B).
\end{code}

\noindent
To learn this program, Metagol invents the predicate symbol \tw{inv} and induces a definition for it using the \emph{chain} metarule.
Metagol uses this new predicate symbol in the definition for the target predicate \tw{f}.

\subsubsection{Lifelong Learning}
\label{sec:lifelong}

\begin{figure}[ht]
\centering
\includegraphics[scale=.4]{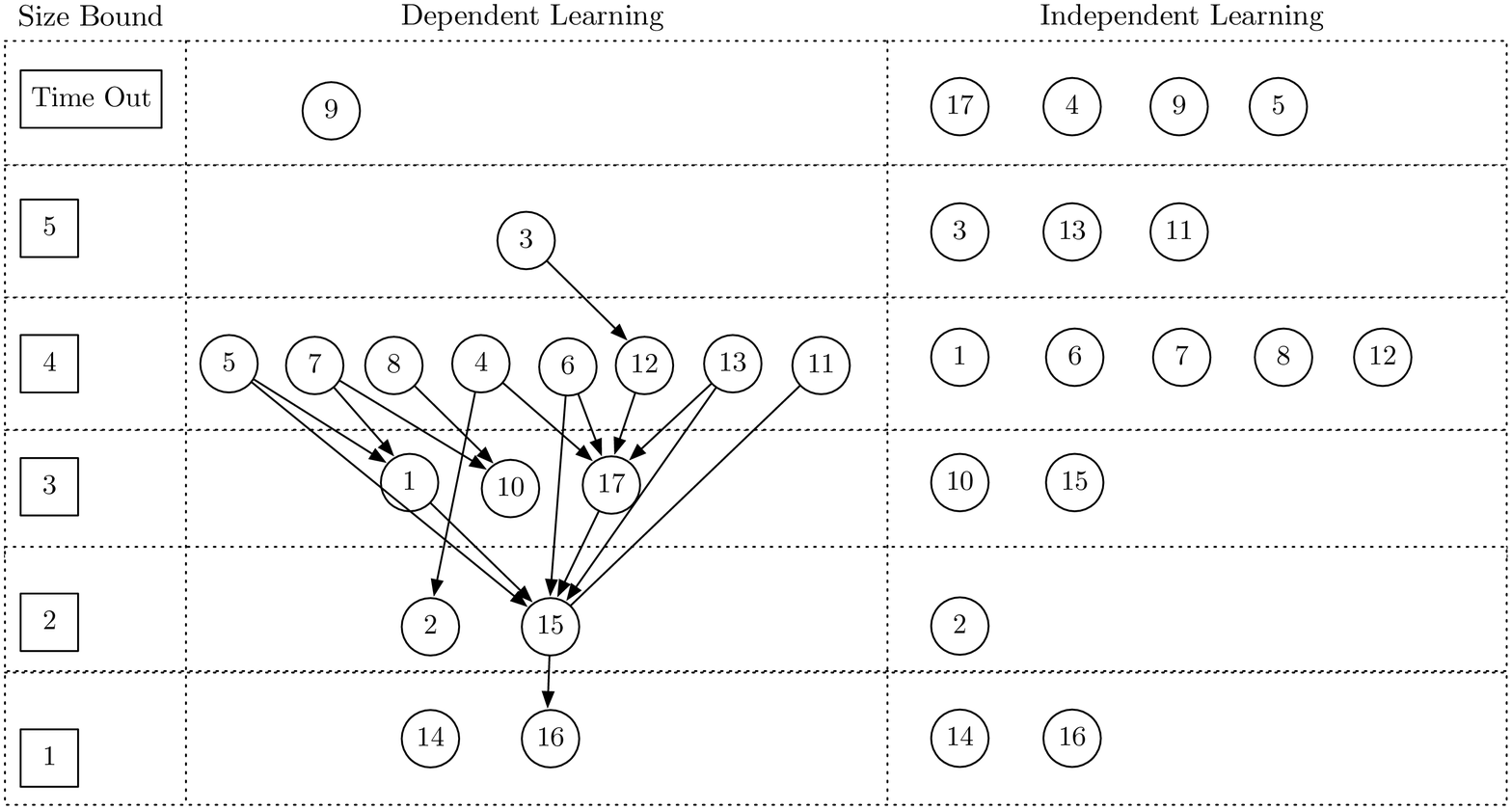}

\caption{
This figure is taken from the work of \shortciteA{metabias}.
It shows the programs learned by dependent (left) and independent (right) learning approaches.
The size bound column denotes the number of clauses in the induced program.
The nodes correspond to programs and the numbers denote the task that the program solves.
For the dependent learning approach, the arrows correspond to the calling relationships of the induced programs.
For instance, the program to solve task 3 reuses the solution to solve task 12, which in turn reuses the solution to task 17, which in turn reuses the solution to task 15.
Tasks 4, 5, and 16 cannot be solved using an independent learning approach, but can when using a dependent learning approach.
}
\label{fig:metabias}
\end{figure}

The aforementioned techniques for PI are aimed at single-task problems.
PI can be performed by continually learning programs (meta-learning).
For instance \shortciteA{metabias} use a technique called \emph{dependent learning} to enable Metagol to learn string transformations programs over time.
Given a set of 17 string transformation tasks, their learner automatically identifies easier problems, learn programs for them, and then reuses the learned programs to help learn programs for more difficult problems.
To determine which problems are easier to solve, the authors initially start with a very strong bias in the form of allowing a learner to only use one rule to find a solution.
They then progressively relax this restriction, each time allowing more rules in a solution.
The authors use PI to reform the bias of the learner where after a solution is learned not only is the target predicate added to the BK but also its constituent invented predicates.
The authors experimentally show that their multi-task approach performs substantially better than a single-task approach because learned programs are frequently reused.
Moreover, they show that this approach leads to a hierarchy of BK composed of reusable programs, where each builds on simpler programs.
Figure \ref{fig:metabias} shows this approach.
Note that this lifelong setting raises challenges, which we discuss in Section \ref{sec:limits}.


\subsubsection{Theory Refinement}
The aim of theory \emph{refinement} \shortcite{wrobel} is to improve the \emph{quality} of a theory.
Theory \emph{revision} approaches \shortcite{AdeRevision,forte} revise a program so that it entails missing answers or does not entail incorrect answers.
Theory \emph{compression} \shortcite{DeRaedtLuc2008CpPp} approaches select a subset of clauses such that the performance is minimally affected with respect to certain examples.
Theory \textit{restructuring} changes the structure of a logic program to optimise its execution or its readability~\shortcite{flach:pi,wrobel}.
We discuss two recent refinement approaches based on PI.


\paragraph{Auto-encoding logic programs.}
\emph{Auto-encoding logic programs} (ALPs) \shortcite{alps} invent predicates by simultaneously learning a pair of logic programs: (i) an \textit{encoder} that maps the examples given as interpretations to new interpretations defined entirely in terms of invented predicates\footnote{The head of every clause in the encoder invents a predicate.}, and (ii) a \textit{decoder} that reconstructs the original interpretations from the invented ones.
The invented interpretations compress the given examples and invent useful predicates by capturing regularities in the data\footnote{
    Evaluating the usefulness of invented predicates via their ability to compress a theory goes back to some of the earliest work in ILP by the Duce system \shortcite{duce}.
}.
ALPs, therefore, change the representation of the problem.
The most important implication of the approach is that the target programs are easier to express via the invented predicates.
The authors experimentally show that learning from the representation invented by ALPs improves the learning performance of generative Markov logic networks (MLN) \shortcite{richardson:mln}.
Generative MLNs learn a (probabilistic) logic program that explains all predicates in an interpretation, not a single target predicate.
The predicates invented by ALPs, therefore, aid the learning of all predicates in the BK.

\paragraph{Program refactoring.}
Knorf~\shortcite{knorf} pushes the idea of ALPs even further.
After learning to solve user-supplied tasks in the lifelong learning setting, Knorf compresses the learnt program by removing redundancies in it.
If the learnt program contains invented predicates, Knorf revises them and introduces new ones that would lead to a smaller program\footnote{
        The idea of learning new predicates to restructure knowledge bases goes back at least to \shortciteA{flach:pi}.
}.
By doing so, Knorf optimises the representation of obtained knowledge.
The refactored program is smaller in size and contains fewer redundant clauses.
The authors experimentally demonstrate that refactoring improves learning performance in lifelong learning.
More precisely, Metagol learns to solve more tasks when using the refactored BK, especially when BK is large.
Moreover, the authors also demonstrate that Knorf substantially reduces the size of the BK program, reducing the number of literals in a program by 50\% or more.

\section{ILP Case Studies}
\label{sec:cases}

We now describe in detail four ILP systems: Aleph \shortcite{aleph}, TILDE \shortcite{tilde}, ASPAL \shortcite{aspal}, and Metagol \shortcite{metagol}.
These systems are not necessarily the best, nor the most popular, but use considerably different techniques and are relatively simple to explain.
Aleph is based on \emph{inverse entailment} \shortcite{progol} and uses \emph{bottom clause construction} to restrict the hypothesis space.
Despite its age, Aleph is still one of the most popular systems.
TILDE is a first-order generalisation of decision trees and uses information gain to divide and conquer the training examples.
ASPAL is a meta-level system that uses an ASP solver to solve the ILP problem, which has influenced much subsequent work, notably ILASP.
Finally, Metagol uses a Prolog meta-interpreter to construct a proof of a set of examples and extracts a program from the proof.
We discuss these systems in turn.

\subsection{Aleph}
\label{sec:aleph}

Progol \shortcite{progol} is arguably the most influential ILP system, having influenced many systems \shortcite{cfinduction,aleph,xhail,atom}, which in turn have inspired many other systems \shortcite{iled,oled,inspire}.
Aleph is based on Progol.
We discuss Aleph, rather than Progol, because the implementation, written in Prolog, is easier to use and the manual is more detailed.

\subsubsection{Aleph Setting}

The Aleph problem setting is:
\begin{description}
    \setlength\itemsep{1pt}
\setlength\parskip{1pt}
    \item [\textbf{Given:}]
    \item [-] A set of mode declarations $M$
    \item [-] Background knowledge $B$ in the form of a normal program
    \item [-] Positive ($E^+$) and negative ($E^-$) examples represented as sets of ground facts
\end{description}
\begin{description}
    \setlength\itemsep{1pt}
\setlength\parskip{1pt}
    \item [\textbf{Return:}] A normal program hypothesis $H$ such that:
    \item [-] $H$ is consistent with $M$
    \item [-] $\forall e \in E^+, \;  H \cup B \models e$ (i.e.~is complete)
    \item [-] $\forall e \in E^-, \;  H \cup B \not\models e$ (i.e.~is consistent)
\end{description}

\noindent
Note that the examples can be different relations to generalise, i.e. they can have different predicate symbols.

\subsubsection{Aleph Algorithm}
Aleph starts with an empty hypothesis and uses the following set covering approach:
\begin{enumerate}
    \setlength\itemsep{1pt}
\setlength\parskip{1pt}
    \item Select a positive example to generalise. If none exists, stop and return the current hypothesis; otherwise proceed to the next step.
    \item Construct the most specific clause (the bottom clause) \shortcite{progol} that is consistent with the mode declarations (Section \ref{sec:modes}) and entails the example.
    \item Search for a clause more general than the bottom clause and has the best score.
    \item Add the clause to the hypothesis and remove all the positive examples covered by it. Return to step 1.
\end{enumerate}

\noindent
We discuss the basic approaches to steps 2 and 3.

\subsubsection*{Step 2: Bottom Clause Construction}
The purpose of constructing a bottom clause is to bound the search in step 3.
The bottom clause is the most specific clause that entails the example to be generalised.
In general, a bottom clause can have infinite cardinality.
Therefore, Aleph uses \emph{mode declarations} (Section \ref{sec:modes}) to restrict them.
Describing how to construct bottom clauses is beyond the scope of this paper.
See the paper by \shortciteA{progol} or the book of \shortciteA{luc:book} for contrasting methods.
Having constructed a bottom clause, Aleph can ignore any clauses that are not more general than it.
In other words, Aleph only considers clauses that are generalisations of the bottom clause, which must all entail the example.
We use the bottom clause definition provided by \shortciteA{luc:book}:

\begin{definition}[Bottom clause]
Let $H$ be a clausal hypothesis and $C$ be a clause.
Then the bottom clause $\bot(C)$ is the most specific clause such that:
\[H \cup \bot(C) \models C\]
\end{definition}

\begin{example}[Bottom clause]
To illustrate bottom clauses in Aleph, we use a modified example from \shortciteA{luc:book}.
Let $M$ be the mode declarations:
\begin{modes}
\tw{:- modeh(*,pos(+shape)).}\\
\tw{:- modeb(*,red(+shape)).}\\
\tw{:- modeb(*,blue(+shape)).}\\
\tw{:- modeb(*,square(+shape)).}\\
\tw{:- modeb(*,triangle(+shape)).}\\
\tw{:- modeb(*,polygon(+shape)).}
\end{modes}

\noindent
Let $B$ be the BK:
\begin{bk}
\tw{red(s1)}.\\
\tw{blue(s2)}.\\
\tw{square(s1)}.\\
\tw{triangle(s2)}.\\
\tw{polygon(A):- rectangle(A).}\\
\tw{rectangle(A):- square(A).}\\
\end{bk}
\noindent
Let $e$ be the positive example \tw{pos(s1)}.
Then:
\begin{code}
$\bot$(e) = pos(A):- red(A),square(A),rectangle(A),polygon(A).
\end{code}
\noindent
This bottom clause contains the literal \tw{rectangle(A)} because it is implied by \tw{square(A)}.
The inclusion of \tw{rectangle(A)} in turn implies the inclusion of \tw{polygon(A)}.
Although \tw{blue} and \tw{triangle} appear in $B$, they are irrelevant to $e$, so do not appear in the bottom clause.
\end{example}
\noindent
Any clause that is not more general than the bottom clause cannot entail $e$ so can be ignored.
For instance, we can ignore the clause \tw{pos(A):- blue(A)} because it is not more general than $\bot(e)$.

\paragraph{Constant symbols.}
Note that $\bot(e)$ contains variables, rather than constant symbols, which would make the bottom clause even more specific.
The reason is that the given mode declarations forbid constant symbols.
Had \tw{modeb(*,polygon(\#shape))} been given in $M$, then $\bot(e)$ would also contain \tw{polygon(s1)}.



\subsubsection*{Step 3: Clause Search}
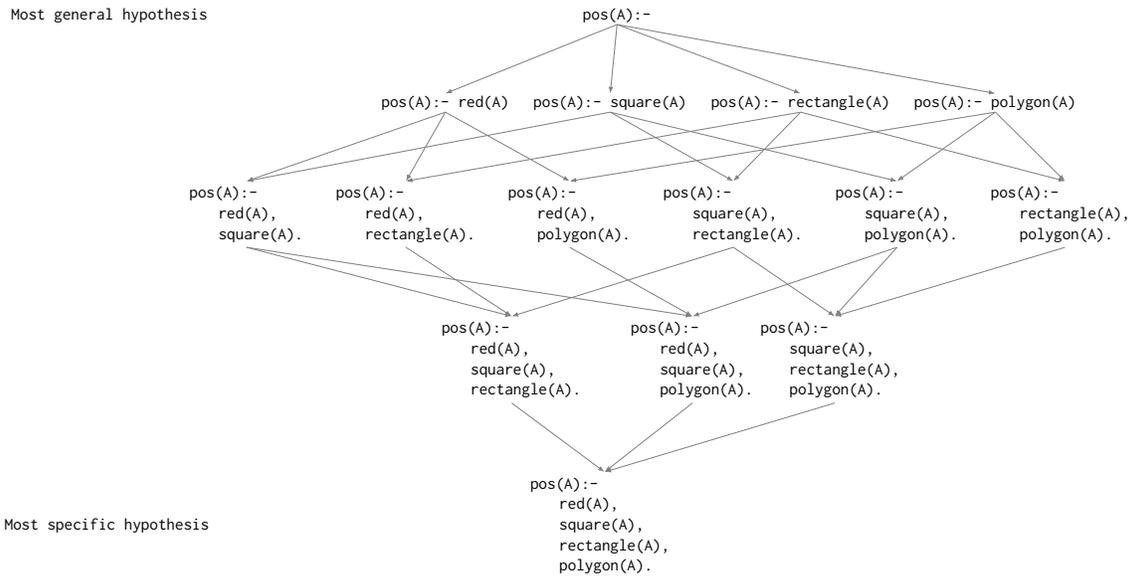
\begin{figure}[ht]
	\centering
	\resizebox{\textwidth}{!}{
	\centering
	\begin{tikzpicture}[scale=1]

		\node (top) at (0,0) {\Huge \tw{pos(A):-}};

		\node[below left=5cm of top] (l1) {\Huge \tw{pos(A):- red(A)}};
		\node[below left=5cm of top, right=of l1] (l2) {\Huge \tw{pos(A):- square(A)}};
		\node[below right=5cm of top, right=of l2] (l3) {\Huge \tw{pos(A):- rectangle(A)}};
		\node[below right=5cm of top, right=of l3] (l4) {\Huge \tw{pos(A):- polygon(A)}};

		\draw[-{Latex}, line width=0.05cm, gray] (top.south)--(l1.north);
		\draw[-{Latex}, line width=0.05cm, gray] (top.south)--(l2.north);
		\draw[-{Latex}, line width=0.05cm, gray] (top.south)--(l3.north);
		\draw[-{Latex}, line width=0.05cm, gray] (top.south)--(l4.north);

		\node[below left=5cm of l1] (l1_1) {
			\Huge
			\begin{tabular}{l}
				\tw{pos(A):- } \\
				\quad\quad \tw{red(A), } \\
				\quad\quad \tw{square(A).}\\
			\end{tabular}
		};

		\node[below=5cm of l1, right=of l1_1] (l1_2) {
			\Huge
			\begin{tabular}{l}
				\tw{pos(A):- } \\
				\quad\quad \tw{red(A), } \\
				\quad\quad \tw{rectangle(A).}\\
			\end{tabular}
		};

		\node[below right=5cm of l1, right=of l1_2] (l1_3) {
			\Huge

			\begin{tabular}{l}
				\tw{pos(A):- } \\
				\quad\quad \tw{red(A), } \\
				\quad\quad \tw{polygon(A).}\\
			\end{tabular}
		};

		\node[below right=of l2, right=of l1_3] (l1_4) {
			\Huge
			\begin{tabular}{l}
				\tw{pos(A):- } \\
				\quad\quad \tw{square(A), } \\
				\quad\quad \tw{rectangle(A).}\\
			\end{tabular}
		};

		\node[below right=of l3, right=of l1_4] (l1_5) {
			\Huge
			\begin{tabular}{l}
				\tw{pos(A):- } \\
				\quad\quad \tw{square(A), } \\
				\quad\quad \tw{polygon(A).}\\
			\end{tabular}
		};

		\node[below right=of l4, right=of l1_5] (l1_6) {
			\Huge
			\begin{tabular}{l}
				\tw{pos(A):- } \\
				\quad\quad \tw{rectangle(A), } \\
				\quad\quad \tw{polygon(A).}\\
			\end{tabular}
		};

		\draw[-{Latex}, line width=0.05cm, gray] (l1.south)--(l1_1.north);
		\draw[-{Latex}, line width=0.05cm, gray] (l1.south)--(l1_2.north);
		\draw[-{Latex}, line width=0.05cm, gray] (l1.south)--(l1_3.north);
		\draw[-{Latex}, line width=0.05cm, gray] (l2.south)--(l1_1.north);
		\draw[-{Latex}, line width=0.05cm, gray] (l2.south)--(l1_4.north);
		\draw[-{Latex}, line width=0.05cm, gray] (l2.south)--(l1_5.north);
		\draw[-{Latex}, line width=0.05cm, gray] (l3.south)--(l1_2.north);
		\draw[-{Latex}, line width=0.05cm, gray] (l3.south)--(l1_4.north);
		\draw[-{Latex}, line width=0.05cm, gray] (l3.south)--(l1_6.north);
		\draw[-{Latex}, line width=0.05cm, gray] (l4.south)--(l1_3.north);
		\draw[-{Latex}, line width=0.05cm, gray] (l4.south)--(l1_5.north);
		\draw[-{Latex}, line width=0.05cm, gray] (l4.south)--(l1_6.north);

		\node[below left=5cm of l1_4] (l2_1) {
			\Huge
			\begin{tabular}{l}
				\tw{pos(A):- } \\
				\quad\quad \tw{red(A), } \\
				\quad\quad \tw{square(A), } \\
				\quad\quad \tw{rectangle(A).}\\
			\end{tabular}
		};

		\node[below left=5cm of l1_5] (l2_2) {
			\Huge
			\begin{tabular}{l}
				\tw{pos(A):- } \\
				\quad\quad \tw{red(A), } \\
				\quad\quad \tw{square(A), } \\
				\quad\quad \tw{polygon(A).}\\
			\end{tabular}
		};

		\node[below left=5cm of l1_6] (l2_3) {
			\Huge
			\begin{tabular}{l}
				\tw{pos(A):- } \\
				\quad\quad \tw{square(A), } \\
				\quad\quad \tw{rectangle(A), } \\
				\quad\quad \tw{polygon(A).}\\
			\end{tabular}

		};

		\draw[-{Latex}, line width=0.05cm, gray] (l1_1.south)--(l2_1.north);
		\draw[-{Latex}, line width=0.05cm, gray] (l1_2.south)--(l2_1.north);
		\draw[-{Latex}, line width=0.05cm, gray] (l1_4.south)--(l2_1.north);
		\draw[-{Latex}, line width=0.05cm, gray] (l1_1.south)--(l2_2.north);
		\draw[-{Latex}, line width=0.05cm, gray] (l1_3.south)--(l2_2.north);
		\draw[-{Latex}, line width=0.05cm, gray] (l1_5.south)--(l2_2.north);
		\draw[-{Latex}, line width=0.05cm, gray] (l1_4.south)--(l2_3.north);
		\draw[-{Latex}, line width=0.05cm, gray] (l1_5.south)--(l2_3.north);
		\draw[-{Latex}, line width=0.05cm, gray] (l1_6.south)--(l2_3.north);

		\node[below left=5cm of l2_3] (l3_1) {
			\Huge
			\begin{tabular}{l}
				\tw{pos(A):- } \\
				\quad\quad \tw{red(A), } \\
				\quad\quad \tw{square(A), } \\
				\quad\quad \tw{rectangle(A), } \\
				\quad\quad \tw{polygon(A).}\\
			\end{tabular}
		};

		\draw[-{Latex}, line width=0.05cm, gray] (l2_1.south)--(l3_1.north);
		\draw[-{Latex}, line width=0.05cm, gray] (l2_2.south)--(l3_1.north);
		\draw[-{Latex}, line width=0.05cm, gray] (l2_3.south)--(l3_1.north);

		\node[left=19cm of top,black] {\tw{\Huge Most general hypothesis}};
		\node[left=16cm of l3_1,black] {\tw{\Huge Most specific hypothesis}};

	\end{tikzpicture}
	}
	\caption{Aleph bounds the hypothesis space from above (the most general hypothesis) and below (the most specific hypothesis). Aleph starts the search from the most general hypothesis and specialises it (by adding literals from the bottom clause) until it finds the best hypothesis.}
	\label{fig:sys:aleph}
\end{figure}

Having constructed a bottom clause, Aleph searches for generalisations of it.
The importance of constructing the bottom clause is that it bounds the search space from below (the bottom clause).
Figure \ref{fig:sys:aleph} illustrates the search space of Aleph when given the bottom clause $\bot$(e) from our previous shape example.
Aleph performs a bounded breadth-first search to enumerate shorter clauses before longer ones, although a user can easily change the search strategy\footnote{Progol, by contrast, uses an A* search \shortcite{progol}.}.
The search is bounded by several parameters, such as a maximum clause size and a maximum proof depth.
In this scenario, Aleph starts with the most general generalisation of $\bot$(e), which is \emph{pos(A)}, which simply says that everything is true.
Aleph evaluates (assigns a score) to each clause in the search, i.e. each clause in the lattice.
Aleph's default evaluation function is \emph{coverage} defined as $P - N$, where $P$ and $N$ are the numbers of positive and negative examples respectively entailed by the clause\footnote{Aleph comes with 13 evaluation functions, such as \emph{entropy} and \emph{compression}.}.
Aleph then tries to specialise a clause by adding literals to the body of it, which it selects from the bottom clause or by instantiating variables.
Each specialisation of a clause is called a \emph{refinement}.
Properties of refinement operators \shortcite{mis} are well-studied in ILP \shortcite{ilp:book,luc:book}, but are beyond the scope of this paper.
The key thing to understand is that Aleph's search is bounded from above (the most general clause) and below (the most specific clause).
Having found the best clause, Aleph adds it to the hypothesis, removes all the positive examples covered by the new hypothesis, and returns to Step 1.
Describing the full clause search mechanism and how the score is computed is beyond the scope of this work, so we refer to the reader to the Progol tutorial by \shortciteA{muggleton2001relational} for a more detailed introduction.

\subsubsection{Discussion}

\paragraph{Advantages.}
Aleph is one of the most popular ILP systems because (i) it has a stable and easily available implementation with many options, and (ii) it has good empirical performance.
Moreover, it is a single Prolog file, which makes it easy to download and use\footnote{Courtesy of Fabrizio Riguzzi and Paolo Niccolò Giubelli, Aleph is now available as a SWIPL package at https://www.swi-prolog.org/pack/list?p=aleph}.
Because it uses a bottom clause to bound the search, Aleph is also efficient at identifying relevant constant symbols that may appear in a hypothesis, which is not the case for pure top-down approaches\footnote{
    As the Aleph manual states, ``the bottom clause is really useful to introduce constants (these are obtained from the seed example''.
}.
Aleph also supports many other features, such as numerical reasoning, inducing constraints, and allowing user-supplied cost functions.

\paragraph{Disadvantages.}
Because it is based on inverse entailment, and only learns a single clause at a time, Aleph struggles to learn recursive programs and optimal programs and does not support PI.
Aleph also uses many parameters, such as parameters that change the search strategy when generalising a bottom clause (step 3) and parameters that change the structure of learnable programs (such as limiting the number of literals in the bottom clause).
These parameters can greatly influence learning performance.
Even for experts, it is non-trivial to find a suitable set of parameters for a problem.
\subsection{TILDE}
\label{sec:tilde}

TILDE \shortcite{tilde} is a first-order generalisation of decision trees, and specifically the C4.5 \shortcite{c45} learning algorithm.
TILDE learns from interpretations, instead of entailment as Aleph, and is an instance of top-down methodology.

The learning setup of TILDE is different from other ILP systems in the sense that it closely mimics the standard classification setup from machine learning.
That is, TILDE learns a program that assigns a class to an example (an interpretation).
Assigning a class is done by checking which class an example entails, given the program and background knowledge.
Class assignment is indicated as a special literal of the form \tw{class(c)}, where \tw{c} comes from a set of classes $C$.
For instance, if a classifier differentiates between cats and dog then $C = \{cat, dog\}$.

\subsubsection{TILDE Setting}

The TILDE problem setting is:
\begin{description}
	\setlength\itemsep{1pt}
    \setlength\parskip{1pt}
    \item [\textbf{Given:}]
    \item [-] A set of classes $C$
    \item [-] A set of mode declarations
    \item [-] A set of examples $E$ (a set of interpretations)
    \item [-] $BK$ in the form of a definite program
    
\end{description}

\begin{description}
	\setlength\itemsep{1pt}
    \setlength\parskip{1pt}
    \item [\textbf{Return:}] A (normal) logic program hypothesis $H$ such that:
    \item [-] $\forall e \in E, \;  H \wedge BK \wedge e \models class(c)$, $c \in C$,
    where c is the class of the example e
    \item [-] $\forall e \in E, \;  H \wedge BK \wedge e \not\models class(c^{\prime})$, $c^{\prime} \in C - \{c\}$
\end{description}

\subsubsection{TILDE Algorithm}

TILDE behaves almost the same as C4.5 limited to binary attributes, meaning that it uses the same heuristics and pruning techniques.
What TILDE does differently is the generation of candidates splits.
Whereas C4.5 generates candidates as attribute-value pairs (or value inequalities in case of continuous attributes), TILDE uses conjunctions of literals.
The conjunctions are explored gradually from the most general to the most specific ones, where $\theta$-subsumption (Section \ref{sec:logic}) is used as an ordering.

To find a hypothesis, TILDE employs a divide-and-conquer strategy recursively repeating the following steps:
\begin{itemize}
	\setlength\itemsep{1pt}
    \setlength\parskip{1pt}
	\item if all examples belong to the same class, create a leaf predicting that class
	\item for each candidate conjunction $conj$, find the normalised information gain when splitting on $conj$
		\begin{itemize}
			\item if no candidate provides information gain, turn the previous node into a leaf predicting the majority class
		\end{itemize}
	\item  create a decision node $n$ that splits on the candidate conjunction with the highest information gain
	\item Recursively split on the subsets of data obtained by the splits and add those nodes as children of $n$
\end{itemize}

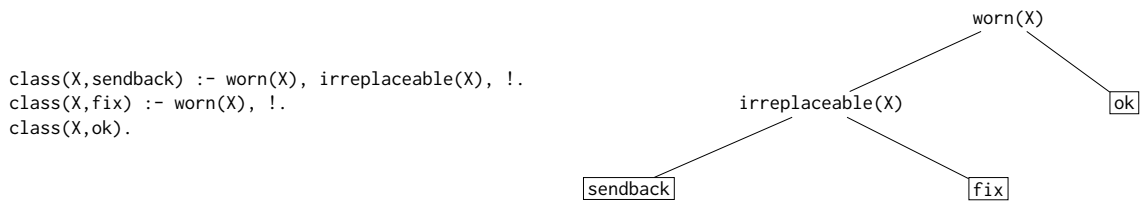
\begin{figure}[t]
	\centering
	\Large
	\resizebox{\textwidth}{!}{
	\begin{tikzpicture}[scale=1]

		\node (top) at (0,0) {\huge \tw{worn(X)}};

		\node[below left=3cm of top] (left_branch) {\huge \tw{irreplaceable(X)}};

		\node[draw, below right=3cm of top] (right_branch) {\huge \tw{ok}};

		\node[draw,below left=3cm of left_branch] (left_branch_2) {\huge \tw{sendback}};

		\node[draw, below right=3cm of left_branch] (right_branch_2) {\huge \tw{fix}};

		\draw[-] (top)--(left_branch);
		\draw[-] (top)--(right_branch);
		\draw[-] (left_branch)--(left_branch_2);
		\draw[-] (left_branch)--(right_branch_2);

		\node[left=7cm of left_branch] {
			\huge
			\begin{tabular}{l}
				\tw{class(X,sendback) :- worn(X), irreplaceable(X), !.} \\
				\tw{class(X,fix) :- worn(X), !.} \\
				\tw{class(X,ok).}
			\end{tabular}
		};

	\end{tikzpicture}
	}
	\caption{TILDE learns tree-shaped (normal) programs. Clauses in the program correspond to paths along the tree.}
	\label{fig:sys:tilde}
\end{figure}

\begin{example}[Machine repair example \shortcite{tilde}]
To illustrate TILDE's learning procedure, consider the following example.
Each example is an interpretation (a set of facts) and it describes (i) a machine with parts that are worn out, and (ii) an action an engineer should perform: \tw{fix} the machine, \tw{send it back} to the manufacturer, or nothing if the machine is \tw{ok}.
These actions are the classes to predict.

\begin{exs}
	\tw{E1: \{worn(gear).  worn(chain).  class(fix).\}} \\
	\tw{E2: \{worn(engine).  worn(chain).  class(sendback).\}}\\
	\tw{E3: \{worn(wheel).  class(sendback).\}} \\
	\tw{E4: \{class(ok).\}} \\
\end{exs}

\noindent
Background knowledge contains information which parts are replaceable and which are not:

\begin{bk}
	\tw{replaceable(gear).}\\
	\tw{replaceable(chain).}\\
	\tw{irreplaceable(engine).}\\
	\tw{irreplaceable(wheel).}\\
\end{bk}

\noindent
Like any top-down approach, TILDE starts with the most general program; in this case, the initial program assigns the majority class to all examples.
TILDE then gradually refines the program (specialises it) until satisfactory performance is reached.
To refine the program, TILDE uses mode declarations.
Due to the top-down nature of TILDE, it is more natural to understand modes as conjunctions that can be added to a current clause.
This interpretation does not conflict the explanation given in Section \ref{sec:bias}.
TILDE interprets an \textit{input} argument as bounding to a variable that already exists in the current clause; this is identical to stating that the argument needs to be instantiated when literal is called as it will be instantiated by the existing literal that introduces that variable.
Similarly, in TILDE, an \textit{output} argument introduces a new variable; this variable will be instantiated after the literal is called.

Assume the mode declarations:
\begin{code}
	modeb(*,replaceable(+X)). \\
	modeb(*,irreplaceable(+X)). \\
	modeb(*,worn(+X)). \\
\end{code}

\noindent
Each mode declaration forms a candidate split:
\begin{code}
	worn(X). \\
	replaceable(X). \\
	irreplaceable(X).
\end{code}

\noindent
Computing the information gain for all of the candidate splits (as with propositional C4.5), the conjunction \tw{worn(X)} results in the highest gain and is set as the root of the tree.
For details about the C4.5 and information gain, we refer the reader to the excellent machine learning book by \shortciteA{tm:book}.

TILDE proceeds by recursively repeating the same procedure over both outcomes of the test: when \tw{worn(X)} is \tw{true} and \tw{false}.
When the root test fails, the dataset contains a single example (\tw{E4}); TILDE forms a branch by creating the leaf predicting the class \tw{ok}.
When the root test succeeds, not all examples (\tw{E1, E2, E3}) belong to the same class.
TILDE thus refines the root node further:

\begin{center}
\begin{minipage}{.48\linewidth}
	\begin{code}
		worn(X), worn(X). \\
		worn(X), replaceable(X). \\
		worn(X), irreplaceable(X)
	\end{code}
\end{minipage}
\begin{minipage}{.48\linewidth}
	\begin{code}
		worn(X), worn(Y). \\
		worn(X), replaceable(Y). \\
		worn(X), irreplaceable(Y)
	\end{code}
\end{minipage}
\end{center}

\noindent
The candidate refinement \tw{worn(X), irreplaceable(X)} perfectly divides the remaining examples and thus \tw{irreplaceable(X)} is added as the subsequent test.
All examples are classified correctly, and thus the learning stops.

The final TILDE tree is (illustrated in Figure \ref{fig:sys:tilde}):
\begin{code}
	class(sendback):- worn(X),irreplaceable(X),!. \\
	class(fix):- worn(X),!. \\
	class(ok).
\end{code}

\end{example}

\noindent
Note the usage of the cut (\tw{!}) operator, which is essential to ensure that only one branch of the decision tree holds for each example.

\subsubsection{Discussion}

\paragraph{Advantages.}
An interesting aspect of TILDE is that it learns normal logic programs (which include negation) instead of definite logic programs.
Another advantage of TILDE is that, compared to other ILP systems, it supports both categorical and numerical data.
Indeed, TILDE is an exception among ILP systems, which usually struggle to handle numerical data.
At any refinement step, TILDE can add a literal of the form \tw{<(X,V)}, or equivalently \tw{X < V} with \tw{V} being a value.
TILDE's stepwise refinement keeps the number of inequality tests tractable.

\paragraph{Disadvantages.}
Although TILDE learns normal programs, it requires them to be in the shape of a tree and does not support recursion.
Furthermore, TILDE inherits the limitations of top-down systems, such as generating many needless candidates.
Another weakness of TILDE is the need for lookahead.
Lookahead is needed when a single literal is useful only in a conjunction with another literal.
Consider, for instance, that the machine repair scenario has a relation \tw{number\_of\_components} and the target rule that a machine needs to be fixed when a part consisting of more than three parts is worn out:
\begin{code}
	class(fix):- worn(X),number\_of\_components(X,Y),Y > 3.
\end{code}

\noindent
To find this clause, TILDE would first refine the clause:
\begin{code}
	class(fix):- worn(X).
\end{code}
\noindent into:
\begin{code}
	class(fix):- worn(X),number\_of\_components(X,Y).
\end{code}
\noindent However, this candidate clause would be rejected as it yields no information gain (every example covered by the first clause is also covered by the second clause).
The introduction of a literal with the \tw{number\_of\_components} predicate is only helpful if it is introduced together with the inequality related to the second argument of the literal.
Informing TILDE about this dependency is known as lookahead.

\subsection{ASPAL}
\label{sec:aspal}

ASPAL \shortcite{aspal} was one of the first meta-level ILP systems, which directly influenced other ILP systems, notably ILASP.
ASPAL builds on TAL \cite{tal}, but is simpler to explain\footnote{
A key difference is that TAL is implemented in Prolog and uses list structure to iteratively build rules. By contrast, using list-like structure in ASP is inefficient and often impossible as the solver completely grounds the program before solving it.}.
Indeed, ASPAL is one of the simplest ILP systems to explain.
It uses the mode declarations to build every possible clause that could be in a hypothesis.
It adds a flag to each clause indicating whether the clause should be in a hypothesis.
It then formulates the problem of deciding which flags to turn on as an ASP problem.

\subsubsection{ASPAL Setting}

The ASPAL problem setting is:
\begin{description}
    \setlength\itemsep{0pt}
    \setlength\parskip{0pt}
    \item [\textbf{Given:}]
    \item [-] A set of mode declarations $M$
    \item [-] $B$ in the form of a normal program
    \item [-] $E^+$ positive examples represented as a set of facts
    \item [-] $E^-$ negative examples represented as a set of facts
    \item [-] A penalty function $\gamma$
\end{description}
\begin{description}
    \setlength\itemsep{0pt}
    \setlength\parskip{0pt}
    \item [\textbf{Return:}] A normal program hypothesis $H$ such that:
    \item [-] $H$ is consistent with $M$
    \item [-] $\forall e \in E^+, \;  H \cup B \models e$ (i.e.~is complete)
    \item [-] $\forall e \in E^-, \;  H \cup B \not\models e$ (i.e.~is consistent)
    \item [-] The penalty function $\gamma$ is minimal
\end{description}

\subsubsection{ASPAL Algorithm}

ASPAL encodes an ILP problem as a meta-level ASP program.
The answer sets of this meta-level program are solutions to the ILP problem.
The ASPAL algorithm is one of the simplest in ILP:

\begin{enumerate}
    \setlength\itemsep{0pt}
    \setlength\parskip{0pt}
    \item Generate all possible rules consistent with the given mode declarations. Assign each rule a unique identifier and add that as an abducible (guessable) atom in each rule.
    \item Use an ASP solver to find a minimal subset of the rules (by formulating the problem as an ASP optimization problem).
\end{enumerate}

\noindent
Step 1 is a little more involved, and we explain why below.
Also, similar to Aleph, ASPAL has several input parameters that constrain the size of the hypothesis space, such as the maximum number of body literals and the maximum number of clauses.
Step 2 uses an ASP optimisation statement to learn a program with a minimal penalty.

\subsubsection{ASPAL Example}

\begin{example}[ASPAL]
To illustrate ASPAL, we slightly modify the example from \shortciteA{aspal}.
We also ignore the penalty statement.
ASPAL is given as input $B$, $E^+$, $E^-$, and $M$:
\begin{bk}
\tw{bird(alice).}\\
\tw{bird(betty).}\\
\tw{can(alice,fly).}\\
\tw{can(betty,swim).}\\
\tw{ability(fly).}\\
\tw{ability(swim).}\\
\end{bk}
\[
\tw{E}^+ = \left\{
\begin{array}{l}
\tw{penguin(betty).}
\end{array}
\right\}
\tw{E}^- = \left\{
\begin{array}{l}
\tw{penguin(alice).}
\end{array}
\right\}
\]
\[
\tw{M} = \left\{
\begin{array}{l}
\tw{modeh(1, penguin(+bird)).}\\
\tw{modeb(1, bird(+bird)).}\\
\tw{modeb(*,not can(+bird,\#ability))}\\
\end{array}
\right\}
\]
\noindent
Given these modes\footnote{Note that `notcan` is used in the original ASPAL paper but we think this usage is a typo.}
, the possible rules are:
\begin{code}
penguin(X):- bird(X).\\
penguin(X):- bird(X), not can(X,fly).\\
penguin(X):- bird(X), not can(X,swim).\\
penguin(X):- bird(X), not can(X,swim), not can(X,fly).\\
\end{code}
\noindent
ASPAL generates skeleton rules which replace constants with variables and adds an extra literal to each rule as an abducible literal:
\begin{code}
penguin(X):- bird(X), rule(r1).\\
penguin(X):- bird(X), not can(X,C1), rule(r2,C1).\\
penguin(X):- bird(X), not can(X,C1), not can(X,C2), rule(r3,C1,C2).\\
\end{code}
\noindent
ASPAL forms a meta-level ASP program from these rules that is passed to an ASP solver:
\begin{code}
bird(alice).\\
bird(betty).\\
can(alice,fly).\\
can(betty,swim).\\
ability(fly).\\
ability(swim).\\
penguin(X):- bird(X), rule(r1).\\
penguin(X):- bird(X), not can(X,C1), rule(r2,C1).\\
penguin(X):- bird(X), not can(X,C1), not can(X,C2), rule(r3,C1,C2).\\
0 \{rule(r1),rule(r2,fly),rule(r2,swim),rule(r3,fly,swim)\}4.\\
goal : - penguin(betty), not penguin(alice).\\
: - not goal.
\end{code}
\noindent
The key statement in this meta-level program is:
\begin{code}
0 \{rule(r1),rule(r2,fly),rule(r2,swim),rule(r3,fly,swim)\}4.\\
\end{code}
\noindent
This statement is a choice rule, which states none or at most four of the literals \{rule(r1), rule(r2,fly), rule(r2,swim), rule(r3,fly,swim)\} could be true.
The job of the ASP solver is to determine which of those literals should be true (formulated as an ASP optimization problem), which corresponds to an answer set for this program:
\begin{code}
rule(r2,c(fly)).
\end{code}
\noindent
Which is translated to a program:
\begin{code}
penguin(A):- not can(A,fly).
\end{code}
\end{example}

\subsubsection{Discussion}

\paragraph{Advantages.}
A major advantage of ASPAL is its sheer simplicity, which has inspired other approaches, notably ILASP.
It also learns optimal programs by employing ASP optimisation constraints.

\paragraph{Disadvantages.}
The main limitation of ASPAL is scalability.
It precomputes every possible rule in a hypothesis, which is infeasible on all but trivial problems.
For instance, when learning game rules from observations \shortcite{iggp}, ASPAL performs poorly for this reason.
\subsection{Metagol}
An \emph{interpreter} is a program that evaluates (interprets) programs.
A \emph{meta-interpreter} is an interpreter written in the same language that it evaluates.
Metagol \shortcite{mugg:metagold,metagol,metaho} is a form of ILP based on a Prolog meta-interpreter.

\subsubsection{Metagol Setting}

The Metagol problem setting is:

\begin{description}
    \setlength\itemsep{0pt}
    \setlength\parskip{0pt}
    \item [\textbf{Given:}]
    \item [-] A set of metarules $M$
    \item [-] $B$ in the form of a normal program
    \item [-] $E^+$ positive examples represented as a set of facts
    \item [-] $E^-$ negative examples represented as a set of facts
\end{description}

\begin{description}
  \setlength\itemsep{0pt}
  \setlength\parskip{0pt}
    \item [\textbf{Return:}] A definite program hypothesis $H$ such that:
    \item [-] $\forall e \in E^+, \;  H \cup B \models e$ (i.e.~is complete)
    \item [-] $\forall e \in E^-, \;  H \cup B \not\models e$ (i.e.~is consistent)
    \item [-] $\forall h \in H, \;
     \exists m \in M$ such that $h=m\theta$, where $\theta$ is a substitution that grounds all the existentially quantified variables in $m$
\end{description}

\noindent
The last condition ensures that a hypothesis is an instance of the given metarules.
It is this condition that enforces the strong inductive bias in Metagol.

\subsubsection{Metagol Algorithm}

Metagol uses the following procedure to find a hypothesis:

\begin{enumerate}
    \setlength\itemsep{0pt}
    \setlength\parskip{0pt}
    \item Select a positive example (an atom) to generalise. If one exists, proceed to step 2.
    If none exists, test the hypothesis on the negative examples. If the hypothesis does not entail any negative example stop and return the hypothesis; otherwise backtrack to a choice point at step 2 and continue.
    \item Try to prove the atom by:
    \begin{enumerate}
      \setlength\itemsep{0pt}
      \setlength\parskip{0pt}
      \item using given BK or an already induced clause
      \item unifying the atom with the head of a metarule (Section \ref{sec:metarules}), binding the variables in a metarule to symbols in the predicate and constant signatures, saving the substitutions, and then proving the body of the metarule through meta-interpretation (by treating the body atoms as examples and applying step 2 to them)
  \end{enumerate}
\end{enumerate}

\noindent
In other words, Metagol induces a logic program by constructing a proof of the positive examples.
It uses metarules to guide the proof search.
After proving all the examples, Metagol checks the consistency of the hypothesis against the negative examples.
If the program is inconsistent, Metagol backtracks to explore different proofs (hypotheses).

\paragraph{Metarules.}
Metarules are fundamental to Metagol.
For instance, the \emph{chain} metarule is \tw{P(A,B):- Q(A,C), R(C,B)}.
The letters \tw{P}, \tw{Q}, and \tw{R} denote second-order variables.
Metagol internally represents metarules as Prolog atoms of the form \tw{metarule(Name,Subs,Head,Body)}.
Here \tw{Name} denotes the metarule name, \tw{Subs} is a list of variables that Metagol should find substitutions for, and \tw{Head} and \tw{Body} are list representations of a clause.
For example, the internal representation of the \emph{chain} metarule is \tw{metarule(chain,[P,Q,R], [P,A,B], [[Q,A,C],[R,C,B]])}.
Metagol represents substitutions, which we will call \emph{metasubs}, as atoms of the form \tw{sub(Name,Subs)}, where \tw{Name} is the name of the metarule and \tw{Subs} is a list of substitutions.
For instance, binding the variables \tw{P}, \tw{Q}, and \tw{R} with \tw{second}, \tw{tail}, and \tw{head} respectively in the chain metarule leads to the metasub \tw{sub(chain,[second,tail,head])} and the clause
\tw{second(A,B):- tail(A,C),head(C,B)}.

\paragraph{Optimality.}
To learn optimal programs, Metagol enforces a bound on the program size (the number of metasubs).
Metagol uses iterative deepening to search for hypotheses.
At depth $d=1$ Metagol is allowed to induce at most one clause, i.e. use at most one metasub.
If such a hypothesis exists, Metagol returns it.
Otherwise, Metagol continues to the next iteration $d+1$.
At each iteration $d$, Metagol introduces $d-1$ new predicate symbols and is allowed to use $d$ clauses.
New predicates symbols are formed by taking the name of the task and adding underscores and numbers.
For example, if the task is \tw{f} and the depth is $4$ then Metagol will add the predicate symbols \tw{f\_1}, \tw{f\_2}, and \tw{f\_3} to the predicate signature.

\subsubsection{Metagol Example}

\begin{example}[Kinship example]
To illustrate Metagol, suppose you have the following BK:

\begin{bk}
\tw{mother(ann,amy).}\;\;\; \tw{mother(ann,andy).}\\
\tw{mother(amy,amelia).}\;\;\; \tw{mother(amy,bob).}\\
\tw{mother(linda,gavin).}\\
\tw{father(steve,amy).}\;\;\; \tw{father(steve,andy).}\\
\tw{father(andy,spongebob).}\;\;\; \tw{father(gavin,amelia).}
\end{bk}

\noindent
And the following metarules represented in Prolog:

\begin{code}
metarule(ident,[P,Q], [P,A,B], [[Q,A,B]]).\\
metarule(chain,[P,Q,R], [P,A,B], [[Q,A,C],[R,C,B]]).\\
\end{code}

\noindent
We can call Metagol with a lists of positive ($E^+$) and negative ($E^-$) examples:


\[
\tw{E}^+ = \left\{
\begin{array}{l}
\tw{grandparent(ann,amelia).}\\
  \tw{grandparent(steve,amelia).}\\
  \tw{grandparent(steve,spongebob).}\\
  \tw{grandparent(linda,amelia).}\\
\end{array}
\right\}
\tw{E}^- = \left\{
\begin{array}{l}
\tw{grandparent(amy,amelia).}
\end{array}
\right\}
\]

\noindent
In Step 1, Metagol selects an example to generalise.
Suppose Metagol selects \tw{grandparent(ann,amelia)}.
In Step 2a, Metagol tries to prove this atom using the BK or an already induced clause.
Since \tw{grandparent} is not part of the BK and Metagol has not yet induced any clauses, this step fails.
In Step 2b, Metagol tries to prove this atom using a metarule.
Metagol can, for instance, unify the atom with the head of the \emph{ident} metarule to form the clause:

\begin{code}
grandparent(ann,amelia):- Q(ann,amelia).
\end{code}
\noindent
Metagol saves a metasub for this clause:
\begin{code}
\tw{sub(indent,[grandparent,Q])}
\end{code}
\noindent
The symbol \tw{Q} in this metasub is still a variable.
Metagol recursively tries to prove the atom \tw{Q(ann,amelia)}.
Since there is no \tw{Q} such that \tw{Q(ann,amelia)} is true, this step fails.
Because the \emph{ident} metarule failed, Metagol removes the metasub and backtracks to try a different metarule.
Metagol unifies the atom with the \emph{chain} metarule to form the clause:
\begin{code}
grandparent(ann,amelia):- Q(ann,C),R(C,amelia).
\end{code}
\noindent
Metagol saves a metasub for this clause:
\begin{code}
\tw{sub(chain,[grandparent,Q,R])}
\end{code}
\noindent
Metagol recursively tries to prove the atoms \tw{Q(ann,C)} and \tw{R(C,amelia)}.
Suppose the recursive call to prove \tw{Q(ann,C)} succeeds by substituting \tw{Q} with \tw{mother} to form the atom \tw{mother(ann,amy)}.
This substitution binds \tw{Q} to \tw{mother} and \tw{C} to \tw{amy} which is propagated to the other atom which now becomes \tw{R(amy,amelia)}.
Metagol also proves this second atom by substituting \tw{R} with \tw{mother} to form the atom \tw{mother(amy,amelia)}.
The proof is now complete and the metasub is:
\begin{code}
\tw{sub(chain,[grandparent,mother,mother])}
\end{code}
\noindent
This metasub means that Metagol has induced the clause:
\begin{code}
grandparent(A,B):- mother(A,C),mother(C,B).
\end{code}

\noindent
After proving the example, Metagol moves to Step 1 and picks another example to generalise.
Suppose it picks the example \tw{grandparent(steve,amelia)}.
In Step 2a, Metagol tries to prove this atom using the BK, which again fails, so tries to prove this atom using an already induced clause, which also fails.
Therefore, Metagol tries to prove this atom using a metarule.
Metagol can again use the \emph{chain} metarule but with different substitutions to form the metasub:
\begin{code}
\tw{sub(chain,[grandparent,father,mother])}
\end{code}
\noindent
This metasub corresponds to the clause:
\begin{code}
grandparent(A,B):- father(A,C),mother(C,B).
\end{code}
\noindent
Metagol has now proven the first two examples by inducing the clauses:
\begin{code}
grandparent(A,B):- mother(A,C),mother(C,B).\\
grandparent(A,B):- father(A,C),mother(C,B).
\end{code}
\noindent
If given no bound on the program size, then Metagol would prove the other two examples the same way by inducing two more clauses to finally form the program:
\begin{code}
grandparent(A,B):- mother(A,C),mother(C,B).\\
grandparent(A,B):- father(A,C),mother(C,B).\\
grandparent(A,B):- father(A,C),father(C,B).\\
grandparent(A,B):- mother(A,C),father(C,B).
\end{code}

\noindent
In practice, however, Metagol would not learn this program. It would induce the following program:
\begin{code}
grandparent(A,B):- grandparent\_1(A,C),grandparent\_1(C,B).\\
grandparent\_1(A,B):- father(A,B).\\
grandparent\_1(A,B):- mother(A,B).
\end{code}
\end{example}

\noindent
In this program, the symbol \tw{grandparent\_1} is \emph{invented} and corresponds to the parent relation.
However, it is difficult to concisely illustrate PI in this example.
We, therefore, illustrate PI in Metagol with a simpler example.

\begin{example}[Predicate invention]
Suppose we have the single positive example:
\begin{epos}
\tw{f([i,l,p],p).}
\end{epos}
\noindent
Also suppose that we only have the \emph{chain} metarule and the background relations \tw{head} and \tw{tail}.
Given this input, in Step 2b, Metagol will try to use the \emph{chain} metarule to prove the example.
However, using only the given the BK and metarules, the only programs that Metagol can construct are combinations of the four clauses:
\begin{code}
f(A,B):- head(A,C),head(C,B).\\
f(A,B):- head(A,C),tail(C,B).\\
f(A,B):- tail(A,C),tail(C,B).\\
f(A,B):- tail(A,C),head(C,B).
\end{code}
\noindent
No combination of these clauses can prove the examples, so Metagol must use PI to learn a solution.
To use PI, Metagol will try to prove the example using the \emph{chain} metarule, which will lead to the
construction of the program:
\begin{code}
f([i,l,p],p):- Q([i,l,p],C),R(C,p).\\
\end{code}
\noindent
Metagol would save a metasub for this clause:
\begin{code}
sub(chain,[f,Q,R]).
\end{code}
\noindent
Metagol will then try to recursively prove both \tw{Q([i,l,p],C)} and \tw{R(C,p)}.
To prove \tw{Q([i,l,p],C)}, Metagol will say that it cannot prove it using a relation in the BK, so it will try to invent a new predicate symbol, which leads to the new atom \tw{f\_1([i,l,p],C)} and the program:
\begin{code}
f([i,l,p],p):- f\_1([i,l,p],C),R(C,p).\\
\end{code}
\noindent
Note that this binds \tw{Q} in the metasub to \tw{f\_1}.
Metagol then tries to prove the \tw{f\_1([i,l,p],C)} and \tw{R(C,p)} atoms.
To prove \tw{f\_1([i,l,p],C)}, Metagol could use the chain metarule to form the clause:
\begin{code}
f\_1([i,l,p],C):- Q2([i,l,p],D),R2(D,C).\\
\end{code}
\noindent
Metagol would save another metasub for this clause:
\begin{code}
sub(chain,[f\_1,Q2,R2]).
\end{code}
\noindent
Metagol then tries to prove the \tw{Q2([i,l,p],D)} and \tw{R2(D,C)} atoms.
Metagol can prove \tw{Q2([i,l,p],D)} by binding \tw{Q2} to \tw{tail} so that \tw{D} is bound to \tw{[l,p]}.
Metagol can then prove \tw{R2([l,p],C)} by binding \tw{R2} to \tw{tail} so that \tw{C} is bound \tw{[p]}.
Remember that the binding of variables is propagated through the program, so \tw{C} in \tw{R(C,p)} is now bound to \tw{R([p],p)}.
Metagol then tries to prove the remaining atom \tw{R([p],p)}, which it can by binding \tw{R} to \tw{head}.
The proof of all the atoms is now complete and the final metasubs are:
\begin{code}
sub(chain,[f,f\_1,head]).\\
sub(chain,[f\_1,tail,tail]).
\end{code}
\noindent
These metasubs correspond to the program:
\begin{code}
f(A,B):- f\_1(A,C),head(C,B).\\
f\_1(A,B):- tail(A,C),tail(C,B).\\
\end{code}

\end{example}

\subsubsection{Discussion}

\paragraph{Advantages.}
Metagol supports PI, learning recursive programs, and is guaranteed to learn the smallest program.
Because it uses metarules, Metagol can tightly restrict the hypothesis space, which means that it is extremely efficient at finding solutions.
The basic Metagol implementation is less than 100 lines of Prolog code, which makes Metagol easy to adapt, such as to support NAF \shortcite{siebers:negation}, types \shortcite{morel:typed}, learning higher-order programs \shortcite{metaho}, learning efficient programs \shortcite{metagolo,metaopt}, and Bayesian inference \shortcite{mugg:metabayes}.

\paragraph{Disadvantages.}
Deciding which metarules to use for a given task is a major problem.
Given too many metarules, the hypothesis space might be so large that the search is intractable.
Given insufficient metarules, the hypothesis space might be too small as to exclude a good hypothesis.
For some tasks, such as string transformations, it is straightforward to choose a suitable set of metarules because one already knows the general form of hypotheses.
However, when one has little knowledge of the solutions, then Metagol is unsuitable.
Although there is preliminary work in identifying universal sets of metarules \shortcite{minmeta,reduce-jelia,reduce}, this work mostly focuses on dyadic logic.
If a problem contains predicates of arities greater than two, then Metagol is unsuitable.
Finally, Metagol cannot handle noisy examples and struggles to learn large programs \shortcite{crop:thesis,brute,popper}.

\section{Applications}
\label{sec:apps}

\noindent
We now briefly discuss some application areas of ILP.

\paragraph{Bioinformatics and drug design.}
Perhaps the most prominent application of ILP is in bioinformatics and drug design.
ILP is especially suitable for such problems because biological structures, including molecules and protein interaction networks, can easily be expressed as relations: molecular bonds define relations between atoms and interactions define relations between proteins.
Moreover, as mentioned in the introduction, ILP induces human-readable models.
ILP can, therefore, make predictions based on the (sub)structured present in biological structures which domain experts can interpret.
The types of task ILP has been applied to include identifying and predicting ligands (substructures responsible for medical activity)~\shortcite{Finn98,DBLP:journals/ml/SrinivasanPCK06,Kaalia16}, predicting mutagenic activity of molecules and identifying structural alerts for the causes of chemical cancers~\shortcite{DBLP:conf/ilp/SrinivasanKMS97,DBLP:journals/ai/SrinivasanMSK96}, learning protein folding signatures~\shortcite{DBLP:journals/ml/TurcotteMS01}, inferring missing pathways in protein signalling networks~\shortcite{inoue:mla}, and modelling inhibition in metabolic networks~\shortcite{Kakas2006}.

\paragraph{Robot scientist.}
One of the most notable applications of ILP was in the \emph{Robot Scientist} project \shortcite{king2009automation}.
The Robot Scientist uses logical BK to represent the relationships between protein-coding sequences, enzymes, and metabolites in a pathway.
The Robot Scientist uses ILP to automatically generate hypotheses to explain data and then devises experiments to test hypotheses, run the experiments, interpret the results, and then repeat the cycle \shortcite{king2004functional}.
Whilst researching yeast-based functional genomics, the Robot Scientist became the first machine to independently discover new scientific knowledge \shortcite{king2009automation}.

\paragraph{Ecology.}
There has been much recent work on applying ILP in ecology \shortcite{bohan2011automated,DBLP:conf/ilp/Tamaddoni-Nezhad14,bohan2017next}.
For instance, \shortciteA{bohan2011automated} use ILP to generate plausible and testable hypotheses for trophic relations (`who eats whom') from ecological data.

\paragraph{Program analysis.}
Due to the expressivity of logic programs as a representation language, ILP systems have found successful applications in software design.
ILP systems have proven effective in learning SQL queries \shortcite{awscp,Sivaraman2019} and programming language semantics \shortcite{DBLP:conf/ilp/BarthaC19}.
Other applications include code search~\shortcite{Sivaraman2019}, in which an ILP system interactively learns a search query from examples, and software specification recovery from execution behaviour~\shortcite{Cohen1994,Cohen95}.

\paragraph{Data curation and transformation.}
Another successful application of ILP is in data curation and transformation, which is again largely because ILP can learn executable programs.
The most prominent example of such tasks is string transformations, such as the example given in the introduction.
There is much interest in this topic, largely due to success in synthesising programs for end-user problems, such as string transformations in Microsoft Excel \shortcite{flashfill}.
String transformations have become a standard benchmark for some recent ILP papers~\shortcite{metabias,metaho,brute,popper,playgol}.
Other transformation tasks include extracting values from semi-structured data (e.g. XML files or medical records), extracting relations from ecological papers, and spreadsheet manipulation~\shortcite{datacurate}.

\paragraph{Learning from trajectories.}
Learning from interpretation transitions (LFIT) \shortcite{inoue:lfit} automatically constructs a model of the dynamics of a system from the observation of its state transitions\footnote{The LFIT implementations are available at \url{https://github.com/Tony-sama/pylfit}}.
Given time-series data of discrete gene expression, it can learn gene interactions, thus allowing to explain and predict states changes over time \shortcite{TRMLJ2020}.
LFIT has been applied to learn biological models, like Boolean Networks, under several semantics: memory-less deterministic systems \shortcite{inoue:lfit,DBLP:conf/ilp/RibeiroI14}, probabilistic systems \shortcite{DMTRICLP15} and their multi-valued extensions \shortcite{TRICMLA15,DMTRICAPS16}.
\shortciteA{DMTRICLP15,DMTRICAPS16} combine LFIT with a reinforcement learning algorithm to learn probabilistic models with exogenous effects (effects not related to any action) from scratch.
The learner was notably integrated into a robot to perform the task of clearing the tableware on a table.
In this task external agents interacted, people brought new tableware continuously and the manipulator robot had to cooperate with mobile robots to take the tableware to the kitchen.
The learner was able to learn a usable model in just five episodes of 30 action executions.
\shortciteA{apperception} apply the \emph{Apperception Engine} to explain sequential data, such as rhythms and simple nursery tunes, image occlusion tasks, and sequence induction intelligence tests.
They show that their system can perform human-level performance.


\paragraph{Natural language processing.}
Many natural language processing tasks require an understanding of the syntax and semantics of the language.
ILP is well-suited for addressing such tasks for three reasons (i) it is based on an expressive formal language that can capture/respect the syntax and semantics of the natural language, (ii) linguistics knowledge and principles can be integrated into ILP systems, and (iii) the learnt clauses are understandable to a linguist.
ILP has been applied to learn grammars~\shortcite{Mooney95,mugg:metalearn,law:asg} and parsers~\shortcite{Zelle96,zelle96:comparative,Mooney2000} from examples.
For an extensive overview of language tasks that can benefit from ILP see the paper by \shortciteA{DzeroskiLLL2001}.

\paragraph{Physics-informed learning.}
A major strength of ILP is its ability to incorporate and exploit background knowledge.
Several ILP applications solve problems from \textit{first principles}: provided physical models of the basic primitives, ILP systems can induce the target hypothesis whose behaviour is derived from the basic primitives.
For instance, ILP systems can use a theory of light to understand images~\shortcite{DaiMWTZ17,mugg:vision}.
Similarly, simple electronic circuits can be constructed from the examples of the target behaviour and the physics of basic electrical components \shortcite{Grobelnik1992MarkusAO} and models of simple dynamical systems can be learned given the knowledge about differential equations~\shortcite{Bratko94}.

\paragraph{Robotics.}
\label{sec:robotics}
Similarly to the previous category, robotics applications often require incorporating domain knowledge or imposing certain requirements on the learnt programs.
For instance, The Robot Engineer~\shortcite{SammutSHW15} uses ILP to design tools for robots and even complete robots, which are tests in simulations and real-world environments.
Metagol$_o$~\shortcite{metagolo} learns robot strategies considering their resource efficiency and \shortciteA{AntanasMR15} recognise graspable points on objects through relational representations of objects.

\paragraph{Games.}
Inducing game rules has a long history in ILP, where chess has often been the focus \shortcite{goodacre1996inductive,DBLP:journals/ci/Morales96,chess-revision}.
For instance, \shortciteA{bain1994learning} studies inducing rules to determine the legality of moves in the chess KRK (king-rook-king) endgame.
\shortciteA{DBLP:conf/ijcai/CastilloW03} uses a top-down ILP system and active learning to induce a rule for when a square is safe in the game minesweeper.
\shortciteA{DBLP:conf/ilp/LegrasRV18} show that Aleph and TILDE can outperform an SVM learner in the game of Bridge.
\shortciteA{ilasp} uses ILASP to induce the rules for Sudoku and show that this more expressive formalism allows for game rules to be expressed more compactly.
\shortciteA{iggp} introduce the ILP problem of \emph{inductive general game playing}: the problem of inducing game rules from observations, such as \emph{Checkers}, \emph{Sokoban}, and \emph{Connect Four}.

\paragraph{Other.}
Other notable applications include learning event recognition systems ~\shortcite{iled,oled}, tracking the evolution of online communities~\shortcite{DBLP:conf/time/AthanasopoulosP18}, the MNIST dataset \shortcite{dilp}, and requirements engineering \shortcite{DBLP:journals/tse/AlrajehKRU13}.
\section{Related Work}
\label{sec:related}

\subsection{Program Synthesis}

Because ILP induces programs, it is also a form of \emph{program synthesis} \shortcite{mis}, where the goal is to build a program from a specification.
Universal induction methods, such as Solomonoff induction \shortcite{solomonoff:64a,solomonoff:64b} and Levin search \shortcite{levin:search} are forms of program synthesis.
However, universal methods are impractical because they learn only from examples and, as \shortciteA{tm:book} points out, bias-free learning is futile.

Deductive program synthesis approaches \shortcite{mana:synthesis} take full specifications as input and are efficient at building programs.
Universal induction methods take only examples as input and are inefficient at building programs.
There is an area in between called \emph{inductive program synthesis}\footnote{
Inductive program synthesis is often called \emph{program induction} \shortcite{metabias,lake2015human,crop:thesis,ellis:scc}, \emph{programming by example} \shortcite{lieberman:pbe}, and \emph{inductive programming} \shortcite{cacm:ip}, amongst many other names.
\shortciteA{gulwani2017program} divide inductive program synthesis into two categories: (i) program induction, and (ii) program synthesis.
They say that program induction approaches are neural architectures that learn a network that is capable of replicating the behaviour of a program.
By contrast, they say that program synthesis approaches output or return an interpretable program.
}.
Similar to universal induction methods, inductive program synthesis systems learn programs from incomplete specifications, typically input/output examples.
In contrast to universal induction methods, inductive program synthesis systems use BK, and are thus less general than universal methods, but are more practical because the BK is a form of inductive bias \shortcite{tm:book} which restricts the hypothesis space.
When given no BK, and thus no inductive bias, inductive program synthesis methods are equivalent to universal induction methods.

Early work on inductive program synthesis includes \shortciteA{plotkin:thesis} on least generalisation, \shortciteA{vere1975} on induction algorithms for predicate calculus, \shortciteA{summers:lisp} on inducing Lisp programs, and \shortciteA{mis} on inducing Prolog programs.
Interest in inductive program synthesis has grown recently, partly due to applications in real-world problems, such as end-user programming \shortcite{flashfill}.
Inductive program synthesis interests researchers from many areas of computer science, notably ML and programming languages (PL).
The two major\footnote{
    Minor differences include the form of specification and theoretical results.
} differences between ML and PL approaches are (i) the generality of solutions (synthesised programs) and (ii) noise handling.
PL approaches often aim to find \emph{any} program that fits the specification, regardless of whether it generalises.
Indeed, PL approaches rarely evaluate the ability of their systems to synthesise solutions that generalise, i.e.~they do not measure predictive accuracy \shortcite{lambda2,DBLP:conf/pldi/OseraZ15,awscp,syntaxguided,DBLP:journals/pacmpl/RaghothamanMZNS20}.
By contrast, the major challenge in ML (and thus ILP) is learning hypotheses that \emph{generalise} to unseen examples.
Indeed, it is often trivial to learn an overly specific solution for a given problem.
For instance, an ILP system can trivially construct the bottom clause \shortcite{progol} for each example.
Similarly, noise handling is a major problem in ML, yet is rarely considered in the PL literature.

Besides ILP, inductive program synthesis has been studied in many areas of ML, including deep learning \shortcite{deepcoder,ellis:scc,ellis:repl}.
The main advantages of neural approaches are that they can handle noisy BK, as illustrated by \dilp{}, and can harness tremendous computational power \shortcite{ellis:repl}.
However, neural methods often require many more examples \shortcite{nandopoo,DBLP:conf/iclr/DongMLWLZ19} to learn concepts that symbolic ILP can learn from just a few.
Another disadvantage of neural approaches is that they often require hand-crafted neural architectures for each domain.
For instance, the REPL approach \shortcite{ellis:repl} needs a hand-crafted grammar, interpreter, and neural architecture for each domain.
By contrast, because ILP uses logic programming as a uniform representation for examples, BK, and hypotheses, it can easily be applied to arbitrary domains.

\subsection{StarAI}

As ILP builds upon logic programs and logical foundations of knowledge representation, ILP also inherits one of their major limitations: the inability to handle uncertain or incorrect BK.
To overcome this limitation, the field of statistical relational artificial intelligence (StarAI)~\shortcite{pilp,DeRaedtKerstingEtAl16} unites logic programming with probabilistic reasoning.

StarAI formalisms allow a user to explicitly quantify the confidence in the correctness of the BK by annotating parts of BK with probabilities.
Perhaps the simplest flavour of StarAI languages, and the one that directly builds upon logic programming and Prolog, is a family of languages based on distribution semantics~\shortcite{Sato95astatistical,Sato2001ParameterLO,raedt:problog}.
Problog~\shortcite{raedt:problog}, a prominent member of this family, represents a minimal extension of Prolog that supports such stochastic execution.
Problog introduces two types of probabilistic choices: probabilistic facts and annotated disjunctions.
Probabilistic facts are the most basic stochastic unit in Problog.
They take the form of logical facts labeled with a probability $p$ and represent a Boolean random variable that is \tw{true} with probability $p$ and \tw{false} with probability $1-p$.
For instance, the following probabilistic fact states that there is 1\% chance of an earthquake in Naples.
\begin{code}
    0.01::earthquake(naples).
\end{code}
\noindent
In contrast to deterministic logic programs in which a fact is always true if so stated, the Problog engine determines the truth assignment of a probabilistic fact when it encounters it during SLD resolution, the main execution principle of logic programs.
How often a fact is deemed true or false is guided by the stated probability.
An alternative interpretation of this statement is that 1\% of executions of the probabilistic program would observe an earthquake.
Whereas probabilistic facts introduce non-deterministic behaviour on the level of facts, annotated disjunctions introduce non-determinism on the level of clauses.
Annotated disjunctions allow for multiple literals in the head, but only one of the head literals can be \tw{true} at a time.
For instance, the following annotated disjunction states that a ball can be either green, red, or blue, but not a combination of colours:
\begin{code}
    $\frac{1}{3}$::colour(B,green); $\frac{1}{3}$::colour(B,red); $\frac{1}{3}$::colour(B,blue) :- ball(B).
\end{code}
\noindent
Though StarAI frameworks allow for incorrect BK, they add another level of complexity to learning: besides identifying the right program (also called structure in StarAI), the learning task also consists of learning the corresponding probabilities of probabilistic choices (also called parameters).
Learning probabilistic logic programs is largely unexplored, with only a few existing approaches~\shortcite{probfoil,DBLP:journals/tplp/BellodiR15}.

\subsection{Neural ILP}

The successes of deep learning on various tasks have prompted researchers to investigate whether these techniques can be used to solve the ILP problem.
What makes this research direction interesting is that the techniques are based on numerical optimisation and, if successful, could learn programs without combinatorial search, which is the core reason why ILP is a difficult problem.
However, to leverage these techniques, ILP needs to be framed as a problem over a fixed set of variables, which is perhaps unnatural given that hypothesis spaces in ILP are combinatorial and essentially infinite.
Additionally, integrating neural networks into ILP would make it possible for ILP to deal with unstructured data, such as images and sound, and noise associated with unstructured data. 
While integrating ILP and program synthesis with neural networks is a proliferating area, relatively few approaches tackle the ILP problem.

The majority of existing neural-ILP systems reframe the ILP problem as {\it structure learning by parameter learning}.
Neural-ILP techniques make the ILP problem amenable to numerical optimisation by assuming that the entire hypothesis space is the solution program, instead of one member of it.
As such a solution program would be able to entail almost anything, neural-ILP techniques relax the notion of entailment.
That is, every member $c$ of the hypothesis space is associated with valuation $w_c$ which indicate the confidence that the entailments of $c$ are correct.
A fact entailed by $c$ is said to be entailed with valuation $w_c$; a valuation of a fact is then the sum of valuations of all $c$ that entail it.
Consequently, the notion of entailment loses its crispness and needs to be interpreted in the continuous spectrum. 
The learning task is then to fit the valuations such that positive examples have high valuations and the valuations of negative examples are low, for which any numerical optimisation techniques can be used.
These techniques make the hypothesis space finite by limiting its complexity and consequently only learn syntactically simple programs (e.g., the ones up to two literals and two clauses).
The prominent examples of this paradigm are \dilp{}, neural theorem provers \shortcite{DBLP:conf/nips/Rocktaschel017}, DiffLog \shortcite{DBLP:conf/ijcai/SiRHN19} and LRNN \shortcite{Sourek2018}.

At the time of publishing, only one approach can simultaneously learn a logic program that trains a neural network to solve a sensory part of the task  \shortcite{dai2020abductive}.
Dai and Muggleton \shortcite{dai2020abductive} extends ILP with an abductive step \shortcite{flach2013abduction} that deduces labels for training neural networks from background knowledge and currently explored program.
The authors demonstrate that their framework can learn arithmetic and sorting operations from digits represented as images.
This is an underexplored research direction that holds a lot of promise in making ILP more capable of handling noise and unstructured data.

\subsection{Representation Learning}

As introduced in Section \ref{sec:pi}, predicate invention (PI) -- changing the representation of a problem by introducing new predicates symbols defined in terms of provided predicates -- is one of the major open challenges in ILP. 
PI shares the goal with \emph{feature} or \emph{representation learning} \shortcite{replearn} stream of research originating in the deep learning community \shortcite{deeplearning}.
Representation learning aims to re-represent given data, be it an image or a relational knowledge base, in a \emph{vector space} such that certain semantic properties are preserved.
For instance, to represent a knowledge base (specified as a logic program) in a vector space, representation learning methods replace every constant and predicate with a vector such that the vectors of constants often appears in facts together are close in vector space.
The benefit of such representation change is that now any tabular machine learning approach can operate on complex relational structures.
The central goal of representation learning coincides with the one behind PI: improving learning performance by changing the representation of a problem.

Despite strong connections, there is little interaction between PI and representation learning.
The main challenges in transferring the ideas from representation learning to PI are their different operating principles.
It is not clear how symbolic concepts can be invented through table-based learning principles that current representation learning approaches use.
Only a few approaches~\shortcite{curled,alps,Sourek2018} start from the core ideas in representation learning, strip them of numerical principles and re-invent them from symbolic principles.
A more common approach is to transform relational data into a propositional tabular form that can be used as an input to a neural network~\shortcite{DashSVOK18,KaurKJKN19,KaurKN20}.
A disadvantage of the latter approaches is that they only apply to propositional learning tasks, not to first-order program induction tasks where infinite domains are impossible to propositionalise.
Approaches that force neural networks to invent symbolic constructs, such as \dilp{} and \emph{neural theorem provers} ~\shortcite{DBLP:conf/nips/Rocktaschel017}, do so by sacrificing the expressivity of logic (they can only learn short Datalog programs).
\section{Summary And Limitations}
\label{sec:cons}


In a survey paper from a decade ago, \shortciteA{ilp20} proposed directions for future research.
There have since been major advances in many of these directions, including in PI (Section \ref{sec:pi}), using higher-order logic as a representation language (Section \ref{sec:metarules}) and for hypotheses (Section \ref{sec:ho}), and applications in learning actions and strategies (Section \ref{sec:robotics}).
We think that these and other recent advances put ILP in a prime position to have a significant impact on AI over the next decade, especially to address the key limitations of standard forms of ML.
There are, however, still many limitations that future work should address.

\subsection{Limitations}
\label{sec:limits}

\paragraph{User-friendly systems.}

\shortciteA{ilp20} argue that a problem with ILP is the lack of well-engineered tools.
They state that whilst over 100 ILP systems have been built since the founding of ILP in 1991, less than a handful of systems can be meaningfully used by ILP researchers.
One reason is that ILP systems are often only designed as prototypes and are often not well-engineered or maintained.
Another major problem is that ILP systems are notoriously difficult to use: you often need a PhD in ILP to use any of the tools.
Even then, it is still often only the developers of a system that know how to properly use it.
This difficulty of use is compounded by ILP systems often using many different biases or even different syntax for the same biases.
For instance, although they all use mode declarations, the way of specifying a learning task in Progol, Aleph, TILDE, and ILASP varies considerably.
If it is difficult for ILP researchers to use ILP tools, then what hope do non-ILP researchers have?
For ILP to be more widely adopted both inside and outside of academia, we must develop more standardised, user-friendly, and better-engineered tools.

\paragraph{Language biases.}
ILP allows a user to provide BK and a language bias.
Both are important and powerful features, but only when used correctly.
For instance, Metagol employs metarules (Section \ref{sec:metarules}) to restrict the syntax of hypotheses and thus the hypothesis space.
If a user can provide suitable metarules, then Metagol is extremely efficient.
However, if a user cannot provide suitable metarules (which is often the case), then Metagol is almost useless.
This same brittleness applies to ILP systems that employ mode declarations (Section \ref{sec:modes}).
In theory, a user can provide very general mode declarations, such as only using a single type and allowing unlimited recall.
In practice, however, weak mode declarations often lead to very poor performance.
For good performance, users of mode-based systems often need to manually analyse a given learning task to tweak the mode declarations, often through a process of trial and error.
Moreover, if a user makes a small mistake with a mode declaration, such as giving the wrong argument type, then the ILP system is unlikely to find a good solution.
This need for an almost perfect language bias is severely holding back ILP from being widely adopted.
To address this limitation, we think that an important direction for future work is to develop techniques for automatically identifying suitable language biases.
Although there is some work on mode learning \shortcite{modelearning,DBLP:conf/ilp/FerilliEBM04,DBLP:conf/sigmod/PicadoTFP17} and work on identifying suitable metarules \shortcite{reduce}, this area of research is largely under-researched.

\paragraph{PI and abstraction.}
\shortciteA{russell:humancomp} argues that the automatic invention of new high-level concepts is the most important step needed to reach human-level AI.
New methods for PI (Section \ref{sec:pi}) have improved the ability of ILP to invent such high-level concepts. However, PI is still difficult and there are many challenges to overcome.
For instance, in \emph{inductive general game playing} \shortcite{iggp}, the task is to learn the symbolic rules of games from observations of gameplay, such as learning the rules of \emph{connect four}.
The reference solutions for the games come from the general game playing competition \shortcite{ggp} and often contain auxiliary predicates to make them simpler.
For instance, the rules for \emph{connect four} are defined in terms of definitions for lines which are themselves defined in terms of columns, rows, and diagonals.
Although these auxiliary predicates are not strictly necessary to learn the reference solution, inventing such predicates significantly reduces the size of the solution (sometimes by multiple orders of magnitude), which in turn makes them much easier to learn.
Although new methods for PI (Section \ref{sec:pi}) can invent high-level concepts, they are not yet sufficiently powerful enough to perform well on the IGGP dataset.
Making progress in this area would constitute a major advancement in ILP and a major step towards human-level AI.

\paragraph{Lifelong learning.}
Because of its symbolic representation, a key advantage of ILP is that learned knowledge can be remembered and explicitly stored in the BK.
For this reason, ILP naturally supports \emph{lifelong} \shortcite{lifelong}, \emph{multi-task} \shortcite{caruana:mtl}, and
\emph{transfer learning} \shortcite{torrey2009transfer}, which are considered essential for human-like AI \shortcite{lake:ai}.
The general idea behind all of these approaches is to reuse knowledge gained from solving one problem to help solve a different problem.
Although early work in ILP explored this form of learning \shortcite{marvin,foil}, it has been under-explored until recently \shortcite{metabias,playgol,forgetgol,celine:bottom,knorf}, mostly because of new techniques for PI.
For instance, \shortciteA{metabias} learn 17 string transformations programs over time and show that their multi-task approach performs better than a single-task approach because learned programs are frequently reused.
However, these approaches have only been demonstrated on a small number of tasks.
To reach human-level AI, we would expect a learner to learn thousands or even millions of concepts.
But handling the complexity of thousands of tasks is challenging because, as we explained in Section \ref{sec:bk}, ILP systems struggle to handle large amounts of BK.
This situation leads to the problem of \emph{catastrophic remembering} \shortcite{forgetgol}: the inability for a learner to forget knowledge.
Although there is initial work on this topic \shortcite{forgetgol}, we think that a key area for future work is handling the complexity of lifelong learning.

\paragraph{Relevance.}
The \emph{catastrophic remembering} problem is essentially the problem of \emph{relevance}: given a new ILP problem with lots of BK, how does an ILP system decide which BK is relevant?
Although too much irrelevant BK is detrimental to learning performance \shortcite{ashwin:1995,ashwin:badbk}, there is almost no work in ILP on trying to identify relevant BK.
One emerging technique is to train a neural network to score how relevant programs are in the BK and to then only use BK with the highest score to learn programs \shortcite{deepcoder,ellis:scc}.
However, the empirical efficacy of this approach has yet to be clearly demonstrated.
Moreover, these approaches have only been demonstrated on small amounts of BK and it is unclear how they scale to BK with thousands of relations.
Without efficient relevancy methods, it is unclear how lifelong learning can be achieved.

\paragraph{Noisy BK.}
Another issue related to lifelong learning is the underlying uncertainty associated with adding learned programs to the BK.
By the inherent nature of induction, induced programs are not guaranteed to be correct (i.e.~are expected to be noisy), yet they are the building blocks for subsequent induction.
Building noisy programs on top of other noisy programs could lead to eventual incoherence of the learned program.
This issue is especially problematic because, as mentioned in Section \ref{sec:noise}, most ILP approaches assume noiseless BK, i.e.~a relation is true or false without any room for uncertainty.
One of the appealing features of \dilp{} is that it takes a differentiable approach to ILP, where it can be provided with fuzzy or ambiguous data.
Developing similar techniques to handle noisy BK is an under-explored topic in ILP.

\paragraph{Probabilistic ILP.}
A principled way to handle noise is to unify logical and probabilistic reasoning, which is the focus of \emph{statistical relational artificial intelligence} (StarAI) \shortcite{DeRaedtKerstingEtAl16}.
While StarAI is a growing field, inducing probabilistic logic programs has received little attention, with few notable exceptions \shortcite{DBLP:journals/tplp/BellodiR15,probfoil}, as inference remains the main challenge. Addressing this issue, i.e.~unifying probability and logic in an inductive setting, would be a major achievement \shortcite{marcus:2018}.

\paragraph{Explainability.}
Explainability is one of the claimed advantages of a symbolic representation. Recent work \shortcite{mugg:compmlj,mugg:beneficial} evaluates the comprehensibility of ILP hypotheses using Michie's \citeyear{usml} framework of \emph{ultra-strong ML}, where a learned hypothesis is expected to not only be accurate but to also demonstrably improve the performance of a human being provided with the learned hypothesis.
\shortciteA{mugg:compmlj} empirically demonstrate improved human understanding directly through learned hypotheses. However, more work is required to better understand the conditions under which this can be achieved, especially given the rise of PI.

\paragraph{Learning from raw data.}
\label{sec:noisyinput}
Most ILP systems require data in perfect symbolic form.
However, much real-world data, such as images and speech, cannot easily be translated into a symbolic form.
Perhaps the biggest challenge in ILP is to learn how to both perceive sensory input and learn a symbolic logic program to explain the input.
For instance, consider a task of learning to perform addition from MNIST digits.
Current ILP systems need to be given as BK symbolic representations of the digits, which could be achieved by first training a neural network to recognise the digits.
Ideally, we would not want to treat the two problems separately, but rather simultaneously learn how to recognise the digits and learn a program to perform the addition.
A handful of approaches have started to tackle this problem \shortcite{deepproblog,NIPS2019_8548,apperception,dai2020abductive}, but developing better ILP techniques that can both perceive sensory input and learn complex relational programs would be a major breakthrough not only for ILP, but the whole of AI.

\subsection*{Further Reading}
For an introduction to the fundamentals of logic and automated reasoning, we recommend the book of \shortciteA{harrison:atp}.
To read more about ILP, then we suggest starting with the founding paper by \shortciteA{mugg:ilp} and a survey paper that soon followed \shortcite{mugg:ilp94}.
For a detailed exposition of the theory of ILP, we thoroughly recommend the books of \shortciteA{ilp:book} and \shortciteA{luc:book}.
\section*{Acknowledgements}
We thank Céline Hocquette, Jonas Schouterden, Jonas Soenen, Tom Silver, Tony Ribeiro, and Oghenejokpeme Orhobor for helpful comments and suggestions.
We also sincerely thank the anonymous reviewers, especially Reviewer 2, whose feedback and suggestions greatly improved the paper.
Andrew Cropper was supported by the EPSRC fellowship \emph{The Automatic Computer Scientist} (EP/V040340/1).
Sebastijan Dumančić was partially supported by Research Foundation - Flanders (FWO; 12ZE520N).

\bibliography{ourbib15}
\bibliographystyle{theapa}

\end{document}